\documentclass{article}

\usepackage{microtype}
\usepackage{graphicx}
\usepackage{subfigure}
\usepackage{booktabs} 
\usepackage{framed}
\usepackage[framemethod=TikZ]{mdframed}
\usepackage{tikz}
\usepackage{pgfplots}
\usetikzlibrary{calc}
\usepackage{etoc}
\usepackage[table,dvipsnames]{xcolor}

\usepackage{hyperref}
\newcommand*\rot{\rotatebox{90}}
\newcommand{\VICReg}{\operatorname{VICReg}}

\usepackage[accepted]{icml2026}

\usepackage{amsmath}
\usepackage{amssymb}
\usepackage{mathtools}
\usepackage{amsthm}

\usepackage[capitalize,noabbrev]{cleveref}

\usepackage{booktabs}
\usepackage{multirow}
\usepackage{float}

\usepackage[utf8]{inputenc} 
\usepackage[T1]{fontenc}    
\usepackage{hyperref}       
\usepackage{url}            
\usepackage{booktabs}       
\usepackage{amsfonts}       
\usepackage{nicefrac}       
\usepackage{microtype}      
\usepackage{booktabs}
\usepackage[colorinlistoftodos,prependcaption,textsize=tiny]{todonotes}
\usepackage{lipsum}
\usepackage{bbm}
\usepackage{arydshln}
\usepackage{amsmath}
\usepackage{comment}
\usepackage{wrapfig}
\usepackage{soul}
\usepackage[dvipsnames]{xcolor}

\usepackage[most]{tcolorbox}
\usepackage{listings}

\newtcolorbox{promptbox}{
    colback=gray!5!white,
    colframe=gray!20!white,
    arc=5pt,
    boxrule=0.5pt,
    left=15pt,
    right=15pt,
    top=15pt,
    bottom=15pt,
    fontupper=\small\ttfamily, 
    enhanced,
    breakable
}

\usepackage{amsmath,amssymb,amsfonts}
\usepackage{amsmath,amstext}
\usepackage{graphicx}
\usepackage{textcomp}
\usepackage{bm}
\usepackage{xspace}
\newcommand{\methodname}{FairSSL\xspace} 

\usepackage{comment}

\theoremstyle{plain}
\newtheorem{theorem}{Theorem}[section]

\newtheorem{lemma}[theorem]{Lemma}
\newtheorem{remark}[theorem]{Remark}
\newtheorem{corollary}[theorem]{Corollary}
\theoremstyle{definition}

\theoremstyle{remark}

\newcommand{\first}[1]{\textbf{#1}}
\newcommand{\second}[1]{\textcolor{gray}{\underline{#1}}}

\newcommand{\methodsuperscript}{\scriptsize \mathrm{FSSL}}

\usepackage[textsize=tiny]{todonotes}
\usepackage{setspace}

\icmltitlerunning{\methodname: Fair Multimodal Self-Supervised Learning}

\begin{document}

\twocolumn
[
\icmltitle{\methodname: Fair Multimodal Self-Supervised Learning}




\icmlsetsymbol{equal}{*}

\begin{icmlauthorlist}
\icmlauthor{Jiaee Cheong}{equal,1,2,5}
\icmlauthor{Abtin Mogharabin}{equal,3,4}
\icmlauthor{Paul Pu Liang}{5}
\icmlauthor{Hatice Gunes}{2}
\icmlauthor{Sinan Kalkan}{6}
\end{icmlauthorlist}

\icmlaffiliation{1}{Harvard University, US.}
\icmlaffiliation{2}{University of Cambridge, UK.}
\icmlaffiliation{3}{University of Tübingen, Germany.}
\icmlaffiliation{4}{Zuse School ELIZA, Germany.}
\icmlaffiliation{5}{MIT, US.}
\icmlaffiliation{6}{ROMER \& Comp. Eng, METU, Türkiye}


\icmlcorrespondingauthor{Jiaee Cheong}{jc2208@cam.ac.uk}

\icmlkeywords{Machine Learning, ICML}

\vskip 0.3in
]



\printAffiliationsAndNotice{\icmlEqualContribution} 

\begin{abstract}


Prevalent multimodal self-supervised learning (SSL) methods rely on the \textbf{redundancy assumption}: that different views share substantial task-relevant information. We argue that this assumption fails in complex, real-world settings characterized by heterogeneity (e.g., variable-length healthcare or behavioral data), where enforcing strict alignment can discard unique, modality-specific signals and inadvertently amplify bias.
In this work, we propose \textbf{\methodname}, a framework that leverages data heterogeneity as a resource for fairness rather than a hindrance. Unlike standard contrastive approaches, \methodname uses a subject-aware Variance-Invariance-Covariance Regularization objective, where alignment is enforced across segments drawn from the same subject. We introduce a segment-based pooling strategy to handle variable-length modalities, and we regularize representations to encourage (i) sufficient within-subject variability, (ii) cross-modal and cross-subject invariance, and (iii) representation decorrelation. Theoretical analysis shows that our objective bounds the score gap between protected groups. Empirically, \methodname significantly outperforms existing baselines on heterogeneous multimodal datasets, improving fairness without sacrificing downstream predictive performance.
Code available at: \href{https://github.com/abtinmU/FairSSL}{https://github.com/abtinmU/FairSSL}

\end{abstract}


\section{Introduction}

\begin{figure}[hbt!]
    \centering
    \includegraphics[width=0.99\linewidth]{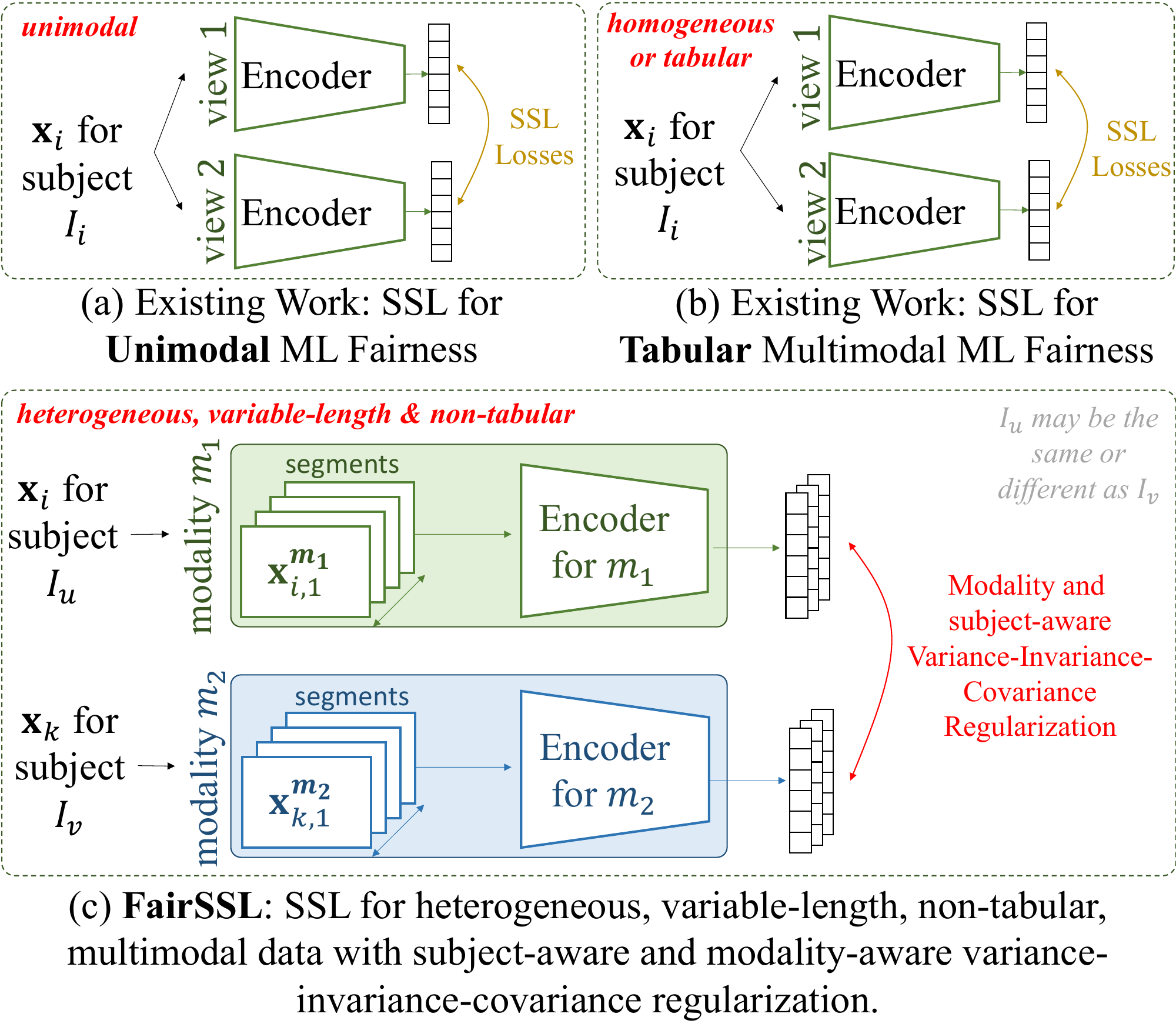}
    \caption{(a,b) Prior work has explored SSL for ML fairness in unimodal or tabular data settings. (c) \methodname addresses the challenges of non-tabular multimodal data in a subject-aware and modality-aware manner. }
    \label{fig:teaser}
    \vspace{-5mm}
\end{figure}

Multimodal machine learning (ML) has emerged as a fundamental paradigm in modern ML research
\cite{chen2020simple,liang2024foundations,radford2021learning},
with self-supervised learning (SSL) playing a key role in its recent success
\cite{zong2024self,chen2021multimodal,song2024quest,liang2023factorized}.
Although early efforts have started investigating multimodal fairness, findings remain contradictory or inconclusive. Some works show that multimodal ML can marginally improve prediction at the cost of reducing fairness
\cite{booth2021bias}, while others report improvements in both performance and fairness beyond the existing performance-fairness frontier \cite{cheong_u-fair}.
Further, it is unclear how multimodal SSL differs from conventional multimodal ML with respect to both predictive performance and fairness. 
\textbf{Research Gap 1 (RG1) (Fairness of multimodal SSL vs. supervised multimodal ML):} prior work reports mixed fairness outcomes for multimodal supervised models, and it remains unclear whether (and when) multimodal SSL changes the performance--fairness trade-off.
%


\begin{table*}[hbt!]
\addtolength{\tabcolsep}{-0.5mm}
\caption{Comparative Summary with existing Multimodal Fairness SSL studies. Abbreviations (sorted): 
  A: Audio. 
  BM: Bias Mitigation.
  Cont: Contrastive.
  EAcc: Equal Accuracy. 
  EEG: Electroencephalogram.
  EOdd: Equalised Odds. 
  EOpp: Equality of Opportunity. 
  MM: Multi-modal.
  Red: Redundancy reduction.
  SP: Statistical Parity. T: Text. V: Visual.
  VL: deals with heterogeneous data of \underline{v}arying \underline{l}ength.
  }   \label{tab:comparative_summary}
  \centering
  \scriptsize
  \begin{tabular}{l|c|ccc|ccc|cccc}
    \hline
     &        & \multicolumn{3}{c|}{\textbf{Approach}}      &\multicolumn{3}{c|}{\textbf{Evaluation}}     &\multicolumn{4}{c}{\textbf{Fairness Measures}}  \\
   
    
    \hspace*{1cm}
    {\textbf{Study}} & \textbf{Task} & {MM} &{Modality} &{\textcolor{Green}{SSL}}  &BM &VL 
     & AU-ROC 
    &SP &EOpp  &EOdd &EAcc  \\
    \hline

      \citet{alasadi_cyberbullying}  &Cyberbullying Detection  &\checkmark & VT &  &\checkmark  &\checkmark  &   &\checkmark &&\checkmark   &  \\ 
   
    \citet{schmitz2022bias}  &Emotion Detection &\checkmark & AVT &  &  &\checkmark &  &\checkmark &\checkmark &   &   \\ 

    \citet{yan2020mitigating}   &Personality Assessment &\checkmark &AV &  &\checkmark &\checkmark &  &\checkmark & & &\checkmark \\ 

   \citet{kathan2022personalised} &Humour Recognition  &\checkmark & AV &  &\checkmark &\checkmark  & &\checkmark && & \\ 
    
    \citet{chen2023fmmrec} &Recommendation  &\checkmark & AVT &   &\checkmark &\checkmark &  & &\checkmark&\checkmark &  \\ 
    
    \citet{janghorbani2023multi}  &Vision-Language Models  &\checkmark & VT &  &\checkmark  &   &    & && &\\ 
    
    \citet{pena2023human} &Automatic Recruitment  &\checkmark & VT &  
    &\checkmark &\checkmark &  &\checkmark && & 
    \\ 

   
    \cdashline{1-12}[.4pt/2pt]   
    
    \citet{barker2024learning}  &Tabular \& Language &   &  tabular, T & \textcolor{Green}{\checkmark} (Red.) &\checkmark &   & &\checkmark &\checkmark &\checkmark &\checkmark   \\ 

    \citet{yfantidou2024using}
    & Human-centred datasets &  &tabular & \textcolor{Green}{\checkmark} (Cont.) &  & &\checkmark &&& &\\ 
   
   \hline
   \rowcolor{gray!10} \textbf{\methodname (Ours)} & Healthcare   &\checkmark    &AV, A-EEG, tabular & \textcolor{Green}{\checkmark} (Red.) &\checkmark  &\checkmark  &\checkmark  &\checkmark 
   &\checkmark  &\checkmark  &\checkmark     \\

    \hline
  \end{tabular}
  
\end{table*}

Further, while prior work has extensively examined \textit{when, how and why} multimodal SSL can improve downstream task performance \cite{zong2024self,chen2021multimodal,song2024quest,liang2023factorized,wang2024decur}.
For instance, on tasks that typically require unique information from different modalities (e.g. sarcasm detection), inter-modality alignment, and performance are weakly correlated or even negatively correlated \cite{liang2023factorized,tjandrasuwitaunderstanding}.
From a fairness perspective, this thus introduces the following additional fundamental question and challenge: assuming a prediction task involving two different subgroups, 
how can we maximise the intra and inter-modality task-relevant signals and yet remove the spurious features introduced by a sensitive attribute, say gender? 
Conventional multimodal SSL wisdom dictates that, when the two modalities are fully redundant, the alignment is strongly correlated with performance -- is this also true for fairness?
\textbf{RG2 (What enables fairness under heterogeneity):} existing multimodal SSL insights are largely performance-centric, leaving it unclear which properties of multimodal SSL objectives and alignment strategies actually drive fairness improvements in heterogeneous settings.

As a result, there remain fundamental challenges and open questions concerning the applicability and effectiveness of existing SSL methods in highly heterogeneous real-world settings.
In particular, data heterogeneity made it challenging to advance multimodal fairness using SSL-based methods.
This is because most existing SSL approaches are contrastive, focusing on maximizing mutual information between views
\cite{tschannenmutual,radford2021learning,chen2020simple}
while ignoring modality-unique information \cite{liang2023factorized}.
These methods rely on the critical assumption that different views 
share substantial task-relevant information, commonly referred to as multi-view redundancy
\cite{liang2023factorized,song2024quest,liu2023focal}.
However, this assumption frequently fails in complex datasets characterized by highly heterogeneous modalities with varying lengths and temporal structures, resulting in minimal inter-modal overlap.

In such settings, prior studies have shown that standard contrastive learning
(CL) will discard task-relevant unique information,
and ultimately degrade downstream performance \cite{robinson2021can,xiaoshould,liu2023focal,liang2023factorized}.
\textbf{RG 3 (Heterogeneity-aware SSL for variable-length modalities):} most multimodal SSL methods assume substantial shared information across modalities, an assumption that often breaks for variable-length and temporally heterogeneous data; this raises the need for SSL objectives and alignment mechanisms that can preserve modality-unique signals while mitigating bias.
Our key contributions (\textbf{KC}) are as follows: 
\textbf{KC1:} We identify and formalize fairness challenges intrinsic to multimodal ML and demonstrate how modality-specific data heterogeneity can be leveraged as an advantage for both multimodal representation learning and fairness (addresses \textbf{RG2}).
\textbf{KC2:}
We introduce a novel subject-aware self-supervised method, \methodname,
that mitigates bias in multimodal settings with highly heterogeneous data of variable-length
(Fig. \ref{fig:teaser}) (addresses \textbf{RGs 2-3}). 
We further exploit a key underpinning currently missed within the literature: 
different modalities may contain varying levels of individual and sensitive-attribute dependent information which can guide fairer and more robust learnt representations at different time-points.
First,
we perform subject-aware changes on the loss function such that the variance term reduces its reliance on the protected attribute as a trivial solution, 
the invariance term ensures consistent predictions for similar individuals, 
and the covariance term minimizes correlational dependence on the protected attribute.
Second, 
we introduce (i) segment-based encoding and (ii) segment-based pooling 
such that we are still able to learn good and fair representations from
modalities of variable feature length with temporality varying levels of task-relevant signals to address data heterogeneity.
\textbf{KC3:}
We provide extensive empirical experiments and theoretical proof validating our method (addresses \textbf{RGs 1-3}).



\section{Related Work}
\label{sect:rel_work}

We review additional background in Appendix~\ref{app:extended_related_work}; here we summarize the most relevant threads and position our contributions (see Table \ref{tab:comparative_summary}).

\noindent\textbf{Multimodal self-supervised learning.}
Multimodal SSL has been widely studied via contrastive and non-contrastive objectives that encourage agreement across views/modalities \cite{radford2021learning,chen2020simple,chen2021multimodal}. However, many approaches implicitly benefit from (or assume) substantial task-relevant overlap across modalities; when modalities are heterogeneous and contain significant modality-unique information, stronger alignment can be weakly correlated or even negatively correlated with downstream performance \cite{liang2023factorized,tjandrasuwitaunderstanding}.

\noindent\textbf{Heterogeneity and modality-unique information.}
Recent work has emphasized that multimodal objectives can suppress modality-unique signals in low-overlap settings, leading to shortcut learning and degraded downstream performance \cite{robinson2021can,liu2023focal,liang2023factorized}. In contrast to prior studies that primarily analyze these phenomena through the lens of predictive performance, we focus on how these objective-level choices interact with downstream fairness.

\noindent\textbf{Fairness in multimodal learning and SSL.}
Existing supervised multimodal fairness methods typically target specific modalities or fusion architectures, and do not systematically address fairness concerns that arise from multimodal interactions (the degree of task-relevant information shared across modalities) and heterogeneity (the degree of similarity across modalities independent of the task). Meanwhile, SSL has been shown to learn fairer representations in some unimodal settings \cite{yfantidou2024using}, but the fairness implications of multimodal SSL under variable-length heterogeneous modalities remain less explored.

\textbf{Overall}, our work differs from prior multimodal SSL by explicitly leveraging heterogeneity to preserve modality-unique signals while mitigating bias, and by providing both theoretical and empirical evidence for the resulting fairness-performance trade-offs.

\section{Preliminaries and Background}

\textbf{Problem Definition and Notation:}
We have a dataset $\mathcal{D} = \{(\mathbf{x}_i, y_i)\}_i$ for a supervised classification problem, where $\mathbf{x}_i \in {X}$ is the input representing information about an individual $I_i \in \mathcal{I}$ and $y_i\in Y$ is the outcome (e.g. 1 depressed vs. 0 non-depressed) that we wish to predict. 
Within the context of our work, we work with a binary setting where $y_i\in \{0,1\}$.
Each input $\mathbf{x}_i$ is composed of multiple modalities: i.e., 
$\mathbf{x}_i=\{\mathbf{x}_i^m \in X^m\}_m$, 
where $m$ can be e.g., ``image'', ``eeg'', or ``audio''. 
Note that, although our
experiments focus on a bi-modal setting,
\methodname can easily be extended to problems with more than two modalities. 
The input for each modality $\mathbf{x}_i^m$ is preprocessed into $N_m$-many fixed-length segments:
\begin{equation}\footnotesize
\mathbf{x}_i^m = \bigl\{\mathbf{x}^m_{i,1},\dots,\mathbf{x}_{i,N_{m}}^m\bigr\}.
\end{equation}
Each input $\mathbf{x}_i$ is associated (through an individual $I_i$) with a demographic group (sensitive attribute) $g_i \in G$ where, e.g., $G = \{\textrm{male}, \textrm{female}\}$. 
The goal in fair ML is to ensure that the outcomes for two different demographic groups $g_1$ and $g_2$ satisfy the fairness measures listed in Section \ref{sect:fairness_measures}.
%


\subsection{Background: VICReg}
\label{sect:vicreg}

Variance-Invariance-Covariance Regularization (VICReg) \cite{bardesvicreg} is a self-supervised learning (SSL) method that can be applied in multimodal settings. 
In a conventional SSL setting, we first generate two different views $\{\mathbf{x}'_i=t'(\mathbf{x}_i)\}$ and $\{\mathbf{x}''_i=t''(\mathbf{x}_i)\}$ of the same inputs $\{\mathbf{x}_i\}$ using some random transformations $t'()$ and $t''()$ (e.g., rotation, translation, cropping). The goal in SSL is to ensure that the representations $\{\mathbf{z}'_i=f'_{\theta'}(\mathbf{x}'_i)\}$ and $\{\mathbf{z}''_i=f''_{\theta''}(\mathbf{x}''_i)\}$ for the two different views obtained by deep networks $f'_{\theta'}$ and $f''_{\theta''}$ 
are similar.  
In other words, we want the representations of both modalities to be \textit{aligned}.
VICReg defines three regularization terms to enforce similarity and discriminativeness of the representations $\{\mathbf{z}'_i\}$ and $\{\mathbf{z}''_i\}$:

\noindent\textbf{(1) Variance regularization} aims to have at least certain standard deviation ($\gamma$) among the embeddings in one branch (modality) to avoid feature collapse:
\begin{equation}\label{eq:variance}  \footnotesize
    V_{reg}(\{\mathbf{z}_i\}) = \frac{1}{d} \sum_{j=1}^d \max\left(0, \gamma - \sqrt{\textrm{Var}(\{\mathbf{z}_i[j]\}) + \epsilon}\right),
\end{equation}
\noindent where $d$ is the no. of dimensions of $\mathbf{z}$; 
$\mathbf{z}[j]$ denotes the $j^{th}$ dimension; 
$\epsilon$ is a constant (set to 1 in the original paper) and $\gamma$ is a hyperparameter.
%


\vspace{2mm}
\noindent\textbf{(2) Invariance regularization} ensures that the representations through the two branches are similar:
\begin{equation}    \footnotesize
\label{invarianceVICReg}
    I_{reg}(\{\mathbf{z}'_i\}, \{\mathbf{z}''_i\}) = \frac{1}{n} \sum_{i} \lVert \mathbf{z}'_i -  \mathbf{z}''_i\rVert^2_2,
\end{equation}
where $n$ is the batch size and $\lVert \mathbf{z}'_i -  \mathbf{z}''_i\rVert^2_2$ is the Euclidean distance between vectors $\mathbf{z}'_i$ and $\mathbf{z}''_i$.

\vspace{2mm}
\noindent\textbf{(3) Covariance regularization} enforces different dimensions to be decorrelated:
\begin{equation}\footnotesize
    C_{reg}(\{\mathbf{z}_i\}) = \frac{1}{d} \sum_{j\neq k} [\textrm{Cov}(\{\mathbf{z}_i\})]_{j,k}^2,
    \label{eqCovReg}
\end{equation}
where $\textrm{Cov}(\cdot)$ is the covariance matrix for its argument set and $j\neq k$ ensures that values at off-diagonal positions of the covariance matrix are minimized.

%
For convenience, we will use $\textrm{VICReg}(\cdot, \cdot)$ to denote the following combination of the individual loss functions:
\begin{equation}    \footnotesize
\scriptstyle
\begin{aligned}
\VICReg(F_1,F_2)
  &= I_{\text{reg}}(F_1,F_2) \\
  &\quad + \mu\bigl(V_{\text{reg}}(F_1) + V_{\text{reg}}(F_2)\bigr) \\
  &\quad + \nu\bigl(C_{\text{reg}}(F_1) + C_{\text{reg}}(F_2)\bigr).
\end{aligned}
\end{equation}
where $F_1=\{\mathbf{z}'_i\}$ and $F_2=\{\mathbf{z}''_i\}$ are the sets of feature vectors from two different views (or modalities in our case).


\begin{figure}
    \centering
    \includegraphics[width=0.99\linewidth]{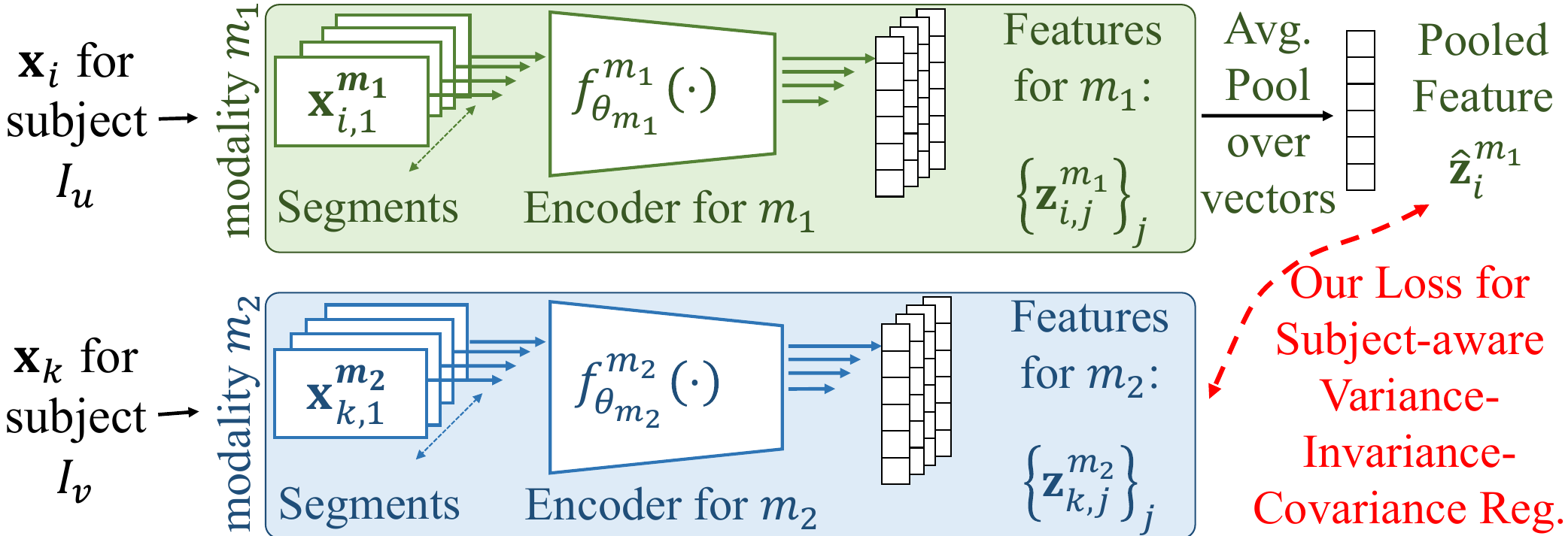}
    \caption{\methodname processes each modality for the same or different subjects and regularizes  representations in a subject-aware manner.}
    \label{fig:overview}
\end{figure}


\section{Proposed Method: \methodname}
\label{sect:method}

As summarized in Fig. \ref{fig:overview}, \methodname
takes in data from two different modalities with varying length, encodes them and modifies the VICReg loss in a subject-aware manner to eliminate any biases associated with demographic groups.

\subsection{\methodname: Overall Approach}

Denoting the embeddings extracted from the encoders for two modalities $m_1$ and $m_2$ for input $\mathbf{x}_i$ for a subject $I_i$ as $\mathbf{z}^{m_1}_{i}$ and $\mathbf{z}^{m_2}_{i}$ respectively, we introduce four variations of \methodname. 
To be able to work with variable-length data, we split such data into segments. Thus, 
given a variable length input $\mathbf{x}^m_i$ for individual $I_i$ for modality $m\in \{m_1, m_2\}$, we have $\{\mathbf{x}^{m}_{i,j}\}_{j}$, 
with $j$ denoting the index for the $j^{\textrm{th}}$ segment. 
Each segment $\mathbf{x}^{m}_{i,j}$ is processed by the encoder of the modality $f^m_{\theta^m}$ separately, yielding a set of representations for that modality: $\{\mathbf{z}^{m}_{i,j}=f^m_{\theta^m}(\mathbf{x}^m_{i,j})\}_{j}$.
Among the three terms (variance, invariance and covariance), variance and covariance are applied to each encoder (modality) independently and hence, do not require any modifications. 
However, the invariance term, enforcing a constraint between the representations of the different encoders, needs to be adapted. 
We do so by using average pooling over the embeddings of modality 1, \(\{\mathbf{z}^{m_1}_{i,j}\}_j\) to compute a pooled vector \(\hat{\mathbf{z}}^{m_1}_i\). For modality 2, we use the set of segment embeddings \(\{\mathbf{z}^{m_2}_{i,j}\}_j\) directly. The \methodname-specific invariance term then becomes:
\begin{equation} \footnotesize
I_{reg}^{{\methodsuperscript}}(\hat{\mathbf{z}}^{m_1}_i, \{\mathbf{z}^{m_2}_{i,j}\}_j)
=\frac{1}{{N_{m_2}}}\sum_{j=1}^{N_{m_2}}\|\hat{\mathbf{z}}^{m_1}_i - \mathbf{z}^{m_2}_{i,j}\|_{2}^{2}.
\label{eq:align}
\end{equation}
We apply average pooling to only one modality with the purpose of aligning its pooled vector against each segment in the other modality, thus enabling the model to identify which segments most strongly drive the invariance loss. 
In an ablation study demonstrated in Section \ref{sect:experiments}, we also experimented with 
\textit{double-pooling}, i.e.
pooling both modalities.
We now outline the four different ways \methodname has been deployed within our experiments. 


\subsection{\methodname: Intra-Subject Regularization (M1)}

This version of \methodname aims to align the average-pooled modality-1 vector $\hat{\mathbf{z}}^{m_1}_i$ for each subject $I_i$ with that same subject’s modality-2 segments $\{\mathbf{z}^{m_2,s_j}_i\}_j$. Given a subject in the current batch \(I_i \in\mathcal B\), this method applies the invariance term within subject only:
\begin{align} \footnotesize
\mathcal{L}_{\mathrm{M1}}^{\mathrm{\methodsuperscript}}
  = \frac{1}{|\mathcal{B}|} & \sum_{I_i \in \mathcal B}  \Big[
      \lambda \, I^{\mathrm{\methodsuperscript}}_{reg}\!\bigl(\hat{\mathbf{z}}^{m_1}_i, \{\mathbf{z}^{m_2}_{i,j}\}_j\bigr)\\
  &\quad + \mu\bigl(V_{\text{reg}}(\{\mathbf{z}^{m_1}_{i,j}\}_j) + V_{\text{reg}}(\{\mathbf{z}^{m_2}_{i,j}\}_j)\bigr) \nonumber \\
  &\quad + \nu\bigl(C_{\text{reg}}(\{\mathbf{z}^{m_1}_{i,j}\}_j) + C_{\text{reg}}(\{\mathbf{z}^{m_2}_{i,j}\}_j)\bigr) \Big] \nonumber.
\end{align}

\subsection{\methodname: Inter-Subject Regularization (M2)}

Different subjects can share common information within their multimodal data. 
To exploit this, we introduce an inter-subject regularization 
within each batch $\mathcal{B}$ as follows:
\begin{align}   \footnotesize
\mathcal{L}_{\mathrm{M2}}^{\mathrm{\methodsuperscript}}
  = \frac{1}{|\mathcal{B}|^2} & \sum_{I_i, I_k \in \mathcal B} \Big[
      \lambda \, I^{\mathrm{\methodsuperscript}}_{reg}\!\bigl(\hat{\mathbf{z}}^{m_1}_i, \{\mathbf{z}^{m_2}_{k,j}\}_j\bigr)\\
  &\quad + \mu\bigl(V_{\text{reg}}(\{\mathbf{z}^{m_1}_{i,j}\}_j) + V_{\text{reg}}(\{\mathbf{z}^{m_2}_{k,j}\}_j)\bigr) \nonumber \\
  &\quad + \nu\bigl(C_{\text{reg}}(\{\mathbf{z}^{m_1}_{i,j}\}_j) + C_{\text{reg}}(\{\mathbf{z}^{m_2}_{k,j}\}_j)\bigr) \Big] \nonumber.
\end{align}


\subsection{\methodname: Class-based Regularization (M3)}

Method 2 neglects the target prediction class of the ML task. 
Since subjects with the same class
are likely to share similar task-relevant features
captured through their multimodal data, Method 3 explores a regularization approach by enforcing $y_{i} = y_{k}$ for subjects $I_i$ and $I_k$:
\begin{align}   \footnotesize
\mathcal{L}_{\mathrm{M3}}^{\mathrm{\methodsuperscript}}
  = \frac{1}{|\mathcal{B}|^2} & \sum_{I_i, I_k \in \mathcal B} \Big[
      \lambda \, I^{\mathrm{\methodsuperscript}}_{reg}\!\bigl(\hat{\mathbf{z}}^{m_1}_i, \{\mathbf{z}^{m_2}_{k,j}\}_j\bigr)\\
  &\quad + \mu\bigl(V_{\text{reg}}(\{\mathbf{z}^{m_1}_{i,j}\}_j) + V_{\text{reg}}(\{\mathbf{z}^{m_2}_{k,j}\}_j)\bigr) \nonumber \\
  &\quad + \nu\bigl(C_{\text{reg}}(\{\mathbf{z}^{m_1}_{i,j}\}_j) + C_{\text{reg}}(\{\mathbf{z}^{m_2}_{k,j}\}_j)\bigr) \Big] \nonumber,
\end{align}
\[
\quad\text{subject to}\ \ y_{i}=y_k, \quad \forall\,I_i, I_k \in \mathcal{B}.
\]
%


\subsection{\methodname: Alternating Regularization (M4)}
Method 3 enforces the whole batch to have the same prediction class, which can introduce bias into the training dynamics. To address this, Method 4 alternates
between $\mathcal{L}_{\mathrm{M2}}^{\mathrm{\methodsuperscript}}$ and $\mathcal{L}_{\mathrm{M3}}^{\mathrm{\methodsuperscript}}$
as follows ($e$: epoch index):
\begin{equation}  \footnotesize
\mathcal{L}_{\mathrm{M4}}^{\mathrm{\methodsuperscript}} =
\begin{cases}
\mathcal{L}_{\mathrm{M2}}^{\mathrm{\methodsuperscript}}, & e\bmod 2 = 1 \quad

     \\[6pt]
\mathcal{L}_{\mathrm{M3}}^{\mathrm{\methodsuperscript}}, & e\bmod 2 = 0 \quad
\end{cases}
\end{equation}

\subsection{Theoretical Link between \methodname and Fairness}
\label{sect:theoretical_link}

We summarize the main theoretical results here and leave the detailed derivations and the specifics to Appendix~\ref{sect:theory}.
Consider the embedding $\mathbf{z}$ learned by \methodname for a sample with protected attribute $g \in \{0, 1\}$. We follow the standard linear-probe setting for representation learning and analyse a downstream binary classifier: $h(\mathbf{z})=\mathbf{w}^\top \mathbf{z}$, followed by the Sigmoid ($\sigma$) that produces predicted probabilities $\hat{p}(y=1 \mid \mathbf{z},g) = \sigma(h(\mathbf{z}))$.
We define the group mean embeddings $\mu_g := \mathbb{E}[\mathbf{z} \mid g]$ for $g \in \{0,1\}$. The corresponding mean scores are $\bar{h}_g(\mathbf{w}) := \mathbb{E}[h(\mathbf{z}) \mid g] = \mathbf{w}^\top \mu_g$, and we define the group score disparity between two groups as:
$\Delta_g(\mathbf{w}) := \bar{h}_1(\mathbf{w}) - \bar{h}_0(\mathbf{w})        = \mathbf{w}^\top (\mu_1 - \mu_0)$.
For each group $g$, we also define the within group score deviation as $v_g(\mathbf{w}) := \mathbb{E}[\,\lvert h(\mathbf{z}) - \bar{h}_g(\mathbf{w}) \rvert \mid g ]$. Define statistical parity ratio $R_{\mathrm{SP}} := \mathbb{P}(\hat{y}=1 \mid g=0)\, / \,\mathbb{P}(\hat{y}=1 \mid g=1)$,
, so that $\lvert R_{\mathrm{SP}} - 1 \rvert$ quantifies deviation from perfect
statistical parity (1).

\begin{figure}[hbt!]
    \centering
    \includegraphics[width=1.03\linewidth]{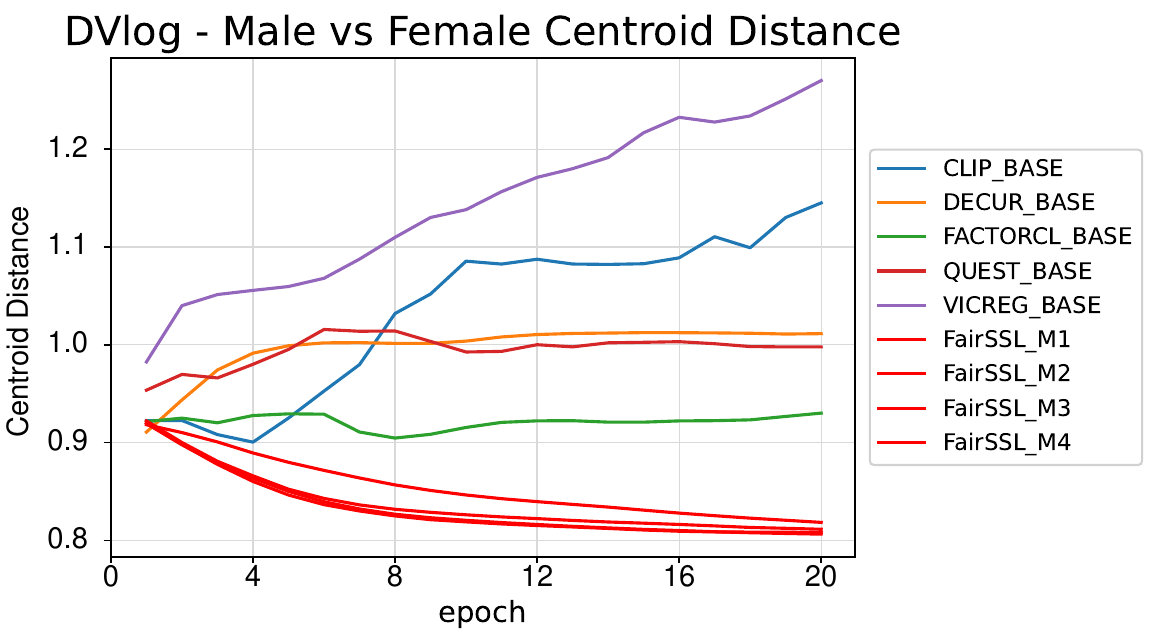} 
    \caption{Euclidean distance between the two gender centroids in a 2D PCA projection of the embeddings.}
    \label{fig:centroid_distance} 
\end{figure}

\noindent\textbf{Main Theoretical Result 1}. The statistical parity difference is bounded above by the group mean score disparity and the within-group score deviations:
\begin{equation}\footnotesize
\label{eq:main_text_SP_bound}
\bigl\lvert R_{\mathrm{SP}} - 1 \bigr\rvert
\;\le\; {\mathcal{O}\bigl(\Delta_g(\mathbf{w}) \bigr)}
\;+\;
{\mathcal{O}\bigl( v_0(\mathbf{w}) + v_1(\mathbf{w}) \bigr)}.
\end{equation}
\noindent\textbf{Main Theoretical Result 2}. Minimizing $I^{\methodsuperscript}_{\mathrm{reg}}$ lowers the first bound $\mathcal{O}\bigl(\Delta_g(\mathbf{w}) \bigr)$, which is illustrated in Fig. \ref{fig:centroid_distance}.

\noindent\textbf{Main Theoretical Result 3}. The three regularization terms bound the dispersion of the scores. $C_{reg}$ prevents the embeddings from collapsing while $I^{\methodsuperscript}_{reg}$ bounds the dispersion:
\begin{equation}\footnotesize
\Omega((c-\sqrt{C_{\rm reg}})_+)
\le
v_0(\mathbf{w})+v_1(\mathbf{w})
\le
\mathcal{O}(\sqrt{I^{\methodsuperscript}_{\mathrm{reg}}}).
\end{equation}

\noindent\textbf{Main Theoretical Result 4}. Such bounds are derived in Appendix~\ref{app:theory_guarantees} for other group fairness measures as well (Remark~\ref{rem:apply_ratio_to_metrics}).

\noindent
\textbf{Remark}. 
The theory is scoped to the standard linear-probe setting, following a common representation-learning practice in SSL to establish guarantees in the linear-probe or linear-prediction regime to isolate what the pretraining objective guarantees about the representation itself 
(e.g., \cite{saunshi2019theoretical,haochen2021provable,haochen2022beyond,johnson2022contrastive}.
%
The regularization terms jointly affect the bounds of a fairness measure and prevent the embeddings from collapsing.
Segment pooling is not separate from the theory; it is \textit{the mechanism} 
that makes the subject-aware invariance term well-defined for heterogeneous, variable-length modalities. 
%
In other words, pooling is the architectural step that enables the invariance regularization analyzed in the theory to be applied in our setting. 
%
We provide further discussions in Section \ref{sect:discussion}.
%
By contrast, alternating regularization (M4) is intended as a training-stability mechanism. 
M3 can bias the training dynamics by enforcing same-class batches, and M4 alternates M2 and M3 to address this.

\section{Experiment Setup and Details}

\subsection{Datasets}
We use the following datasets -- see Table \ref{tab:dataset_breakdown} for sample distribution and Appendix~\ref{sect:dataset_splits} for the data splits:

\begin{table}[h]
\caption{
    Dataset and target attribute breakdown across datasets. Abbreviations: F: Female. M: Male. T: Total. $Y_0$: Control group. $Y_1$: positive event group.
    \textcolor{red} {Red} highlights imbalanced splits.  \textcolor{teal}{Green} denotes relatively balanced splits.
    }
\label{tab:dataset_breakdown}
    \centering\footnotesize
    \addtolength{\tabcolsep}{-0.8mm}
    \scalebox{0.9}
    {\begin{tabular}{l|ccc|ccc|ccc|ccc} \hline
         & \multicolumn{3}{c|}{\textbf{D-Vlog}} & \multicolumn{3}{c|}{\textbf{MIMIC-CXR}} & \multicolumn{3}{c|}{\textbf{MODMA}} & \multicolumn{3}{c}{\textbf{MIMIC-III}} \\ 
         & $Y_0$ &$Y_1$ & {T} &  $Y_0$ & $Y_1$ & {T} & $Y_0$ & $Y_1$ & {T} &  $Y_0$ & $Y_1$ & {T} \\ \hline
    M  &.16 &.17 &\textcolor{red}{\textbf{.34}}
       &.20 &.29 &\textcolor{teal}{\textbf{.49}}
       &.35 &.33 &\textcolor{red}{\textbf{.68}}
       &.35 &.09 &\textcolor{teal}{\textbf{.44}}\\
    
    F  &.30 &.37 &\textcolor{red}{\textbf{.66}}
       &.18 &.34 &\textcolor{teal}{\textbf{.51}}
       &.16 &.16 &\textcolor{red}{\textbf{.32}}
       &.46 &.10 &\textcolor{teal}{\textbf{.56}}\\
    
    T  &\textcolor{teal}{\textbf{.42}} &\textcolor{teal}{\textbf{.58}} &1.00
       &\textcolor{red}{\textbf{.37}}  &\textcolor{red}{\textbf{.63}}  &1.00
       &\textcolor{teal}{\textbf{.51}} &\textcolor{teal}{\textbf{.49}} &1.00
       &\textcolor{red}{\textbf{.81}}  &\textcolor{red}{\textbf{.19}}  &1.00\\

            \hline
    \end{tabular}}
\end{table}

\noindent\textbf{D-Vlog (audio, visual)}
consists of 555 depressed and 406 non-depressed vlogs of 639 females and 322 males \cite{Yoon_Kang_Kim_Han_2022}.
The dataset owners provided a standard train-test split which we adhered to in our experiments.


\noindent\textbf{MIMIC-CXR (text, visual)} is a large-scale chest X-ray dataset containing 377,110 radiology images and free-text reports. We consider the ``No Finding'' classification task.

\noindent\textbf{MODMA (EEG, audio)}
consists of 
data from clinically depressed patients and 
healthy controls (HC) from 33 males and 20 females so females are the minority \cite{cai2022multi}. 
24 out of the 53 participants were diagnosed as depressed based on the DSM criteria. 
%


\noindent\textbf{MIMIC-III (tabular)}
contains more than 31 million clinical events that correspond to 17 clinical variables (e.g., heart rate, oxygen saturation, temperature).  
Our task involves prediction of in-hospital mortality from observations recorded within 48 hours of an intensive care unit (ICU) admission.

Variation in dataset size is intentional. 
We use these datasets to demonstrate that the method remains effective across very different regimes, including scarce and imbalanced settings such as MODMA and large-scale settings such as MIMIC-CXR. 


\subsection{Compared Methods}\label{sect:compared_methods}
We briefly list the methods and their categories here and leave their detailed descriptions to Section \ref{sect:app_compared_methods}.

\noindent\textbf{D-Vlog:}
{(i) X-add, X-concat} \cite{he2024lmvd}. 
{(ii) SEResnet} \cite{hu2018squeeze}. 
{(iii) Depression Detector (DeprDet)} \cite{Yoon_Kang_Kim_Han_2022}.  {(iv) Bi-cross, Bi-concat}. 
{(v) Perceiver} \cite{gimenogomez2024reading}.
\noindent\textbf{MIMIC-III:} 
We use SSL by splitting the tabular data columnwise into two, following \citet{yfantidou2024using}. 
\noindent\textbf{MODMA:}
{(i) MultiDepr} \cite{ahmed2023taking}.
{(ii) Effnetv2s} \cite{qayyum2023highdensity}.
{(iii) FeatNet} \cite{singh2024learning}.
{(iv) EMO-GCN} \cite{xing2024adaptive}.  {(v) EAV} 
\cite{lee2024eav}.

\noindent\textbf{Multimodal SSL:}
\underline{Contrastive:} 
{CoMM}  \cite{dufumier2025what}. 
{FOCAL} \cite{liu2023focal}. 
{QUEST} \cite{song2024quest}. 
{FACTORCL}  \cite{liang2023factorized}. 
{SimCLR}  \cite{yfantidou2024using}. 
\underline{Redundancy reduction:}
{DeCUR} \cite{wang2024decur}. 
VicReg \cite{bardesvicreg}.


\subsection{Implementation Details}

Processing and training details are in Appendix~\ref{sect:implementation_details}.

\subsection{Fairness Measures}
\label{sect:fairness_measures}

We use different fairness measures to capture different aspects of fairness: Statistical Parity (SP), Equal Opportunity (EOpp), Equalized Odds (EOdd) and Equal Accuracy (EAcc) following prior work \cite{barker2024learning,cheong_gender_fairness}. 
Specific formulations are in Appendix~\ref{sect:appendix_fairness_measures}.
For each fairness measure, the closer the value to $1$, the fairer the outcome.
In addition, 
for ease of interpretation,
following \citet{liu2025can}, we compute an aggregated form of the set $\mathcal{F}$ of fairness measures we used:
\begin{equation}\footnotesize
\label{eqn:agg_fairness}
    \textrm{AGG}_{F}= \bigg|1- \sum_{F_i \in \mathcal{F}}
    \frac{|F_i-1|}{|\mathcal{F}|}\bigg|.
\end{equation}

\section{Experiments and Results}
\label{sect:experiments}

We now evaluate our claims, corresponding to the research gaps in Introduction.


\begin{table}[hbt!]
\caption{\textbf{DVlog:} Comparison across \textbf{multimodal} supervised and self-supervised methods. *\methodname is applied on DepressionDet. \first{Bold} and \second{underline} mark best and second-best, respectively.
    }\label{tab:dvlog_multimodal}\addtolength{\tabcolsep}{-1.4mm}
  \footnotesize
  \centering
  \resizebox{0.95\linewidth}{!}{%
  \begin{tabular}{l|l|cc|cccc|c}
    \hline

      & \multirow{2}{*}{\textbf{Method}} & \multicolumn{2}{c|}{\textbf{Perf.}} &\multicolumn{5}{c}{\textbf{Fairness}}   \\

     & & Acc	&F1	& SP & EOpp  & EOdd  & EAcc &$\textrm{AGG}_{F}$ \\
   \hline
   
\multirow{9}{*}{\rot{Supervised}}%

    &  \multicolumn{8}{c}{\cellcolor{gray!10}\textbf{CNNs}}\\
    \cdashline{2-9}[.4pt/2pt]
& X-add	    &0.60		&0.66		&0.83	&1.84	&0.90	&0.89  &0.70 \\
& X-concat	&0.58		&0.65		&0.63	&1.40	&0.71	& \first{0.99}  &0.73    \\
& SEResnet	&0.57		&\second{0.72}	&0.82	&2.13	&1.04	&0.82 &0.62 \\
    \cdashline{2-9}[.4pt/2pt]

    & \multicolumn{8}{c}{\cellcolor{gray!10}\textbf{Transformers}}\\ 
     \cdashline{2-9}[.4pt/2pt]

 & Bi-cross	    &\first{0.67} &\second{0.72} &0.72	&1.60	&0.76	&0.90 &0.70 \\
 & Bi-concat	& 0.58		& 0.65		     &0.76	&1.69	& \second{1.02}	&0.74	&0.70 \\
 & Perceiver	& \second{0.62}	 &0.66		& \second{1.08}	&2.38	&1.62	&0.87	&0.45 \\ 
 & DepressionDet$^*$	& \second{0.62}		&0.69	&0.86	&1.91	&1.17	&0.79 &0.64 \\

    \hline
\multirow{12}{*}{\rot{Self-Supervised}}%

    & \multicolumn{8}{c}{\textbf{\cellcolor{gray!10} Contrastive methods}}\\
    \cdashline{2-9}[.4pt/2pt]
    & {CoMM}  & \second{0.62}	 	 &0.63	 &1.30	 &2.88	 &4.49	 &0.77 &0.47\\   
    & {FOCAL} & \second{0.62}		&0.63	&1.25	&2.77	&3.26	&0.86	&0.10\\
    & {QUEST} &\first{0.64}		&\second{0.67}	&\second{1.09}	&2.41	&2.64	&0.80	&0.17\\
    & FACTORCL & 0.61 & 0.62 & 0.95 & 2.10 & 1.64 & 0.85 & 0.52 \\ 
    & {CLIP}  &0.61		&0.60	&1.16	&2.57	&2.34	&0.93	  & 0.22\\

    \cdashline{2-9}[.4pt/2pt]
    & \multicolumn{8}{c}{\textbf{\cellcolor{gray!10} Redundancy-reduction methods}}\\
    \cdashline{2-9}[.4pt/2pt]
    
    & {DeCUR} &0.59		&0.52	&1.47	&3.25	&6.64	&0.95	  &0.06\\

    & {VICReg} {\scriptsize (baseline)$^*$} & 0.57		&0.52	&1.41	&3.13	&2.25	&0.95	&0.04\\
    
    \cdashline{2-9}[.4pt/2pt]   
    & \methodname (M1)	&0.58		&\textbf{0.73}		& \first{1.00}	&2.21	& \first{1.00} & 0.90 &0.67 \\
    & \methodname (M2) &0.59		&0.64		&0.74	& \first{1.19}	&0.89	& \first{0.99} & \first{0.86} \\
    & \methodname (M3) &0.60		&0.64	&0.84	&\second{1.38}	&\second{0.92}	&0.89 & \second{0.82} \\
    & \methodname (M4) &0.57		&0.53		&0.54	& \first{1.19}	&0.55	& \second{0.96} &0.71 \\

   \hline  
\end{tabular}
}
\end{table}


\begin{table}[hbt!]
\caption{\textbf{MODMA:} 
    Comparison across \textbf{multimodal} supervised and self-supervised methods. *\methodname is applied on FeatNet. \first{Bold} and \second{underline} mark best and second-best, respectively.\label{tab:modma_multimodal}
    }
\footnotesize
\addtolength{\tabcolsep}{-0.5mm}
  \centering
  \resizebox{1\linewidth}{!}{%
  \begin{tabular}{l|l|cc|cccc|c}
    \hline
      & \multirow{2}{*}{\textbf{Method}} & \multicolumn{2}{c|}{\textbf{Perf.}} &\multicolumn{5}{c}{\textbf{Fairness}}   \\
       &  & Acc.		&F1	& SP & EOpp  & EOdd  & EAcc  & $\textrm{AGG}_{F}$\\ 
       \cline{1-9}
     
\multirow{5}{*}{\rot{Supervised}}%
 
 & MultiDepr	&\first{0.79}		&\first{0.67}		&0.00	&0.00	&$0.00$ 	&0.76 &0.19 \\
 & Effnetv2s	&\second{0.71}	&0.50		&0.00	&0.00	&0.00	&0.88 &0.22\\
 & FeatNet	&\second{0.71}	& \first{0.67}		&0.67	&0.50	&$0.00$	&0.33 &0.38\\ 

    \cdashline{3-9}[.4pt/2pt]    
 & EMO-GCN	 &0.60	  & \first{0.67}	 &0.78	 &0.59	 &0.86	 &0.82 &0.76 \\    

    \cdashline{3-9}[.4pt/2pt]
 & EAV* &0.54	&0.57		&0.86	&0.60	&1.43	&0.62 &0.66 \\    

    \hline
\multirow{12}{*}{\rot{Self-Supervised}}%

 & \multicolumn{8}{c}{\cellcolor{gray!10}\textbf{Contrastive methods}}\\
\cdashline{2-9}[.4pt/2pt]
 & CoMM   & 0.58 & 0.14 & 0.83 & 0.63 & 4.00 & 1.11 & 0.09 \\
 & QUEST   & 0.60 & 0.32 & 0.56 & 0.42 & 2.50 & 0.88 & 0.34 \\
 & FACTORCL   & 0.63 & 0.50 & 1.21 & \second{0.91} & 2.00 & 1.39 & 0.58 \\
 & CLIP   & 0.65 & 0.37 & 0.25 & 0.33 & 0.00 & 1.45 & 0.28 \\

\cdashline{2-9}[.4pt/2pt]
 & \multicolumn{8}{c}{\cellcolor{gray!10}\textbf{Redundancy-reduction methods}}\\
\cdashline{2-9}[.4pt/2pt]
 & DeCUR   & \second{0.66} & 0.35 & 0.00 & 0.00 & 0.00 & \second{1.01} & 0.25 \\

 & VICReg {\scriptsize (baseline)$^*$} & 0.57 & \first{0.66} & 1.33 & \first{1.00} & 2.00 & 0.44 & 0.53 \\

\cdashline{2-9}[.4pt/2pt]
 & \methodname (M1) & 0.59 & 0.57 & \first{0.95} & 0.71 &\textbf{1.00} & 0.95 &\first{0.90} \\
 & \methodname (M2) &\first{0.67} & \second{0.63} &\second{0.89} &0.66 & \second{1.03} & \second{1.01} & \second{0.88}
\\
 & \methodname (M3) &0.60 &0.42 &0.67 &0.50 & \first{1.00}	& \first{1.00} &0.79
\\
 & \methodname (M4) &0.65 & \first{0.66} &0.85 &0.64 &1.08	&1.10 &0.83 
\\
\hline
\end{tabular}
}
\end{table}

\begin{table}[hbt!]
\caption{\textbf{MIMIC-CXR:} Comparison across different multimodal SSL methods \textit{across \textbf{gender}}. 
*Encoders used are ResNet-18 for the image modality and a Transformer encoder for the text modality.
    \first{Bold} and \second{underline} mark best and second-best, respectively. \label{tab:MIMIC_CXR_ssl_accuracy}
    }
\footnotesize
\addtolength{\tabcolsep}{-0.5mm}
  \centering\resizebox{1\linewidth}{!}{%
  \begin{tabular}{l|cc|cccc|c}
    \hline
       \multirow{2}{*}{\textbf{Method}} & \multicolumn{2}{c|}{\textbf{Perf.}} &\multicolumn{5}{c}{\textbf{Fairness}}   \\
       & Acc. & F1 & SP & EOpp & EOdd & EAcc &$\textrm{AGG}_{F}$\\ \hline

    \rowcolor{gray!10} \multicolumn{8}{c}{\textbf{Contrastive methods}}\\
    \hline
CoMM     &\first{0.74} & \second{0.79} &1.04 &\second{1.04} &1.03 &\second{1.04} &0.96\\
QUEST    &0.69 &0.74 &0.96 &0.95 &0.96 &\first{1.01} &0.96\\
FACTORCL &0.69 &0.77 &0.98 &\first{0.98} &0.97 &\second{1.04} &\second{0.97}\\
CLIP     &\first{0.74} & \second{0.79} &1.05 &1.05 &1.03 &\second{1.04} &0.96\\

    \hline
    \rowcolor{gray!10} \multicolumn{8}{c}{\textbf{Redundancy-reduction methods}}\\
    \hline
DeCUR    &0.72 &0.78 &1.05 &1.05 &1.04 &1.06 &0.95\\ 
VICReg {\scriptsize (baseline)$^*$} &\first{0.74} & \second{0.79} &1.08 &1.07 &1.05 &1.05 &0.94\\
\cdashline{1-8}[.4pt/2pt]
\methodname (M1) &0.66 &0.76 &\first{1.01} &1.07 & \second{1.01} &1.06 &0.96\\ 
\methodname (M2) &0.69 & \second{0.79} &\first{0.99} &\second{1.04} &0.98 &1.06 &\second{0.97}\\
\methodname (M3) &\second{0.73} &\first{0.81} & \second{1.02} &1.05 &\textbf{1.00} &\second{1.04} &\first{0.98}\\ 
\methodname (M4) &0.70 & \second{0.79} &\second{1.02} &\second{1.04} &\second{1.01} &1.05 &\second{0.97}\\

   \hline
    \end{tabular}
}
\end{table}


\subsection{Multimodal vs. Multimodal SSL (RG 1)}

For D-Vlog in Table \ref{tab:dvlog_multimodal}, 
we see that although multimodal methods typically perform better than unimodal methods, this does not always hold for multimodal SSL methods. 
D-Vlog is balanced across class but not gender (Table \ref{tab:dataset_breakdown}).
For D-Vlog,
SOTA SSL methods provide comparable performance but generally perform poorer across fairness compared to the multimodal supervised methods, suggesting that, in class-balanced settings,
factors that lead to improved performance do not necessarily lead to improved fairness.

For MODMA in Table \ref{tab:modma_multimodal}, we see an interesting trend where
supervised
multimodal methods (e.g. MultiDepr, Effnetv2s) provide better classification performance compared to existing SSL methods. However, they perform poorly on fairness; e.g.,
although the strongest baseline (MultiDepr) reaches the highest accuracy and F1, its fairness completely collapses, a pattern repeated by Effnetv2s and FeatNet. EMO-GCN and the Transformer baseline improve stability in $\textrm{AGG}_F$, but still underperform in maintaining consistent fairness. 
For MODMA, 
SOTA SSL methods generally lead to a reduction in both performance and fairness. 
We hypothesize that this is due to the severe class imbalance (Table \ref{tab:dataset_breakdown}).

\setlength\FrameSep{\fboxsep}
\begin{framed}
\noindent\textbf{Takeaway (RG1):}
%
Multimodal methods in general improve task performance, likely due to the richer representation learnt. 
Under class-balanced settings (i.e. D-Vlog), SSL methods provide comparable performance but generally perform poorer across fairness compared to the multimodal non-SSL methods, suggesting that in class-balanced settings
factors that lead to improved performance do not necessarily lead to improved fairness.
Under scarce data and gender-imbalanced settings (i.e. MODMA), 
supervised multimodal methods are better than SSL methods across performance but inferior in terms of fairness. 
\end{framed}


\begin{table}[hbt!]
\caption{\textbf{MIMIC-III:} Comparison across different  multimodal SSL methods. *VICReg is applied on the same architecture as SimCLR.
    \first{Bold} and \second{underline} mark best and second-best, respectively. \label{tab:MIMIC_ssl_accuracy}
    }
\footnotesize
\addtolength{\tabcolsep}{-1mm}
  \centering
  \resizebox{0.95\linewidth}{!}{%
  \begin{tabular}{l|cc|cccc|c}
    \hline
       \multirow{2}{*}{\textbf{Method}} & \multicolumn{2}{c|}{\textbf{Perf.}} &\multicolumn{5}{c}{\textbf{Fairness}}   \\
     
       & Acc.	&F1	   & SP & EOpp  & EOdd  & EAcc  &$\textrm{AGG}_{F}$\\ \hline

    \rowcolor{gray!10} \multicolumn{8}{c}{\textbf{Contrastive methods}}\\
    \hline
CoMM	&\second{0.84}	&0.12		&0.84	&0.79	&0.86	&1.03  &0.86\\
FOCAL	&0.82	&0.15	&1.67	&1.57	&2.04	&0.92 &0.41 \\ 
SimCLR	&0.74	&0.18 &0.92	&0.86	& \first{1.01}	&0.97 &\second{0.94} \\
FACTORCL	&0.81	&0.16	&0.78	&0.67	&0.54	&1.04 &0.74 \\ 
CLIP	&\second{0.84}	&0.14 &1.45	&1.36	& 2.14	&0.94 & 0.50 \\

    \hline
    \rowcolor{gray!10} \multicolumn{8}{c}{\textbf{Redundancy-reduction methods}}\\
    \hline
DeCUR	&0.68	&0.20	&1.36	&1.28	&1.53	&0.83 &0.67 \\ 
 
VICReg {\scriptsize (baseline)$^*$} &0.81&0.11&0.93&\second{0.93}&1.30&\second{0.98}&0.88 \\
\cdashline{1-8}[.4pt/2pt]
\methodname (M1)	&0.82	&0.17	& \second{1.06}	& \first{1.01}	&1.21	&0.96	& {0.92} \\
\methodname (M2) & 0.77 & \first{0.27} & \first{1.05} & \first{0.99} & 1.10 & 0.95 & \first{0.95}\\
\methodname (M3) &0.77	&0.21	&0.84	&0.89	&0.89	& \first{1.00} &0.90 \\
\methodname (M4) &\first{0.86}	& \second{0.26}	&0.89	&0.84	& \second{1.07}	&\second{0.98}  &0.91 \\
 
   \hline  
    \end{tabular} 
}
\end{table}


\subsection{Multimodal SSL characteristics for fairness under heterogeneity (RG2)}

\begin{figure}
    \centering
    \includegraphics[width=1\linewidth]{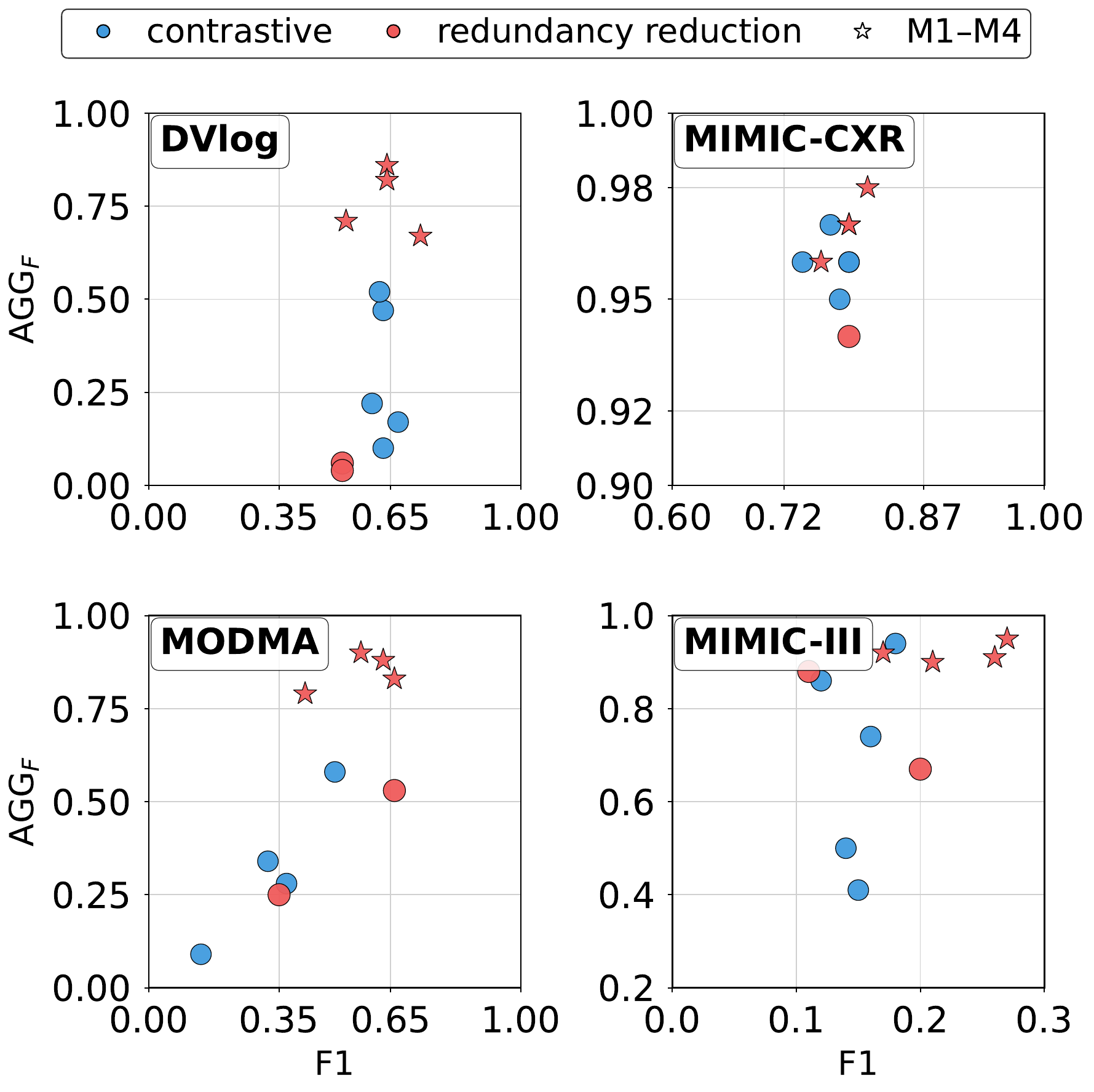} 
    \caption{Comparison of contrastive and redundancy-reduction SSL methods with respect to F1 and AGG$_F$ scores.}
    \label{fig:horizontal_f1plot} 
    \vspace{-2mm}
\end{figure}

\textbf{Effect of contrastive learning.}
Contrastive learning typically attempts to pull together features of different modalities for the same class and push away those for the different class. 
Without appropriate mitigation, such strategies may end up optimizing for class performance at the expense of fairness. We see traces of this in Tables \ref{tab:dvlog_multimodal},  \ref{tab:modma_multimodal}, \ref{tab:MIMIC_ssl_accuracy}, and  \ref{tab:ablation_dvlog_ssl}, which are summarized in Fig. \ref{fig:horizontal_f1plot}. 
%

{To broaden the evaluation, we also computed 
fairness results across age and race for MIMIC-CXR (see Tables \ref{tab:MIMIC_CXR_age} and \ref{tab:MIMIC_CXR_race}).
Across these 
protected attributes, we see that the gains are strongest on SP, EOpp, and EOdd, the fairness measures for which our theory proves direct bounds for (Section~\ref{sect:theoretical_link} and Appendix Remark~\ref{rem:apply_ratio_to_metrics}), while EAcc is slightly worse in some cases, which we view as the known fairness measure trade-off rather than a contradiction of our theory.
%

\begin{table}[hbt!]

\caption{\textbf{MIMIC-CXR:} Comparison across different multimodal SSL methods \textit{across \textbf{age}}. 
*Encoders used are ResNet-18 for the image modality and a Transformer encoder for the text modality.
    \first{Bold} and \second{underline} mark best and second-best, respectively. \label{tab:MIMIC_CXR_age}
    }
\footnotesize
\addtolength{\tabcolsep}{-0.5mm}
  \centering\resizebox{1\linewidth}{!}{%
  \begin{tabular}{l|cccc|c}
    \hline
       \multirow{2}{*}{\textbf{Method}}&\multicolumn{5}{c}{\textbf{Fairness}}   \\
      & SP & EOpp & EOdd & EAcc &$\textrm{AGG}_{F}$\\ 
       \hline

    \hline
CoMM   &\first{1.07} & 1.32 & 1.77 & 1.56 & 0.72 \\
QUEST  & 1.09 & 1.27& 1.40& 1.50   & 0.77  \\
DeCUR   & \second{1.08}& 1.31& 1.71& 1.54 &0.73  \\ 
CLIP  &\second{1.08} & 1.39& 1.80& 1.51&0.72   \\

    \hline

VICReg {\scriptsize (baseline)$^*$}  & \first{1.07} & 1.32& 1.74 & 1.56   & 0.73 \\
\cdashline{1-6}[.4pt/2pt]
\methodname (M1)    & 1.43 & \second{0.85} & \first{1.01}  & \first{1.02}  &\first{0.96} \\ 
\methodname (M2) & 1.21 & \first{1.10} & \second{1.18} & \second{1.20}& \second{0.85}\\
\methodname (M3) & 1.11& 1.33 & 1.49 & 1.45& 0.75 \\ 
\methodname (M4)   & 1.18 & \second{1.15}  & 1.24 & 1.25& 0.83 \\

   \hline
    \end{tabular}
}
\end{table}


\begin{table}[hbt!]

\caption{\textbf{MIMIC-CXR:} Comparison across different multimodal SSL methods \textit{across \textbf{race}}. 
*Encoders used are ResNet-18 for the image modality and a Transformer encoder for the text modality.
    \first{Bold} and \second{underline} mark best and second-best, respectively. \label{tab:MIMIC_CXR_race}
    }
\footnotesize
\addtolength{\tabcolsep}{-0.5mm}
  \centering\resizebox{0.95\linewidth}{!}{%
  \begin{tabular}{l|cccc|c}
    \hline
       \multirow{2}{*}{\textbf{Method}}&\multicolumn{5}{c}{\textbf{Fairness}}   \\
      & SP & EOpp & EOdd & EAcc &$\textrm{AGG}_{F}$\\ 
       \hline

    \hline
CoMM   & 0.92 & 0.25 & 1.05 & \second{0.96} & 0.77 \\
QUEST  & 0.94 & 0.24 & 1.03 & \first{0.97} & \second{0.78} \\
DeCUR  & 0.91 & 0.25 & 1.02 & \second{0.96} & \second{0.78} \\
CLIP   & 0.90 & \second{0.26} & \second{0.99} & \second{0.96} & \second{0.78} \\

    \hline

VICReg {\scriptsize (baseline)$^*$}  & 0.91 & 0.25 & \first{1.00} & \first{0.97} & \second{0.78} \\
\cdashline{1-6}[.4pt/2pt]
\methodname (M1) & 0.95 & \second{0.26} & \first{1.00} & 0.95 & \first{0.79} \\ 
\methodname (M2) & \second{0.97} & 0.25 & \first{1.00} & 0.93 & \first{0.79} \\
\methodname (M3) & 0.95 & \first{0.27} & \second{1.01} & 0.94 & \first{0.79} \\ 
\methodname (M4) & \first{0.99} & 0.25 & 1.02 & 0.90 & \second{0.78} \\

   \hline
    \end{tabular}
}
\end{table}


\textbf{Effect of redundancy reduction.}
Redundancy reduction methods typically attempt to decompose feature embeddings into cross-modal shared components and modality-specific components, encouraging alignment of shared representations across modalities while pushing modality-unique representations apart.
Such a strategy will be effective for datasets that contain high-level of modality specific unique information. We see evidence for this in Tables \ref{tab:dvlog_multimodal},  \ref{tab:modma_multimodal}, \ref{tab:MIMIC_ssl_accuracy}, and  \ref{tab:ablation_dvlog_ssl}, which are summarized in Fig. \ref{fig:horizontal_f1plot}. MIMIC-CXR results in Table \ref{tab:MIMIC_CXR_ssl_accuracy} suggest that redundancy-based approaches can outperform even in a scenario where contrastive approaches provide a good performance-fairness balance.
%


\newcommand{\specialcell}[2][c]{%
  \begin{tabular}[#1]{@{}c@{}}#2\end{tabular}}
  
\begin{table}[hbt!]
\caption{\textbf{D-Vlog:}
    Ablation analysis on pooling. 
    Comparisons include (i) ``No pooling'': no pooling on $\{\mathbf{z}^{m}_{i,j}\}$ for $m=m_1$ or $m_2$, 
    i.e. the baseline VICReg method, (ii) ``Single Pooling'': pooling only on $\{\mathbf{z}^{m_1}_{i,j}\}$ for $m_1$ (as explained in Section \ref{sect:method}) and (iii) ``Double Pooling'':  pooling on both $\{\mathbf{z}^{m_1}_{i,j}\}$ and $\{\mathbf{z}^{m_2}_{i,j}\}$. 
    \first{Bold} and \second{underline} mark best and second-best, respectively. \label{tab:dvlog_pooling}
    }
\footnotesize
\addtolength{\tabcolsep}{-0.5mm}
  \centering\resizebox{0.95\linewidth}{!}{
  \begin{tabular}{l|c|cc|cccc|c}
    \hline
       \multicolumn{2}{c|}{}  & \multicolumn{2}{c|}{\textbf{Performance}} &\multicolumn{5}{c}{\textbf{Fairness}}   \\
     
    \multicolumn{2}{c|}{\textbf{Method}}   & Acc.		&F1   & SP & EOpp  & EOdd  & EAcc &$\textrm{AGG}_{F}$\\ \hline 
\multicolumn{2}{c|}{\textbf{No Pooling}} & \multirow{2}{*}{0.57}		& \multirow{2}{*}{0.52}	& \multirow{2}{*}{1.41}	& \multirow{2}{*}{3.13}	& \multirow{2}{*}{2.25}	& \multirow{2}{*}{0.95}	& \multirow{2}{*}{0.04}\\
\multicolumn{2}{c|}{(baseline)} & 		&	&	&	&	&	& \\
 \hline   
 \multirow{4}{*}{\rot{\specialcell{\bf{Single} \\ Pooling}}}
& M1	&0.58		&\second{0.73}		& \first{1.00}	&2.21	& \first{1.00}	&0.90 &0.67 \\
& M2 &0.59		&0.64		&0.74	& \first{1.19}	&0.89	& \first{0.99} &\first{0.86} \\
& M3	&0.60		&0.64	&0.84	& \second{1.38}	&0.92	&0.89 &\second{0.82} \\
& M4	&0.57		&0.53		&0.54	& \first{1.19}	&0.55	&0.96 &0.71 \\
\hline
\multirow{4}{*}{\rot{\specialcell{\bf Double \\ Pooling}}}
& M1 & \second{0.62} & \second{0.73} & 1.06 & 2.35 & 1.61 & 0.82 & 0.45  \\ 
& M2 & 0.61  & 0.72 & \second{0.99} & 2.09 & 1.13 & 0.84 & 0.66  \\ 
& M3 & \first{0.65}  & \textbf{0.78} & 0.95 & 2.10 & \first{1.00} & \second{0.98} & 0.71  \\ 
& M4 & 0.53  & 0.63 & 0.93 & 2.05 & \second{1.04} & 0.75 & 0.65  \\

    \cdashline{1-9}[.4pt/2pt]
    
   \hline  
    \end{tabular}
}
\end{table}


\textbf{Effect of pooling.}
Subject-aware pooling seems well-suited as a temporal alignment technique to enhance uniqueness, reduce redundancy and improve fairness in SSL settings.
Looking at Tables \ref{tab:dvlog_multimodal} and \ref{tab:dvlog_pooling}, 
both single and double-pooling improve upon the baseline model 
with no pooling,
thus implying that the model has learned fairer 
representations via the pooling mechanism.
However, double‑pooling seems to slightly
underperform across fairness compared to single‑pooling. 
We hypothesize that this because double-pooling may have 
overly reduced redundancy to the point of loss in unique information learnt
and hence induced a slight reduction the modality specific gender de-correlation.

%
%
With reference to Table \ref{tab:ablation_dvlog_ssl} we see that our subject-aware cross-modal
alignment is effective at promoting subject-dependent inter-modality synergy across both contrastive and redundancy reduction based SSL methods.
For DVlog, 
given that males and females tend to exhibit different behavioural cues when depressed  \cite{cheong_gender_fairness}, 
\methodname-M1 (intra-subject reg.)
and M2 (inter-subject reg.)
should give the best outcome as both methods encourage the model to learn more relevant \textbf{\textit{intra-}} and \textbf{\textit{inter-}} subject representations that are indicative of depression for each \textit{individual} of different \textit{gender}. 
This hypothesis is well-supported by our results in Table \ref{tab:dvlog_multimodal}.
This also true for MODMA in Table \ref{tab:modma_multimodal}, where we see \methodname-M1 and M2 giving the top two performance and fairness outcomes.
MIMIC-III, on the other hand, may suffer less from modality alignment issues 
and may have less 
{\textit{intra-}} and {\textit{inter-}} subject differences as it is simply a tabular data split into two separate parts. As a result, with reference to Table \ref{tab:MIMIC_ssl_accuracy}, although \methodname still gives improved performance compared to baseline, the improvements are minimal compared to DVlog and MODMA.


\begin{framed}
\noindent\textbf{Takeaway (RG2):}
Baseline contrastive and redundancy-based approaches achieve competitive overall performance but perform poorly with respect to fairness. These limitations can be mitigated through subject-based pooling and subject-aware cross-modal alignment. 
\end{framed}


\subsection{Fair multimodal SSL for heterogeneous data (RG3)}

Across Tables
\ref{tab:dvlog_multimodal}--\ref{tab:ablation_dvlog_ssl},
we see that variations of \methodname on existing SOTA SSL methods typically improve on performance and fairness across all datasets.
This is supported by Fig. \ref{fig:pareto_plot} where we see 
variations of \methodname, 
which contains 
pooling-based temporal alignment to enhance \textbf{gender-specific} intra-modality uniqueness and reduce inter-modality redundancy, and a subject-aware regularization alignment strategy to promote 
\textbf{gender-specific} inter-modality synergy, 
consistently producing the best outcomes across the performance-fairness Pareto frontier.

\begin{framed}
\noindent\textbf{Takeaway (RG3):}
Across settings, \methodname dominate competing SSL methods in terms of fairness, without performance degradation.
These demonstrate that it is possible to move beyond the conventional performance–fairness Pareto frontier by leveraging heterogeneous data through pooling mechanisms and (ii) subject-aware regularization.
%
\end{framed}

\begin{table}[hbt!]
\caption{\textbf{D-Vlog:} 
    Ablation analysis on \methodname methods applied on other \textbf{SSL} multimodal methods. 
    Best and second-best results are noted in \first{bold} and \second{underline}
    for each method separately. \label{tab:ablation_dvlog_ssl}
    }  
\footnotesize
\addtolength{\tabcolsep}{-0.8mm}
  \centering
  \begin{tabular}{l|cc|cccc|c}
    \hline
    \multirow{2}{*}{\textbf{Method}} & \multicolumn{2}{c|}{\textbf{Perf.}} &\multicolumn{5}{c}{\textbf{Fairness}}   \\
     
        & Acc.		&F1   & SP & EOpp  & EOdd  & EAcc  &$\textrm{AGG}_{F}$\\ \hline

    \rowcolor{gray!10} \multicolumn{8}{c}{\textbf{Contrastive methods}}\\
    \hline
     
 \textbf{CoMM} & \second{0.62}	 	 & \second{0.63}	 &1.30	 &2.88	 &4.49	 &0.77 &0.47\\
 \cdashline{2-8}[.4pt/2pt]   
 w M1	 & \first{0.63}		 & \first{0.65}	 & \first{1.16} &2.57	 &2.11	 &0.86 &0.25\\
 w M2	 &0.57	 	 &0.31	 &0.44	 & \first{0.33}	 & \first{1.00}	 & \first{0.95}  & \second{0.68}\\
 w M3	 & \second{0.62}     & \first{0.65}	 &0.84	 & \second{1.86}	 & \first{1.00}	 & \second{0.89} & \first{0.72}\\
 w M4	 & \second{0.62}     & \first{0.65}	 & \second{0.97}	 &2.14	 & \second{1.51}	 &0.84 &0.54\\

\hline
 \textbf{FOCAL} & \first{0.62}		&0.63	&1.25	&2.77	&3.26	&0.86	&0.10\\
 \cdashline{2-8}[.4pt/2pt]   
 w/ M1	 & \first{0.62}		& \second{0.64}	&1.17	&2.64	&2.64	&0.84	&0.10\\
 w/ M2	 &0.52		&0.38	& \second{1.04}	& \second{0.75}	& \first{1.00}	&\second{1.11}	& \first{0.90}\\
 w/ M3	& \second{0.60}		&0.37	&1.22	& \first{0.91}	&1.38	&1.17	& \second{0.79}\\
 w/ M4	 &0.59		& \first{0.69}	& \first{0.99}	&2.11	& \second{1.12}	&\first{0.93}	&0.67\\

\hline
 \textbf{QUEST} &\first{0.64}		&0.67	&1.09	&2.41	&2.64	&0.80	&0.17\\
 \cdashline{2-8}[.4pt/2pt]   
 w/ M1	 & \second{0.62}		&0.43	&0.89	& \first{0.67}	& \first{1.00}	& \second{1.14}	&\first{0.85}\\
 w/ M2	 &0.58		& {0.67}	& \first{0.97}	&2.15	& \second{1.01}	& \second{0.86}	&0.33\\
 w/ M3	&0.56		&\first{0.69}	&0.88	& \second{1.85}	&0.89	&\first{0.87}	& \second{0.70}\\
 
 w/ M4	&\second{0.62}		&\second{0.68}	& \second{1.08}	&2.39	&1.51	&0.82  &0.46\\

\hline
\rowcolor{gray!10}\multicolumn{8}{c}{\textbf{Redundancy-reduction methods}}\\
\hline
 \textbf{DeCUR} &0.59		&0.52	&1.47	&3.25	&6.64	& \first{0.95}	 
 &0.06\\
 \cdashline{2-8}[.4pt/2pt]   
 w/ M1	 & \second{0.63}		 &\first{0.70}	 & \first{1.10} &2.44	 &1.69	 &0.81	 &0.39 \\
 w/ M2	 &0.62	  &\second{0.66}	 & \second{0.86}	 & \second{1.90}	 &1.22	 &0.81	 &0.64 \\
 w/ M3  &\textbf{0.65}		 &0.56	 &0.67	 & \first{0.50}	 & \first{1.00}	 &0.83	 & \first{0.75}	\\
 w/ M4	  &0.61		 &0.63	 & \first{0.90}	 &2.00	 & \second{0.91}	 & \second{0.88}	 & \second{0.67} \\ \hline

   \end{tabular} 
\end{table}

\begin{figure}
    \centering
    \begin{tikzpicture}

\node (image) at (0,0) {	\includegraphics[width=0.85\linewidth]{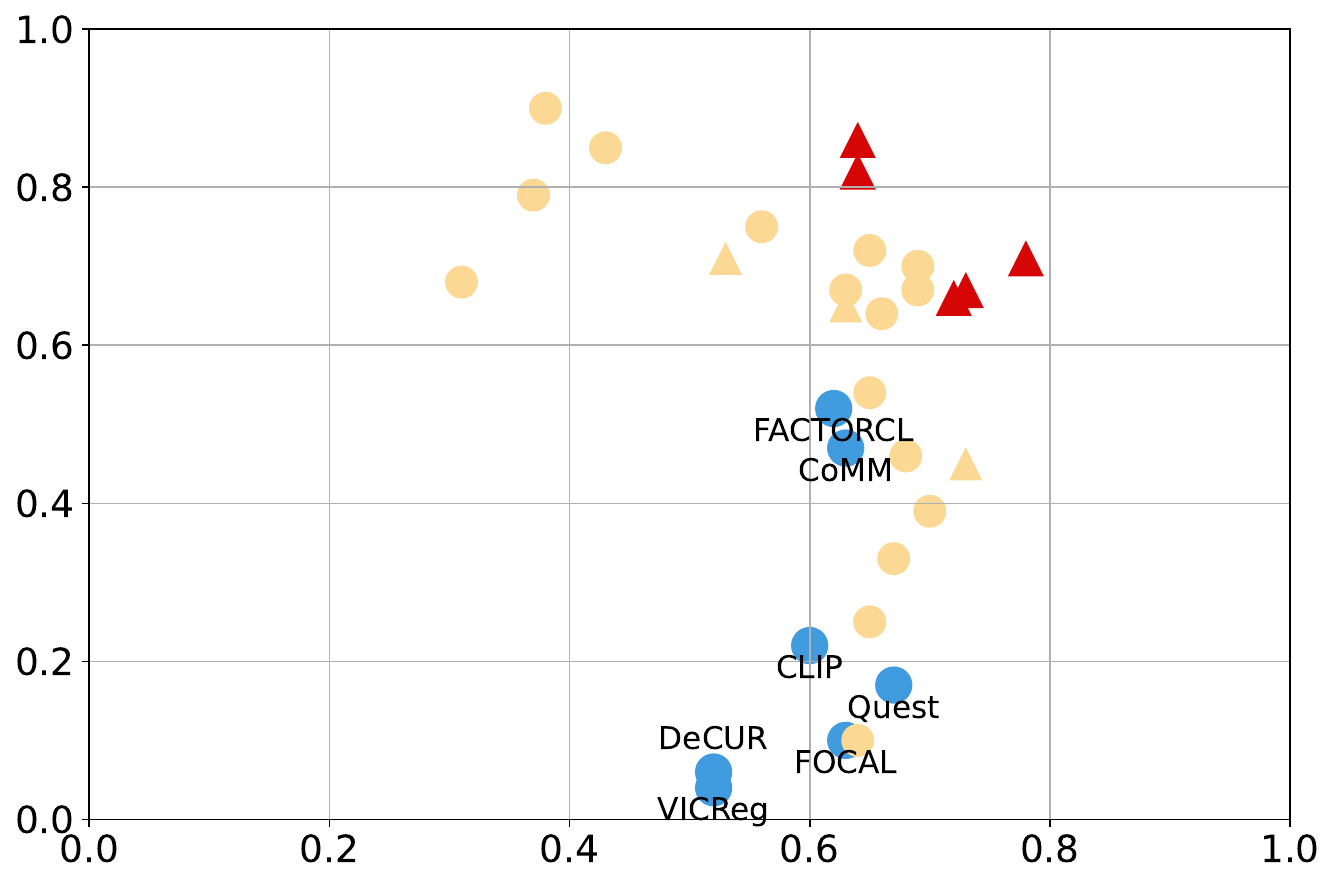}};
{ \node [below] at (0.1,-2.0) {$F1$}; }
 { \node [left,rotate=90] at (-3.5, 0.7) {$\textrm{AGG}_F$}; }
{ \node at (2.9, 2.2) {\textcolor{red}{\footnotesize Optimum}}; }
\draw [red,-stealth](2.5,1.7) -- (3.08,2.03);
\draw[red,dashed,-]  (0.9, 2.0) to[bend left=30] (2.4, 0.5);
{ \node at (1.2, 2.1) {\tiny \textcolor{red}{Pareto Front}}; }
\begin{scope}[
x={($0.1*(image.south east)$)},
y={($0.1*(image.north west)$)}]

\end{scope}

\end{tikzpicture}
    \vspace{-1mm}
    \caption{$\textrm{AGG}_{F}$ vs. F1-score Pareto Plot for DVlog. 
    \textcolor{Mahogany}{\textbf{Red}} 
    triangles represents best results from \methodname.
    \textcolor{MidnightBlue}{\textbf{Blue}} circles represent baseline SSL methods. 
    \textcolor{Dandelion}{\textbf{Yellow}} circles represent the SSL-methods with our \methodname modifications.}
    \label{fig:pareto_plot}
    \vspace{-2mm}
\end{figure}


\section{Discussion and Conclusion}
\label{sect:discussion}

Our method provides a heterogeneity-aware alignment strategy; empirically, it improves fairness while maintaining competitive downstream performance across datasets. The variance- and invariance-based regularization encourage representations that are more reflective of the prediction task and less reliant on the protected attribute.
We further distill the following key principles (\textbf{KP}s) for future efforts to develop fair multimodal SSL methods.

\textbf{KP 1 (RG1):}  
Multimodal SSL does not necessarily improve fairness relative to supervised multimodal learning. When objectives prioritize shared information and alignment, models may still exploit demographic-specific shortcuts.

\textbf{KP 2 (RG2):} 
Intra-modal representations can complement cross-modal representations, especially when modalities encode different degrees of modality-unique information for different demographic subgroups.

\textbf{KP 3 (RG2-3):} Alignment strategies can have a huge impact on fairness; if alignment discards subgroup-relevant signals, fairness can degrade even when predictive performance improves.

\textbf{KP 4 (RG2):} The distribution of sensitive-attribute-dependent signals across modalities matters.
In our experiments, D-Vlog suggests that both modalities contain gender-dependent signals, and optimizing solely for task performance can yield modest performance gains at the expense of fairness.
MIMIC-III exhibits high modality uniqueness, and contrastive and redundancy-based methods show broadly similar behavior.
For MODMA, EEG appears largely gender-invariant, whereas audio exhibits mild gender dependence; the strongest results under M1 suggest that the relevant signals are primarily modality-specific rather than shared across modalities.

\textbf{KP 5 (RG3):} Pooling can amplify subject-level signals within each modality, supporting alignment at the subject level. In our experiments, intra-subject (M1) and inter-subject (M2) variants often provide strong performance--fairness trade-offs.
%
%

\textbf{Practical Guidelines:} 
In a new multimodal task, the key data features for choosing between M1 and M4 are:
(i) how much of the useful signal is shared across modalities within the same subject,
(ii) how similar samples from the same class across different subjects are, and
(iii) how large the subject-specific heterogeneity is.
%
%
%
\textbf{M1} is the better choice when the task is driven mainly by \textit{within-subject cross-modal consistency}, while subject-specific differences are large and same-class samples across subjects are not expected to align tightly.
E.g., depression classification from speech and facial expressions naturally favors M1.
\textbf{M4} becomes more appropriate when there is clear \textit{class-level shared structure across subjects}, so cross-subject alignment is useful, but enforcing 
class-aware constraint throughout training may be too rigid. 
%
%
E.g., pneumonia classification from chest X-rays and Electronic Health Record tabular variables can better motivate M4 because same-class patients share cross-subject structure but remain heterogeneous.

\noindent\textbf{Limitations:} 

Our method is broadly applicable across a range of settings. However, in high-stakes domains such as healthcare, we emphasize that any resulting models must undergo rigorous clinical validation prior to deployment. Without such validation, there is a risk of unintended bias, performance trade-offs across subgroups, or misinterpretation of predictions in clinical workflows. In addition, models may exhibit subgroup-specific failures under distribution shift, particularly when deployed in new clinical environments with differing patient populations or data collection processes. We also note that multimodal healthcare data introduces additional privacy considerations, as combining heterogeneous data sources may increase the risk of re-identification or unintended information leakage. Furthermore, fairness improvements observed on the evaluated datasets may not generalize across institutions or demographic settings, especially in the presence of dataset shift or differing population characteristics. Accordingly, these methods should be deployed with appropriate human oversight, robustness and external validation across diverse settings, and alignment with established clinical, ethical, and regulatory standards.


%

%




\clearpage

\section*{Impact Statement}
We investigate the prevalent, and yet, understudied problem of learning fairer representations
from multiple sources of heterogeneous data 
of varying temporality and feature length,
which is common for healthcare data collected from real-world settings.
We also address the timely need of developing 
fairer ML methods that can work with minimal supervised data. 
We show that SSL-based methods can be highly effective if guided with appropriate strategies.
\methodname is effective across all three datasets of different data types and modalities in our experiments.

\section*{Acknowledgement}

Abtin Mogharabin is supported by the Konrad Zuse School of Excellence in Learning and Intelligent Systems (\href{https://eliza.school/}{ELIZA}) through the DAAD programme Konrad Zuse Schools of Excellence in Artificial Intelligence, sponsored by the Federal Ministry of Education and Research.
Jiaee Cheong is supported by a DAAD Fellowship. 
We also gratefully acknowledge the computational resources kindly provided by METU-ROMER, Center for Robotics and Artificial Intelligence.

\bibliographystyle{icml2026}
\bibliography{icml}


\appendix
\onecolumn

\section*{Appendices} 

\renewcommand\thefigure{\thesection.\arabic{figure}}   
\renewcommand\thetable{\thesection.\arabic{table}}
\renewcommand\theequation{\thesection.\arabic{equation}}

\noindent{\bfseries{Contents}\hfill Page No. \par }
{\setstretch{0.2}
\contentsline {section}{\numberline {A.} More on the Experimental Setup}{\pageref{sect:more_setup}}{}%
\contentsline {subsection}{\numberline {A.1.} Dataset Splits}{\pageref{sect:dataset_splits}}{}%
\contentsline {subsubsection}{\numberline {A.1.1.} D-Vlog}{\pageref{app:dvlog_splits_note}}{}%
\contentsline {subsubsection}{\numberline {A.1.2.} MIMIC-III}{\pageref{app:mimic_splits_note}}{}%
\contentsline {subsubsection}{\numberline {A.1.3.} MODMA}{\pageref{app:modma_splits_note}}{}%
\contentsline {subsubsection}{\numberline {A.1.4.} MIMIC-CXR}{\pageref{app:mimiccxr_splits_note}}{}%
\contentsline {subsection}{\numberline {A.2.} Implementation Details}{\pageref{sect:implementation_details}}{}%
\contentsline {subsubsection}{\numberline {A.2.1.} D-Vlog}{\pageref{app:impl_dvlog}}{}%
\contentsline {subsubsection}{\numberline {A.2.2.} MIMIC-III}{\pageref{app:impl_mimic}}{}%
\contentsline {subsubsection}{\numberline {A.2.3.} MODMA}{\pageref{app:impl_modma}}{}%
\contentsline {subsubsection}{\numberline {A.2.4.} MIMIC-CXR}{\pageref{app:impl_mimiccxr}}{}%
\contentsline {subsection}{\numberline {A.3.} Details of the Compared Methods}{\pageref{sect:app_compared_methods}}{}%
\contentsline {subsubsection}{\numberline {A.3.1.} D-Vlog}{\pageref{app:methods_dvlog}}{}%
\contentsline {subsubsection}{\numberline {A.3.2.} MODMA}{\pageref{app:methods_modma}}{}%
\contentsline {subsubsection}{\numberline {A.3.3.} MIMIC-III}{\pageref{app:methods_mimic}}{}%
\contentsline {subsection}{\numberline {A.4.} Fairness Measure Definitions}{\pageref{sect:appendix_fairness_measures}}{}%

\contentsline {section}{\numberline {B.} Theoretical Validation of \methodname}{\pageref{sect:theory}}{}%
\contentsline {subsection}{\numberline {B.1} Notation}{\pageref{app:theory_notation}}{}%
\contentsline {subsection}{\numberline {B.2.} Theoretical Guarantees}{\pageref{app:theory_guarantees}}{}%
\contentsline {subsection}{\numberline {B.3.} Proofs}{\pageref{app:theory_proofs}}{}%

\contentsline {section}{\numberline {C.} Additional Results}{\pageref{sect:additional_results}}{}%
\contentsline {subsection}{\numberline {C.1.} DVlog}{\pageref{sect:add_results_dvlog}}{}%
\contentsline {subsection}{\numberline {C.2.} MIMIC-III}{\pageref{sect:add_results_mimic}}{}%
\contentsline {subsection}{\numberline {C.3.} MODMA}{\pageref{sect:add_results_modma}}{}%

\contentsline {section}{\numberline {D.} Hyperparameters}{\pageref{sect:hyperparameters}}{}%

\contentsline {section}{\numberline {E.} Extended Related Work}{\pageref{app:extended_related_work}}{}%
\contentsline {subsection}{\numberline {E.1.} Self-supervised Learning}{\pageref{app:ssl_related_work}}{}%
\contentsline {subsection}{\numberline {E.2.} Multimodal Fairness}{\pageref{app:multimodal_ssl_related_work}}{}%
\contentsline {subsection}{\numberline {E.3.} Self-Supervised Learning for Fairness}{\pageref{app:ssl_fairness_related_work}}{}%
\contentsline {subsection}{\numberline {E.4.} Fairness in Healthcare}{\pageref{app:fair_health_related_work}}{}%
}

\section{More on the Experimental Setup}
\label{sect:more_setup}

\subsection{Dataset Splits}
\label{sect:dataset_splits}

\begin{table}[ht]
    \caption{Dataset split proportions for \textbf{D-Vlog}. Abbreviations: $Y_0$: Healthy Controls. $Y_1$: Major Depressive Disorder. M: male. F: female.}
    \centering
    \addtolength{\tabcolsep}{-1.0mm}
    \begin{tabular}{l|ccc|ccc|ccc}\hline
        & \multicolumn{3}{c|}{\textbf{Train}} 
        & \multicolumn{3}{c|}{\textbf{Val}} 
        & \multicolumn{3}{c}{\textbf{Test}} \\ 
        & $Y_0$ & $Y_1$ &T &$Y_0$ & $Y_1$ &T &$Y_0$ & $Y_1$ &T  \\ 
        \hline
        M   & 0.15 & 0.19 &0.34 & 0.19 & 0.21 & 0.40 & 0.19 & 0.12 &0.31 \\
        F & 0.27 & 0.39 & 0.66 & 0.25 & 0.35 & 0.60 & 0.39 & 0.30 &0.69 \\
        T &0.42 &0.58 &1.00 &0.44 &0.56 &1.00 &0.58 &0.42 &1.00\\
        \hline
    \end{tabular}
    \label{tab:supp_dvlog_splits}
\end{table}

\begin{table}[ht]
    \emph{\caption{Subject proportions for \textbf{MIMIC-III} across data splits. Abbreviations: 
    $Y_0$: Healthy Controls.
    $Y_1$: Major Depressive Disorder. 
    M: male. 
    F: female.
    T: total.
    }
    \label{tab:supp_mimic3_splits}}
    \centering
    \addtolength{\tabcolsep}{-0.8mm}
    \scalebox{1}{
    \begin{tabular}{l|ccc|ccc|ccc} \hline
         & \multicolumn{3}{c|}{\textbf{Train}} & \multicolumn{3}{c|}{\textbf{Val}} & \multicolumn{3}{c}{\textbf{Test}} \\ 
         & $Y_0$ & $Y_1$ &T & $Y_0$ & $Y_1$ &T & $Y_0$ & $Y_1$  &T \\ \hline
         M   
          & 0.33 & 0.10 &0.43
         & 0.29 & 0.12 &0.41
         & 0.46 & 0.06 &0.52\\
         F 
         & 0.46 & 0.11 &0.57
        & 0.49 & 0.10  &0.59
        & 0.43 & 0.06  &0.49 \\ 
        T  
        & 0.79 & 0.21 &1.00
        & 0.78 & 0.22 &1.00
       & 0.88  & 0.12 &1.00\\
       \hline
    \end{tabular}}
\end{table}


\begin{table*}[ht]
    \emph{\caption{Subject proportions for \textbf{MODMA} across 5-fold cross-validation and test set. Abbreviations: 
    $Y_0$: Healthy Controls.
    $Y_1$: Major Depressive Disorder.
    M: male. 
    F: female.
    T: total.
    \label{tab:supp_modma_folds}}
    }
    \centering
    \addtolength{\tabcolsep}{-1mm}
    \scalebox{0.80}{
    \begin{tabular}{l|ccc|ccc|ccc|ccc|ccc|ccc} \hline
        & \multicolumn{3}{c|}{\textbf{Fold 1}} & \multicolumn{3}{c|}{\textbf{Fold 2}} & \multicolumn{3}{c|}{\textbf{Fold 3}} & \multicolumn{3}{c|}{\textbf{Fold 4}} & \multicolumn{3}{c|}{\textbf{Fold 5}} & \multicolumn{3}{c}{\textbf{Test}} \\ 
        & $Y_0$ & $Y_1$  &T 
        & $Y_0$ &$Y_1$ &T 
        & $Y_0$ & $Y_1$ &T 
        & $Y_0$ & $Y_1$ &T 
       & $Y_0$  & $Y_1$ &T
        & $Y_0$ & $Y_1$ &T\\ \hline
M   
& 0.50 & 0.25 &0.76 
& 0.37 & 0.25 &0.62 
& 0.38 & 0.38 &0.76 
& 0.33 & 0.33 &0.67 
& 0.38 & 0.38 &0.76 
& 0.36 & 0.27 &0.63 \\

F 
& 0.12 & 0.12 &0.24 
& 0.25 & 0.12 &0.37 
& 0.12 & 0.12 &0.24 
& 0.22 & 0.11 &0.33 
& 0.12 & 0.12 &0.24 
& 0.18 & 0.18 &0.37 \\

T 
&0.62 &0.38 &1.00 
&0.63 &0.37 &1.00 
&0.50 &0.50 &1.00 
&0.55 &0.45 &1.00 
&0.50 &0.50 &1.00 
&0.55 &0.45 &1.00 \\ 
\hline
    \end{tabular}
}
\end{table*}


\begin{table}[ht]
    \emph{\caption{Subject proportions for \textbf{MIMIC-CXR} across data splits. Abbreviations: 
    $Y_0$: Control group.
    $Y_1$: positive event group.
    M: male. 
    F: female.
    T: total.
    }
    \label{tab:supp_mimiccxr_splits}}
    \centering
    \addtolength{\tabcolsep}{-0.8mm}
    \scalebox{1}{
    \begin{tabular}{l|ccc|ccc|ccc} \hline
         & \multicolumn{3}{c|}{\textbf{Train}} & \multicolumn{3}{c|}{\textbf{Val}} & \multicolumn{3}{c}{\textbf{Test}} \\
         & $Y_0$ & $Y_1$ &T & $Y_0$ & $Y_1$ &T & $Y_0$ & $Y_1$  &T \\ \hline
         M   
          & 0.20 & 0.29 &0.48
         & 0.20 & 0.29 &0.49
         & 0.20 & 0.29 &0.49\\
         F 
         & 0.18 & 0.34 &0.52
        & 0.18 & 0.33 &0.51
        & 0.18 & 0.33 &0.51 \\ 
        T  
        & 0.37 & 0.63 &1.00
        & 0.38 & 0.62 &1.00
       & 0.38 & 0.62 &1.00\\
       \hline
    \end{tabular}}
\end{table}


\subsubsection{D-Vlog}
\label{app:dvlog_splits_note}

With reference to Table \ref{tab:supp_dvlog_splits}, we see that there is slight gender imbalance in favour of females. The dataset is considered relatively balanced across the different outcome classes ($Y_0$ vs. $Y_1$).
%

\subsubsection{MIMIC-III}
\label{app:mimic_splits_note}
With reference to Table \ref{tab:supp_mimic3_splits}, we see that gender is comparatively more balanced across gender. 
However, there is severe class imbalance across the different outcome classes ($Y_0$ vs. $Y_1$).
Thus, this may mean that models trained on MIMIC-III should be relatively fair across gender but performance metrics such as F1 may be compromised given the ground-truth outcome class imbalance.

\subsubsection{MODMA}
\label{app:modma_splits_note}
Looking at Table \ref{tab:supp_modma_folds}, we see that the dataset is imbalanced across both gender and class, with males and the class $Y_0$ being the majority in all splits. 
This may be one of the key factor for the difficulty of most models to learn a robust and fair representation for the females of class $Y_1$.

\subsubsection{MIMIC-CXR}
\label{app:mimiccxr_splits_note}
With reference to Table \ref{tab:supp_mimiccxr_splits}, we see that gender is comparatively balanced across the different splits. 
However, there is consistent class imbalance across the different outcome classes ($Y_0$ vs. $Y_1$), with $Y_1$ being the majority in train, validation, and test.

\subsection{Implementation Details}
\label{sect:implementation_details}

We perform Bayesian optimization over the training and validation folds via liner probing for all models.
We adopt the Adam optimizer \cite{kingma2014adam}
unless otherwise specified by the original publication.
%
Early stopping was used if the validation metric did not improve after a patience of 3-10 epochs with a threshold between $10^{-5}$ and $10^{-2}$, 
as decided by Bayesian optimization. 
%
The Bayesian optimization for was done for around 3000 steps per MODMA and Dvlog model, and around 1500 steps per MIMIC-III model.
%
The search space included SL and SSL learning-rates, sampled unifrotmly from a range of $(10^{-5}, 1)$, while the model-specific loss-function hyper-parameters were sampled from the range $(0, 100)$. Over MODMA and Dvlog, the SSL epochs were tuned within (1, 100) and the SL fine-tuning epochs were tuned within (1, 40). While for MIMIC-III, the SL epochs were taken within (1, 10) and SSL epochs within (1, 50).
The hyper-parameter configuration with the highest validation score was retrained on the combined training and validation data and evaluated on the held-out test set. For MODMA and Dvlog, the tuning was done to maximize the accuracy results, but Bayesian optimization sometimes explores close regions and might give similar results for multiple configurations.
See Section \ref{sect:hyperparameters} for the precise hyper-parameters that were found to be optimal via this procedure.
In such instances,
the decision was made based on the maximal aggregated Accuracy and F1.
%
%
%
We pair segments from both modalities in the same extraction order.
In baseline cases where the model requires the same name number of input segments for both encoders, we padded the modality with fewer segments to 
ensure that no data segment was ignored during training.


\subsubsection{D-Vlog}
\label{app:impl_dvlog}

We adopt \citet{Yoon_Kang_Kim_Han_2022} implementation 
%
as the baseline model.
We train the model with the Adam optimizer \cite{kingma2014adam} at a learning rate of 0.0002 and a batch size of 32 for D-Vlog as stated in Yoon \textit{et al}.

\subsubsection{MIMIC-III}
\label{app:impl_mimic}
We adopt 
\citet{harutyunyan2019multitask}'s pre-processing pipeline, 
remove all rows with $>$ 50 \% missing entries, and conduct column-wise mean-imputation
on the remaining missing values
\cite{wang2020mimic}. 

\subsubsection{MODMA}
\label{app:impl_modma}
We adopt 
\citet{wang2022ecapa}'s and \citet{khan2024machine}'s pre-processing pipeline for the audio and EEG modality respectively.
We remove subjects which did not have both modalities which resulted 
%
in a total sample of 41.
%
%
As most of the existing works on MODMA had significant data leakage and did not take gender and label balance into account within their train-val-test splits, 
we conduct our own splits on MODMA where we ensured that subjects and label balance were both taken into account in all folds and test sets as evidenced in Table S7.
%
%

For MODMA, we encountered
challenges unique to EEG datasets
such as
data drift \cite{mari2025machine,pontes2024concept},
and reproducibility challenges (i.e. inability to derive the same results using the same experimental setup) \cite{kinahan2024achieving}.
Moreover, past works did not adopt a subject-independent classification protocol.
%
We 
adopt an evaluation protocol with no data leakage and provide the dataset split in the Supp. Mat. to facilitate reproducibility.

\subsubsection{MIMIC-CXR}
\label{app:impl_mimiccxr}

We use the original MIMIC-CXR provided dataset directly. The data was first filtered to subjects that had a minimum of 2 text segments available, considering that multiple contrastive learning methods required such structure. Encoders used were ResNet-18 for the image modality, and a Transformer encoder for the text modality. For the loss-specific hyperparameters, we selected the same hyperparameters reported by the corresponding papers, and the learning rates for linear probing and pre-training were tuned using grid search over a grid of [5e-3, 1e-4, 5e-4], while the linear probe and pre-training epochs were tuned over a grid of [5, 25, 50]. Following a similar protocol as the one used for MIMIC-CXR, the grid search was done to detect the hyperparameter combination that maximized the AUC score. The model was then evaluated over the test set.



\begin{table}[hbt!]
\caption{\textbf{DVlog:} \textbf{Multimodal Results} across both performance and fairness
    Best and second-best results are highlighted in \first{bold} and \second{underline}, respectively. \label{tab:supp_dvlog_multimodal}}
\small
\addtolength{\tabcolsep}{-1.6mm}

  \centering
  \begin{tabular}{l|l|cccc|cccc|c}
    \hline
      & \multirow{2}{*}{\textbf{Method}} & \multicolumn{4}{c|}{\textbf{Performance}} &\multicolumn{5}{c}{\textbf{Group Fairness}}   \\
     
     &     & Acc.	&Prec.	&Rec.	&F1	
     & $SP$ & $EOpp$  & $EOdd$  & $EAcc$ &$AGG_F$ \\ \cline{1-11}
     
 \multirow{5}{*}{\rot{\scriptsize \textbf{Transformer}}}
 &Bimod-cross	&\first{0.67}	&\first{0.70}	&0.74	
 &0.72	
 &0.72	&1.60	&0.76	&0.90 &0.70 \\
 & Bimod-concat	&0.58	&0.65	&0.67	&0.65	
 &0.76	&1.69	& \second{1.02}	&0.74	&0.70 \\
 & Perceiver	&\second{0.62}	& \second{0.69}	&0.64	&0.66	
 & \second{1.08}	&2.38	&1.62	&0.87	&0.45 \\ 
&SEResnet	&0.57	&0.58	&\second{0.94}	&\second{0.72}	
&0.82	&2.13	&1.04	&0.82 &0.62 \\
&DepressionDet	&\second{0.62}	&0.66	&0.71	&0.69	
&0.86	&1.91	&1.17	&0.79 &0.64 \\

    \cline{1-11}
    
 \multirow{2}{*}{\rot{\scriptsize\textbf{CNN}}}
 &Xception-add	&0.60	&0.66	&0.67	&0.66	
 &0.83	&1.84	&0.90	&0.89  &0.70 \\
 &Xception-concat &0.58	&0.64	&0.65	&0.65	
 &0.63	&1.40	&0.71	&\first{0.99}  &0.73    \\

    \cline{1-11} 
    
\multirow{4}{*}{\rot{\scriptsize \textbf{\methodname}}}
&M1	&0.58	&0.58	&\first{1.00}	&\first{0.73}	
&\first{1.00}	&2.21	& \first{1.00}	&0.90 &0.67 \\
&M2 &0.59	&0.55	&0.76	&0.64	
&0.74	&\first{1.19}	&0.89	&0.99 &\first{0.86} \\
&M3	&0.60	&0.59	&0.71	&0.64	
&0.84	&\second{1.38}	&0.92	&0.89 &\second{0.82} \\

&M4	&0.57	&0.59	&0.48	&0.53	
&0.54	&1.19	&0.55	&\second{0.96} &0.71 \\

    \hline    
    \end{tabular}
    
\end{table}


\begin{table}[hbt!]
\small
\caption{\textbf{DVlog:} \textbf{Unimodal results} of both the audio and visual modalities across performance and classification.
Best and second-best results are highlighted in \first{bold} and \second{underline}, respectively.}
\addtolength{\tabcolsep}{-1.8mm}
\centering
\begin{tabular}{c|l|cccc|cccc|c}
\hline
\multirow{2}{*}{\textbf{Mod.}} & \multirow{2}{*}{\textbf{Method}} & \multicolumn{4}{c|}{\textbf{Performance}} & \multicolumn{5}{c}{\textbf{Group Fairness}} \\
 & & Acc. & Prec. & Rec. & F1 
 & $SP$ & $EOpp$  & $EOdd$  & $EAcc$ &$AGG_F$ \\ \cline{1-11}
\multirow{5}{*}{\rot{\textbf{Audio}}}
& MSCDR                   & 0.62 & 0.64 & \second{0.78} & 0.70 
& 1.14 & 2.51 & 1.76 & 0.77 & 0.34 \\
 & Classify-Net            &\second{0.65}& 0.68 & 0.76 & \second{0.72} 
 & 0.74 & \second{1.63} & 0.76 & 0.87 & 0.68 \\
 & DeiT                    & \first{0.68} & 0.69 & \first{0.80} & \first{0.74} 
 & 0.79 & 1.75 &  \second{0.84}& \second{0.90} & \first{0.70} \\
 & DenseNet-201            & 0.60 &  \second{0.70} & 0.55 & 0.62 
 & 0.62 & \first{1.37}  & 0.64 & 0.86 & \second{0.69} \\
& DeprDet (baseline) & 0.61 & 0.65 & 0.71 & 0.68 
& \second{1.09} &  2.27 & 1.28 & 0.87 & 0.56 \\
\cline{1-11}
\multirow{3}{*}{\rot{\textbf{Visual}}}
& FacialPulse             & 0.60 & 0.62 & 0.27 & 0.62
& 0.82 & 1.82 & \first{0.87}  &\second{0.90}& \second{0.69} \\
 & FL-ST-GCN   & 0.58 &  \first{0.77} & 0.38 & 0.51 
 & 1.12 & 3.07 &0.00 &  \first{0.93} &0.18 \\
& DeprDet (baseline) & 0.63 & 0.66 & 0.73 & 0.69 
&\first{0.94} & 2.16 & 1.61 & 0.75 & 0.48 \\
\hline
\end{tabular}
\label{tab:supp_dvlog_combined}
\end{table}

\begin{table}[hbt!]
\caption{\textbf{D-Vlog:} 
A comparison of the performance and fairness across different \textbf{SSL-based} multimodal methods. 
Best results \textbf{for each method} are highlighted in \textbf{bold}.
Abbreviations: DP: Double pooling. Best and second-best results are highlighted in \first{bold} and \second{underline}, respectively.
}
\small
\addtolength{\tabcolsep}{-1.3mm}
  \centering
  \begin{tabular}{l|l|cccc|cccc|c}
    \hline
      & \multirow{2}{*}{\textbf{Method}} & \multicolumn{4}{c|}{\textbf{Performance}} &\multicolumn{5}{c}{\textbf{Group Fairness}}   \\
     
     &     & Acc.	&Prec.	&Rec.	&F1   & $SP$ & $EOpp$  & $EOdd$  & $EAcc$ &$AGG_F$ \\ \cline{1-11}
     
 \multirow{5}{*}{\rot{\textbf{CoMM}}}
 &Baseline	 &0.62	 &\second{0.73}	 &0.55	 &0.63	 &1.30	 &2.88	 &4.49	 &0.77 &0.47\\
 \cdashline{2-11}[.4pt/2pt]   
 &M1	 &0.63	 &0.70	 &0.60	 &0.65	 &1.16	 &2.57	 &2.11	 &0.86 &0.25\\
 &M2	 &0.57	 &0.50	 &0.22	 &0.31	 &0.44	 &0.33	 &\first{1.00}	 &0.95  &0.68\\
 &M3	 &0.62	 &0.70	 &0.60	 &0.65	 &0.84	 &1.86	 &\first{1.00}	 &0.89 &0.72\\
 &M4	 &0.62	 &0.69	 &0.62	 &0.65	 &0.97	 &2.14	 &1.51	 &0.84 &0.54\\

\cline{1-11}
\multirow{5}{*}{\rot{\textbf{FOCAL}}}
 &Baseline	&0.62	&0.71	&0.57	&0.63	&1.25	&2.77	&3.26	&0.86	&0.10\\
 \cdashline{2-11}[.4pt/2pt]   
 &M1	 &0.62	&0.70	&0.59	&0.64	&1.17	&2.64	&2.64	&0.84	&0.10\\
 &M2	 &0.52	&0.43	&0.33	&0.38	&1.04	&0.75	&\first{1.00}	&1.11	&\textbf{0.90}\\
 &M3	&0.60	&0.57	&0.28	&0.37	&1.22	&\first{0.91}	&1.38	&1.17	&0.79\\
 &M4	 &0.59	&0.61	&0.80	&0.69	&\second{0.99}	&2.11	&1.12	&0.93	&0.67\\

\cline{1-11}
 \multirow{5}{*}{\rot{\textbf{Quest}}}
 &Baseline	&\second{0.64}	&0.72	&0.63	&0.67	&1.09	&2.41	&2.64	&0.80	&0.17\\
 \cdashline{2-11}[.4pt/2pt]   
 &M1	 &0.62	&0.60	&0.33	&0.43	&0.89	&0.67	&\first{1.00}	&1.14	&0.85\\
 &M2	 &0.58	&0.58	&\second{0.97}	&0.67	&0.97	&2.15	&\second{1.01}	&0.86	&0.33\\
 &M3	&0.56	&0.57	&0.87	&0.69	&0.88	&1.85	&0.89	&0.87	&0.70\\
 
 &M4	&0.62	&0.67	&0.70	&0.68	&1.08	&2.39	&1.51	&0.82  &0.46\\

\cline{1-11}
\multirow{5}{*}{\rot{\textbf{DeCUR}}}
 &Baseline	&0.59	&\first{0.74}	&0.41	&0.52	&1.47	&3.25	&0.00	&0.95	 &0.06\\
 \cdashline{2-11}[.4pt/2pt]   
 &M1	 &0.63	 &0.66	 &0.76	 &0.70	 &1.10	 &2.44	 &1.69	 &0.81	 &0.39 \\
 &M2	 &0.62	 &0.68	 &0.64	 &0.66	 &0.86	 &1.90	 &1.22	 &0.81	 &0.64 \\
 &M3  &\first{0.65}	 &0.56	 &0.56	 &0.56	 &0.67	 &0.50	 &\first{1.00}	 &0.83	 &0.75\\
 &M4	  &0.61	 &0.71	 &0.57	 &0.63	 &0.90	 &2.00	 &0.91	 &0.88	 &0.67\\

\cline{1-11}
\multirow{8}{*}{\rot{\textbf{Ours}}}
 &Baseline	&0.57	&\first{0.74}	&0.40	&0.52	&1.41	&3.13	&2.25	&0.95	&0.04\\
 \cdashline{2-11}[.4pt/2pt]   
&M1	&0.58	&0.58	&\first{1.00}	&\second{0.73}		&\first{1.00}	&2.21	&\first{1.00}	&0.90 &0.67 \\
&M2 &0.59	&0.55	&0.76	&0.64		&0.74	&\second{1.19}	&0.89	&\first{0.99} &\second{0.86}\\
&M3	&0.60	&0.59	&0.71	&0.64	&0.84	&1.38	&0.92	&0.89 &0.82\\
&M4	&0.57	&0.59	&0.48	&0.53		&0.54	&\second{1.19}	&0.55	&0.96 &0.71 \\

\cdashline{2-11}[.4pt/2pt]

& M1+DP & 0.62 & 0.68 & 0.80 & \second{0.73} & 1.06 & 2.35 & 1.61 & 0.82 & 0.45  \\ 
& M2+DP & 0.61 & 0.62 & 0.86 & 0.72 & \second{0.99} & 2.09 & 1.13 & 0.84 & 0.66  \\ 
& M3+DP & \first{0.65} & 0.64 & 0.94 & \textbf{0.78} & 0.95 & 2.10 & \first{1.00} & \second{0.98} & 0.71  \\ 
& M4+DP & 0.53 & 0.53 & 0.78 & 0.63 & 0.93 & 2.05 & 1.04 & 0.75 & 0.65  \\

    \cdashline{1-11}[.4pt/2pt]

   \hline  
    \end{tabular}
    \label{tab:supp_dvlog_ssl}
\end{table}


\subsection{Details of the Compared Methods}
\label{sect:app_compared_methods}

\subsubsection{D-Vlog}
\label{app:methods_dvlog}

We 
benchmark \methodname against five 
state-of-the-art (SOTA) methods:
\textbf{(i) Depression Detector} (DeprDet), the original model 
employed by the DVlog authors
\cite{Yoon_Kang_Kim_Han_2022},
where two unimodal modality-specific Transformer encoders (acoustic and visual) are 
fused by a third encoder.
%
\textbf{(iii) Multimodal Xception} 
\cite{he2024lmvd} which integrates depthwise-separable 2D convolutions and 1D depthwise-separable convolutions on the respective video and audio files.
\textbf{(iv) Perceiver} is a 
modality-agnostic encoder that 
iteratively attends between input modalities and a lower-dimensional latent array \cite{gimenogomez2024reading}. 
Finally, \textbf{(v) SEResnet} incorporates Squeeze-and-Excitation (SE) blocks that recalibrate channel-wise feature maps \cite{hu2018squeeze}.
%


\subsubsection{MODMA}
\label{app:methods_modma}

We benchmark our approach on MODMA 
against 
\textbf{(i) MultiDepr}, 
an attention-based network with selective dropout  
\cite{ahmed2023taking}, 
\textbf{(ii) EfficientNetV2-S} 
a convolutional baseline which leverages balanced depth–width–resolution scaling to extract features from EEG and audio spectrograms 
\cite{qayyum2023highdensity},
\textbf{(iii) EMO-GCN} which fuses EEG and audio features using an adaptive multi-graph neural network with modality-specific GCN modules and attention-driven graph pooling featuring structural learning \cite{xing2024adaptive} and
a \textbf{(iv) transformer} baseline 
which combines two modality-specific tansformers
\cite{lee2024eav}.
Finally, we use \textbf{(v) FeatNet} \cite{singh2024learning}
which is a 2-D CNN with Conv–InstanceNorm–LeakyReLU blocks and adaptive average pooling 
which we also adopt 
as the backbone of our MODMA \methodname variant, and the supervised linear probing results of this model are reported in Table \ref{tab:supp_modma_multimodal}. 
%

%


\subsubsection{MIMIC-III}
\label{app:methods_mimic}

We compare \methodname with six recent multimodal SSL frameworks, including contrastive methods and redundancy-reduction methods.
\textbf{CoMM} aligns fused modality embeddings by maximizing mutual information, thereby capturing redundancy, uniqueness, and synergy without enforcing explicit cross-modal projections \cite{dufumier2025what}. 
\textbf{FOCAL} decomposes representations into shared and private orthogonal subspaces and couples contrastive losses with a temporal structural constraint to disentangle consistent versus modality-specific features \cite{liu2023focal}. 
\textbf{QUEST} leverages quaternion-valued embeddings in a quadruple-contrastive setup, introducing orthogonal constraints and self-penalization to prevent feature suppression and disentangle shared and unique information \cite{song2024quest}. 
\textbf{DeCUR} decouples common and unique factors through cross-modal and intra-modal redundancy reduction losses, optionally enhanced by deformable attention for richer modality-specific representations \cite{wang2024decur}. 
\textbf{FACTORCL} 
factorizes contrastive objectives into separate streams for shared and unique information, optimizing mutual information bounds via tailored multimodal augmentations \cite{liang2023factorized}. 
Finally, \textbf{SimCLR} adapts the canonical contrastive paradigm to time-series by employing scaling and signal-inversion augmentations on a three-layer Conv1D encoder with a projection head and gradual unfreezing during fine-tuning \cite{yfantidou2024using}. 
In addition to MIMIC-III, we apply these SSL approaches on D-Vlog and MODMA when applicable, both the original form of each method and when combined with \methodname.


We compare our approach on MODMA and Dvlog against eight unimodal baselines:
\textbf{(i) EEGNet} is a compact CNN for EEG-based BCIs that applies temporal convolution for band-pass filtering, depthwise spatial convolution for channel mixing, and separable convolutions for feature learning \cite{lawhern2018eegnet, zhang2024depressionlevelclassification}. 
\textbf{(ii) DeprNet} proposes a novel 1-D CNN framework comprising a CNN that jointly captures spatial and temporal information from EEG segments for depression detection \cite{seal2021deprnet}. 
\textbf{(iii) HybGNN} integrates a Common GNN branch with a learnable fixed adjacency to capture shared depression patterns and an Individualized GNN branch with adaptive connections, plus a Graph Pooling and Unpooling Module to extract hierarchical EEG features on MODMA \cite{wang2024hybrid};
\textbf{(iv) FeatureNet} implements dual 1-D convolutional pathways, a small branch with narrow, fine-scale filters and a large branch with wide filters, where channel-wise embeddings are flattened, concatenated, and passed through a dropout-regularized two-layer MLP to predict depression \cite{jia2021multiviewsg}. 
\textbf{(v) MSCDR} extracts LPC and MFCC features from speech segments, processes them via parallel 1D-CNNs, and fuses outputs with an LSTM to model correlations \cite{du2023depressionrecognition}. 
\textbf{(vi) Classify-Net} is a fully-connected network that receives fused spectrogram and MFCC features, processes them through a CNN, and applies a linear transform to hierarchically extract salient representations for depression detection \cite{wu2025enhanced}. 
\textbf{(vii) DeiT} (Data-Efficient Image Transformer) enhances the standard ViT by introducing a distillation token and teacher–student training strategy, enabling competitive performance on limited data via positional encodings and distilled supervision \cite{pratiwi2025visiontransformer}, and 
\textbf{(viii) Modified Densenet201} ingests Mel-spectrograms, freezes pretrained convolutional blocks, and fine-tunes 
them via transfer learning \cite{sun2023audio}.
\textbf{Facial Landmarks ST-GCN} propose a privacy-preserving engagement measurement method that feeds facial landmarks into a Spatial-Temporal GCN trained under an ordinal transfer-learning framework, allowing for lightweight and interpretable \cite{abedi2024engagement}. 
\textbf{FacialPulse} uses a Facial Landmark Calibration Module, combining sparse optical flow, two-way denoising, and Kalman filtering to stabilize landmarks. They then apply Facial Motion Modeling Module to encode temporal dynamics \cite{wang2024facialpulse}.
%


\begin{table}[hbt!]
\caption{\textbf{MIMIC-III:} SSL-based methods finetuned on \textbf{accuracy}. 
    Comparison of the performance and fairness across different baseline multimodal SSL methods.
    Best results across all methods are highlighted in \textbf{bold}. Best and second-best results are highlighted in \first{bold} and \second{underline}, respectively.\label{tab:supp_MIMIC_ssl_accuracy}
    }
\small
\addtolength{\tabcolsep}{-1.4mm}
  \centering
  \begin{tabular}{l|l|cccc|cccc|c}
    \hline
    & \multirow{2}{*}{\textbf{Method}} & \multicolumn{4}{c|}{\textbf{Performance}} &\multicolumn{5}{c}{\textbf{Group Fairness}}   \\
     
     &     & Acc.	&Prec.	&Rec.	&F1	   & $SP$ & $EOpp$  & $EOdd$  & $EAcc$ &$AGG_F$ \\ \cline{1-11}
     
 \multirow{5}{*}{\rot{\textbf{MM SSL}}}    
&CoMM	&\second{0.84}	&0.17	&0.10	&0.12		&0.84	&0.79	&0.86	&1.03  &0.86 \\

&DeCUR	&0.68	&0.14	&\second{0.36}	&0.20	&1.36	&1.28	&1.53	&0.83 &0.67 \\

&FOCAL	&0.82	&0.17	&0.14	&0.15	&1.67	&1.57	&2.04	&0.92 &0.41 \\
									
&FACTORCL	&0.81	&0.16	&0.17	&0.16	&0.78	&0.67	&0.54	&1.04 &0.74 \\

&SimCLR	&0.74	&\second{0.32}	&0.35	&0.18 &0.92	&0.86	&\first{1.01}	&0.97 &\second{0.94} \\

    \hline
    
\multirow{4}{*}{\rot{\textbf{Ours}}} 
 &M1	&0.82	&0.19	&0.17	&0.17	&\second{1.06}	&\first{1.01}	&1.21	&0.96	&0.92 \\
 &M2  &0.65	&0.18	&\first{0.57}	&\first{0.28}	&\first{1.03}	&0.95	&1.01	&\second{0.99} &\first{0.98}\\
 &M3 &0.77	&0.17	&0.26	&0.21	&0.84	&0.89	&\second{0.89}	&\first{1.00} &0.90 \\
 &M4 &\first{0.86}	& \first{0.34}	&0.20	&\second{0.26}	&0.89	&0.84	&1.07	&0.98  &0.91 \\
 
   \hline  
    \end{tabular}
    
\end{table}


\begin{table}[hbt!]
\caption{\textbf{MIMIC-III:} 
    SSL-based methods finetuned on \textbf{AUROC}.
   Comparison of the performance and fairness across different multimodal SSL methods. Best and second-best results are highlighted in \first{bold} and \second{underline}, respectively.
   \label{tab:supp_MIMIC_ssl_auroc}  }
\small
\addtolength{\tabcolsep}{-1.7mm}
\centering
\begin{tabular}{l|l|cccc|cccc|c|c}
\hline
\scriptsize
& \multirow{2}{*}{\textbf{Method}}
& \multicolumn{4}{c|}{\textbf{Performance}}
& \multicolumn{4}{c|}{\textbf{Fairness}}
& \multirow{2}{*}{AUROC}
& \multirow{2}{*}{$AGG_F$} \\[-2pt]
& & Acc. & Prec. & Rec. & F1
& $SP$ & $EOpp$  & $EOdd$  & $EAcc$ \\ \cline{1-12}
%
\multirow{5}{*}{\rot{\textbf{CoMM}}}
& Baseline   & 0.86 & 0.37 & 0.14 & 0.21 & 0.94 & 0.89 & 1.79 & 0.95 & 0.68 & 0.75 \\
 \cdashline{2-12}[.4pt/2pt]   
& M1   & 0.82&	0.22	&0.20	&0.21&	0.92&	0.87&	1.56&	0.95&	0.67	&0.79      \\
& M2         & 0.84 & 0.21 & 0.18 & 0.19 & \second{0.97} & 0.92 & 1.14 & 0.96 & 0.68 & \second{0.93} \\
& M3         & 0.86 & 0.26 & 0.12 & 0.16 & 0.93 & 0.88 & 1.28 & 0.97 & 0.67 & 0.88 \\
& M4         & 0.87 & 0.36 & 0.18 & 0.24 & 1.13 & \second{1.06} & 1.43 & 0.97 & 0.66 & 0.84 \\
\cline{1-12}
%
\multirow{5}{*}{\rot{\textbf{DeCUR}}}
& Baseline   & 0.87 & 0.17 & 0.14 & 0.15 & 2.48 & 2.33 & 4.28 & 0.96 & 0.61 & 0.54 \\
 \cdashline{2-12}[.4pt/2pt]   
& M1   &0.85&	0.23	&0.17	&0.19	&2.23	&2.10	&4.59	&0.88	&0.66	&0.51     \\
& M2         & 0.87 & 0.25 & 0.08 & 0.12 & 0.64 & 0.60 & \second{1.07} & \second{0.99} & 0.60 & 0.79 \\
& M3         & 0.85 & 0.26 & 0.17 & 0.20 & 1.23 & 1.15 & 2.09 & 0.92 & 0.63 & 0.61 \\
& M4         & 0.85 & 0.32 & 0.23 & 0.27 & 1.13 & \second{1.06} & 1.65 & 0.94 & 0.67 & 0.77 \\
\cline{1-12}
\multirow{5}{*}{\rot{\textbf{FOCAL}}}
& Baseline   & 0.87 & 0.33 & 0.11 & 0.16 & 0.80 & 0.75 & 1.43 & 0.96 & 0.67 & 0.77 \\
 \cdashline{2-12}[.4pt/2pt]   
& M1 & 0.86	&0.33	&0.10	&0.15&	0.74&	0.70&	0.86	&0.98  & 0.65 & 0.82      \\
& M2         & 0.87 & 0.29 & 0.12 & 0.15 & \first{1.02} & \first{0.96} & 1.41 & 0.97 & 0.64 & 0.88 \\
& M3         & \second{0.88} & 0.42 & 0.04 & 0.07 & 0.76 & 0.71 & 1.43 & 0.98 & 0.62 & 0.76 \\
& M4         & 0.87 & 0.35 & 0.11 & 0.17 & 0.84 & 0.79 & \second{1.07} & 0.98 & 0.67 & 0.89 \\
\cline{1-12}
\multirow{5}{*}{\rot{\textbf{FACTORCL}}}
& Baseline   & 0.84 & 0.26 & \second{0.24} & 0.25 & 0.64 & 0.60 & 0.68 & 1.03 & 0.65 & 0.72 \\
 \cdashline{2-12}[.4pt/2pt]   
& M1  & 0.80&	0.19&	0.22&	0.20	&0.61&	0.57	&0.65&	1.03&	0.65&	0.70       \\
& M2         & 0.86 & 0.33 & \second{0.24} & \second{0.28} & \first{1.02} & \first{0.96} & 1.14 & 0.98 & \second{0.69} & \first{0.94} \\
& M3         & \first{0.89} & 0.50 & 0.21 & \first{0.30} & 0.89 & 0.93 & 1.43 & 0.97 & 0.68 & 0.84 \\
& M4         & \first{0.89} & \second{0.70} & 0.11 & 0.18 & 0.87 & 0.82 & 2.14 & 0.98 & 0.67 & 0.63 \\
\cline{1-12}
\multirow{5}{*}{\rot{\textbf{SimCLR}}}
& Baseline   & \second{0.88} & \first{0.86} & 0.05 & 0.09 & 0.43 & 0.40 & 0.00 & 0.98 & \first{0.70} & 0.45 \\
 \cdashline{2-12}[.4pt/2pt]   
& M1  &0.88&	0.75&	0.05	&0.90	&0.36	&0.37&	0.00&	0.98&	0.66	&0.43        \\
& M2         & 0.86 & 0.32 & 0.18 & 0.23 & 0.77 & 0.72 & \first{0.99} & 0.97 & 0.61 & 0.86 \\
& M3         & \first{0.89} & 0.58 & 0.11 & 0.19 & 1.24 & 1.17 & 1.77 & 0.98 & 0.63 & 0.70 \\
& M4         & \second{0.88} & 0.50 & 0.09 & 0.15 & 0.90 & 0.85 & 2.14 & 0.97 & 0.67 & 0.64 \\
\cline{1-12}
\multirow{4}{*}{\rot{\textbf{Ours}}}
& M1          & 0.87 & 0.32 & 0.14 & 0.20 & 0.90 & 0.84 & 0.90 & \first{1.00} & 0.62 & 0.91 \\
& M2          & 0.86 & 0.34 & 0.20 & 0.25 & 0.96 & 0.90 & 1.16 & 0.97 & \second{0.69} & 0.92 \\
& M3          & 0.85 & 0.21 & 0.21 & 0.25 & 0.82 & 0.77 & 0.89 & \second{0.99} & 0.65 & 0.87 \\
& M4          & 0.82 & 0.26 & \first{0.31} & \second{0.28} & 0.83 & 0.78 & 0.90 & \second{0.99} & 0.67 & 0.88 \\
\cdashline{1-12}[.4pt/2pt]
\hline
\end{tabular}
    
\end{table}


\subsection{Fairness Measure Definitions}
\label{sect:appendix_fairness_measures}

Let each individual belong to one of two demographic groups, denoted by 
$g_i\in\{0,1\}$, 
where $g=0$ is taken as female subjects and $g=1$ male subjects. 
We write $P(\hat y=1\mid g_j)$ for the probability that the classifier predicts the positive label for group~$g=j$, 
and $P(\hat y=1\mid y=i,\,g=j)$ for the probability of predicting positive given true label $y=i$ in group~$g=j$.

\begin{itemize}   
\item\textbf{Statistical Parity}, or demographic parity, is based purely on predicted outcome $\hat{y}$ and independent of actual outcome $y$:
    \begin{equation}
    \label{eqn:SP}
     SP= \frac{P(\hat{y}=1|g=0 ) }{ P(\hat{y}=1 | g=1)} .
    \end{equation}
    According to this measure, in order for a classifier to be deemed fair, $P(\hat{y}=1 | g=1) = P(\hat{y}=1|g=0 ) $.

\item\textbf{Equal opportunity} states that both demographic groups $g_0$ and $g_1$ should have equal True Positive Rate (TPR). 
    \begin{equation}
    \label{eqn:EOpp}
    EOpp = \frac{P(\hat{y}=1|y=1, g=0 )}{P(\hat{y}=1 | y=1, g=1)}.
    \end{equation}
    According to this measure, in order for a classifier to be deemed fair,  $P(\hat{y}=1 | y=1, g=1) = P(\hat{y}=1|y=1, g=0 ) $.
     
\item\textbf{Equalised odds} can be considered as a generalization of Equal Opportunity where the rates are not only equal for $y=1$, but for all values of $y \in \{1, ... k\}$, i.e.: 
    \begin{equation}
        \label{eqn:EOdd}
        EOdd = \frac{P(\hat{y}=1|y=i, g=0 )}{P(\hat{y}=1 | y=i, g=1)} .
    \end{equation}
    According to this measure, in order for a classifier to be deemed fair, $P(\hat{y}=1 | y=i, g=1) = P(\hat{y}=1|y=i, g=0 ),  \forall  i \in \{1, ... k\}$.

\item\textbf{Equal Accuracy} states that both subgroups $g_0$ and $g_1$ should have equal rates of accuracy.
    \begin{equation}
        \label{eqn:Wacc}
        EAcc = \frac{Acc_{g=0}}{Acc_{g=1}} .
    \end{equation}
\end{itemize}


\section{Theoretical Validation of \methodname}
\label{sect:theory}

In this section, we provide a theoretical justification for why \methodname improves group fairness for linear downstream classifiers. Specifically, we show that reducing the distance between group mean embeddings and controlling within-group score variability jointly bounds statistical parity and other group fairness ratios (and thus $AGG_F$) toward $1$.

\subsection{Notation}
\label{app:theory_notation}

Let $z$ denote the embedding (e.g., the one learned by \methodname) for a sample with protected attribute $g \in \{0, 1\}$. We use capital letters $(Z,G)$ to denote the random variables corresponding to $(z,g)$, and lowercase $(z,g)$ for their realizations.

We follow the standard linear-probe setting for representation learning and analyse a downstream binary classifier whose score is linear in the learned embedding, i.e., $h(z)=w^\top z$ where any bias can be absorbed into $z$ by appending a constant coordinate, followed by an activation function
$\sigma$ (e.g., the Sigmoid) that produces predicted probabilities
$\hat{p}(y=1 \mid z,g) = \sigma(h(z))$. We interpret the prediction
$\hat{y}$ as a Bernoulli random variable with
$\mathbb{P}(\hat{y}=1 \mid z,g) = \hat{p}(y=1 \mid z,g)$, so that
$\mathbb{P}(\hat{y}=1 \mid g) = \mathbb{E}[\sigma(h(z)) \mid g]$.
Throughout we will use the fact that that $\sigma$ is $L$-Lipschitz, that is
$\lvert \sigma(a) - \sigma(b) \rvert \le L \lvert a - b \rvert$ for all $a,b$.

We define the group
mean embeddings $\mu_g := \mathbb{E}[Z \mid G=g]$ for $g \in \{0,1\}$. The corresponding mean scores are
$\bar{h}_g(w) := \mathbb{E}[h(Z) \mid G=g] = w^\top \mu_g$, and we define
the group score disparity between two groups as:
\begin{equation}
\Delta_g(w) := \bar{h}_1(w) - \bar{h}_0(w)
             = w^\top (\mu_1 - \mu_0).
\end{equation}
For each group $g$, we also define the within group score deviation as
$v_g(w) := \mathbb{E}\!\left[\,\lvert h(Z) - \bar{h}_g(w) \rvert \,\middle|\, G=g \right]$.

\paragraph{Conditioning on an event.}
Let $A$ be any event with $\mathbb{P}(A)>0$ such that $\mathbb{P}(G=g,A)>0$ for both $g\in\{0,1\}$.
We use the shorthand
$
\pi_{g,A} := \mathbb{P}(G=g\mid A) $ and $
\mu_{g,A} := \mathbb{E}[Z \mid G=g, A].
$ We also write $\Sigma_{g,A} := \mathrm{Cov}(Z\mid G=g,A)$ for the conditional covariance matrix. We also write $\pi_{\min,A}:=\min(\pi_{0,A},\pi_{1,A})$ and $(t)_+:=\max(t,0)$.

For a linear probe $h(z)=w^\top z$, we then define the conditional mean score and score gap
\[
\bar h_{g,A}(w) := \mathbb{E}[h(Z)\mid G=g,A],
\qquad
\Delta_{g,A}(w) := \bar h_{1,A}(w) - \bar h_{0,A}(w),
\]
and the conditional within-group score deviation
\[
v_{g,A}(w) := \mathbb{E}\!\left[\,\bigl|h(Z)-\bar h_{g,A}(w)\bigr|\,\middle|\,G=g,A\right].
\]
We also write the conditional positive prediction probability and ratio as
\[
p_{g,A}(w) := \mathbb{P}(\hat{y}=1 \mid G=g,A) = \mathbb{E}[\sigma(h(Z))\mid G=g,A],
\qquad
R_A(w) := \frac{p_{0,A}(w)}{p_{1,A}(w)}.
\]
Finally, letting $(Z',G')$ be an independent draw from the same (conditional) distribution as $(Z,G)$ (under the conditional law given $A$), we define the population-level inter-subject invariance term
\[
I^{\methodsuperscript}_{\mathrm{reg},A} := \mathbb{E}\big[\|Z-Z'\|_2^2 \mid A\big].
\]

\subsection{Theoretical Guarantees}
\label{app:theory_guarantees}

\begin{lemma}[Bounding the score gap]
\label{lem:score-gap}
Fix any event $A$ with $\mathbb{P}(A)>0$ such that $\mathbb{P}(G=g,A)>0$ for both $g\in\{0,1\}$.
Then, for any $w$,
\[
|\Delta_{g,A}(w)| \le \|w\|_2\,\|\mu_{1,A}-\mu_{0,A}\|_2.
\]
\end{lemma}


\begin{lemma}[Bounding the conditional positive-rate difference]
\label{lem:rate-bound}
Fix any event $A$ with $\mathbb{P}(A)>0$ such that $\mathbb{P}(G=g, A)>0$ for both $g\in\{0,1\}$, and consider the conditional distribution given $A$. Define $\Delta_A(w):=p_{1,A}(w)-p_{0,A}(w)$.
Then
\begin{equation}
\label{eqn:rate_bound}
\bigl|\Delta_A(w)\bigr|
\;\le\;
L\,\bigl|\Delta_{g,A}(w)\bigr|
\;+\;
L\bigl(v_{1,A}(w)+v_{0,A}(w)\bigr).
\end{equation}
\end{lemma}


\begin{corollary}[Conditional fairness ratio bound]
\label{cor:conditional_ratio_bound}
Let $\mathcal{H}$ be a class of linear classifiers, for example
$\mathcal{H} = \{w \in \mathbb{R}^d : \lVert w \rVert_2 \le C\}$ for some
$C>0$, and let $h(z) = w^\top z$ denote the score of a linear probe
$w \in \mathcal{H}$. Then, fix an event $A$ with $\mathbb{P}(A)>0$ such that $\mathbb{P}(G=g,A)>0$ for both $g\in\{0,1\}$.

Assume that there exists $p_{\min,A}>0$ such that $p_{1,A}(w)\ge p_{\min,A}$ for all $w\in\mathcal{H}$.
Then, for all $w\in\mathcal{H}$,
\begin{equation}
\label{eq:ratio_explicit_bound_A}
\bigl|R_A(w) - 1\bigr|
\;\le\;
\frac{L}{p_{\min,A}}\Bigl(
  \bigl|\Delta_{g,A}(w)\bigr|
  + v_{0,A}(w) + v_{1,A}(w)
\Bigr).
\end{equation}

\end{corollary}


\begin{remark}[Application to $SP$, $EOpp$, $EOdd$, and $AGG_F$]
\label{rem:apply_ratio_to_metrics}
Let $Y$ denote the ground-truth label random variable. Under the assumptions of Corollary~\ref{cor:conditional_ratio_bound} for the corresponding event $A$, the ratios $SP$, $EOpp$, and $EOdd$ defined in Section~\ref{sect:appendix_fairness_measures} are obtained as special cases of $R_A(w)$.

\begin{itemize}
\item \emph{Statistical parity (SP).}
Taking $A=\Omega$ (the whole sample space) yields
\begin{equation}
R_{\mathrm{SP}}(w)
:= \frac{\mathbb{P}(\hat{y}=1\mid G=0)}{\mathbb{P}(\hat{y}=1\mid G=1)}
= R_A(w)\big|_{A=\Omega},
\end{equation}
with
$\Delta_{g,\Omega}(w)=\Delta_g(w)$ and
$v_{g,\Omega}(w)=v_g(w)$.
The bound~\eqref{eq:ratio_explicit_bound_A} specializes to
\begin{equation}
\bigl|R_{\mathrm{SP}}(w)-1\bigr|
\;\le\;
\frac{L}{p_{\min,\Omega}}\Bigl(
  \bigl|\Delta_g(w)\bigr|
  + v_0(w) + v_1(w)
\Bigr),
\end{equation}
which is the explicit decomposition into a mean-score term and a
within-group variability term used for statistical parity.

\item \emph{Equal opportunity (EOpp).}
Taking $A=\{Y=1\}$ gives
$
R_{\mathrm{EOpp}}(w)
= R_A(w)\big|_{A=\{Y=1\}},
$
with corresponding conditional mean-score gap $\Delta_{g,\{Y=1\}}(w)$ and
conditional deviations $v_{g,\{Y=1\}}(w)$. Corollary~\ref{cor:conditional_ratio_bound}
then provides the same type of bound as in Eq. \eqref{eq:ratio_explicit_bound_A},
now for the equal opportunity ratio.

\item \emph{Equalized odds (EOdd).}
For each class $i\in\{1,\dots,k\}$, taking $A=\{Y=i\}$ yields the per-class equalized odds
ratio
$
R_{\mathrm{EOdd},i}(w)
= R_A(w)\big|_{A=\{Y=i\}},
$
again with the same decomposition into a conditional mean-score term and
a conditional within-group variability term.


\item \emph{Aggregated fairness score ($AGG_F$).}
The overall measure $AGG_F$ used in our experiments combines the group fairness
ratios above across events $A=\Omega$, $A=\{Y=1\}$ and $A=\{Y=i\}$, $i\in\{1,\dots,k\}$,
via an aggregation over their deviations from $1$.
Thus, Corollary~\ref{cor:conditional_ratio_bound} provides a unified
view of how the learned representations influence all components of
$AGG_F$ through their mean-score gaps and within-group score variability.
\end{itemize}
\end{remark}


\begin{lemma}[\methodname's inter-subject invariance effect on protected-group mean alignment]
\label{lem:fw_inv_mean_gap}
Let $(Z,G)$ be a random pair where $Z\in\mathbb{R}^d$ is the learned embedding
and $G\in\{0,1\}$ is the protected attribute. Let $A$ be any event with
$\mathbb{P}(A)>0$ such that $\mathbb{P}(G=g,A)>0$ for both $g\in\{0,1\}$.
Let $(Z',G')$ be an independent draw from the conditional distribution of $(Z,G)$ given $A$, and assume $\mathbb{E}\!\left[\|Z\|_2^2 \mid A\right]<\infty$. Then
\begin{equation}
\|\mu_{1,A}-\mu_{0,A}\|_2^2 \;\le\; \frac{I^{\methodsuperscript}_{\mathrm{reg},A}}{2\pi_{0,A}\pi_{1,A}}.
\end{equation}
Equivalently, for any $\varepsilon>0$,
\begin{equation}
I^{\methodsuperscript}_{\mathrm{reg},A}\le \varepsilon
\quad\Longrightarrow\quad
\|\mu_{1,A}-\mu_{0,A}\|_2 \le \sqrt{\frac{\varepsilon}{2\pi_{0,A}\pi_{1,A}}}.
\end{equation}
\end{lemma}


\begin{lemma}[Population-level bound on $v_{0,A}(w)+v_{1,A}(w)$ under \methodname]
\label{lem:two_sided_bound}
Let $(Z,G)$ be a population embedding--group pair with $G\in\{0,1\}$, and let
$A$ be any event with $\mathbb{P}(A)>0$ such that $\mathbb{P}(G=g,A)>0$ for both $g\in\{0,1\}$. Let $h(z)=w^\top z$ with $\|w\|_2=C>0$. 
For each $g\in\{0,1\}$, define the population-level variance and covariance terms of \methodname as
\begin{equation}
V_{\mathrm{reg},A}^{(g)}
\;:=\;
\frac1d\sum_{j=1}^d\big(\gamma-\mathrm{Std}(Z_j\mid G=g,A)\big)_+,
\qquad
C_{\mathrm{reg},A}^{(g)}
\;:=\;
\frac1d\sum_{j\neq k}\mathrm{Cov}(Z_j,Z_k\mid G=g,A)^2,
\end{equation}
for some $\gamma>0$. Suppose that $V_{\mathrm{reg},A}^{(g)}=0$ and
$C_{\mathrm{reg},A}^{(g)}\le \delta$ for both $g\in\{0,1\}$, for some
$\delta\ge 0$, and that there exists $M_A>0$ such that
$\|Z\|_2\le M_A$ almost surely under $A$.
Then the within-group score deviations satisfy the two-sided bound
\begin{equation}
\frac{C}{M_A}\Big(\gamma^2-\sqrt{d(d-1)\,\delta}\Big)_+
\;\le\;
v_{0,A}(w)+v_{1,A}(w)
\;\le\;
C\sqrt{\frac{I^{\methodsuperscript}_{\mathrm{reg},A}}{2}}\Big(\frac1{\sqrt{\pi_{0,A}}}
+\frac1{\sqrt{\pi_{1,A}}}\Big)
\;\le\;
C\sqrt{\frac{2I^{\methodsuperscript}_{\mathrm{reg},A}}{\pi_{\min,A}}}.
\end{equation}
In particular, in the idealized population limit where \methodname's objective
is optimized under $A$, the invariance term $I^{\methodsuperscript}_{\mathrm{reg},A}$ gives us an explicit
upper bound preventing $v_{0,A}(w)+v_{1,A}(w)$ from becoming arbitrarily large,
while the variance/covariance constraints $V_{\mathrm{reg},A}^{(g)}=0$ and small
$C_{\mathrm{reg},A}^{(g)}$ yield an explicit lower bound preventing
$v_{0,A}(w)+v_{1,A}(w)$ from collapsing to $0$ (for any nontrivial probe with
$\|w\|_2=C$).
\end{lemma}

\begin{remark}[Connection between \methodname and group fairness ratios]
\label{rem:fw_connects_to_terms}
The bounds in Corollary~\ref{cor:conditional_ratio_bound} decompose the deviation
from perfect fairness for any event $A$ into a mean-score term
$\bigl|\Delta_{g,A}(w)\bigr|$ and a within-group variability term
$v_{0,A}(w)+v_{1,A}(w)$.

First, Lemma~\ref{lem:score-gap} implies
$
\bigl|\Delta_{g,A}(w)\bigr|
\;\le\;
\|w\|_2\,\|\mu_{1,A}-\mu_{0,A}\|_2.
$
Combined with Lemma~\ref{lem:fw_inv_mean_gap}, this shows that the inter-subject
invariance loss $I^{\methodsuperscript}_{\mathrm{reg},A}$ directly controls the mean-score gap term
for any bounded-norm probe $w\in\mathcal{H}$ and any conditioning event $A$.

Second, Lemma~\ref{lem:two_sided_bound} shows that, under the idealized population 
versions of \methodname's variance and covariance regularizers, the conditional
within-group deviations $v_{0,A}(w)+v_{1,A}(w)$ admit both an explicit positive
lower bound (preventing collapse) and an explicit upper bound controlled by the
same invariance term $I^{\methodsuperscript}_{\mathrm{reg},A}$ (preventing explosion). Thus, for the group fairness ratios in Section~\ref{sect:appendix_fairness_measures} that can be written as $R_A(w)=p_{0,A}(w)/p_{1,A}(w)$ (statistical parity, equal opportunity, and the equalized odds ratios), the corresponding deviation from $1$ admits a bound of the form Eq. \eqref{eq:ratio_explicit_bound_A} in which both right-hand-side terms are linked to \methodname's objective.
\end{remark}


\subsection{Proofs}
\label{app:theory_proofs}

\begin{proof}[Proof of Lemma~\ref{lem:score-gap}]
Working under the conditional law given $A$, we have
$\Delta_{g,A}(w)=\bar h_{1,A}(w)-\bar h_{0,A}(w)=w^\top(\mu_{1,A}-\mu_{0,A})$.
The Cauchy--Schwarz inequality then trivially gives,
$
|\Delta_{g,A}(w)|
\le \|w\|_2\,\|\mu_{1,A}-\mu_{0,A}\|_2.
$
\end{proof}


\begin{proof}[Proof of Lemma~\ref{lem:rate-bound}]
By definition of the conditional positive-rate difference,
\begin{align}
\Delta_A(w)
&=
\mathbb{P}(\hat y = 1 \mid G=1, A)
-
\mathbb{P}(\hat y = 1 \mid G=0, A)
\\
&=
\mathbb{E}[\sigma(h(Z)) \mid G=1, A]
-
\mathbb{E}[\sigma(h(Z)) \mid G=0, A],
\end{align}
where we used
$\mathbb{P}(\hat y = 1 \mid G=g, A)
= \mathbb{E}[\sigma(h(Z)) \mid G=g, A]$.
We now add and subtract the group mean scores passed through $\sigma$:
\begin{align}
\Delta_A(w)
&=
\Bigl(
  \mathbb{E}[\sigma(h(Z)) \mid G=1, A]
  - \sigma(\bar h_{1,A}(w))
\Bigr)
\\
&\quad
+ \Bigl(
  \sigma(\bar h_{1,A}(w)) - \sigma(\bar h_{0,A}(w))
\Bigr)
\\
&\quad
+ \Bigl(
  \sigma(\bar h_{0,A}(w))
  - \mathbb{E}[\sigma(h(Z)) \mid G=0, A]
\Bigr).
\end{align}
Taking the absolute values and applying the triangle inequality gives
\begin{align}
\bigl|\Delta_A(w)\bigr|
&\le
\bigl|
  \mathbb{E}[\sigma(h(Z)) \mid G=1, A]
  - \sigma(\bar h_{1,A}(w))
\bigr|
\label{eqn:first_term_A}
\\
&\quad
+
\bigl|
  \sigma(\bar h_{1,A}(w)) - \sigma(\bar h_{0,A}(w))
\bigr|
\label{eqn:middle_term_A}
\\
&\quad
+
\bigl|
  \sigma(\bar h_{0,A}(w))
  - \mathbb{E}[\sigma(h(Z)) \mid G=0, A]
\bigr|.
\label{eqn:last_term_A}
\end{align}

For the middle term \eqref{eqn:middle_term_A}, Lipschitz continuity of
$\sigma$ implies
\begin{equation}
\bigl|\sigma(\bar h_{1,A}(w)) - \sigma(\bar h_{0,A}(w))\bigr|
\;\le\;
L\,\bigl|\bar h_{1,A}(w) - \bar h_{0,A}(w)\bigr|
=
L\,\bigl|\Delta_{g,A}(w)\bigr|.
\end{equation}
For the first term \eqref{eqn:first_term_A}, Jensen's inequality and Lipschitz continuity of $\sigma$ give
\begin{align}
\bigl|
  \mathbb{E}[\sigma(h(Z)) \mid G=1, A]
  - \sigma(\bar h_{1,A}(w))
\bigr|
&\le
\mathbb{E}\bigl[
  \bigl|\sigma(h(Z)) - \sigma(\bar h_{1,A}(w))\bigr|
\bigm|\,
  G=1, A
\bigr]
\\
&\le
L\,\mathbb{E}\bigl[
  \bigl|h(Z) - \bar h_{1,A}(w)\bigr|
\bigm|\,
  G=1, A
\bigr]
=
L\,v_{1,A}(w),
\end{align}
and similarly
\begin{equation}
\bigl|
  \sigma(\bar h_{0,A}(w))
  - \mathbb{E}[\sigma(h(Z)) \mid G=0, A]
\bigr|
\;\le\;
L\,v_{0,A}(w).
\end{equation}

Combining these three bounds
\eqref{eqn:first_term_A}–\eqref{eqn:last_term_A}
yields
\begin{equation}
\bigl|\Delta_A(w)\bigr|
\;\le\;
L\,\bigl|\Delta_{g,A}(w)\bigr|
\;+\;
L\bigl(v_{1,A}(w)+v_{0,A}(w)\bigr),
\end{equation}
which is Eq. \eqref{eqn:rate_bound}.
\end{proof}


\begin{proof}[Proof of Corollary~\ref{cor:conditional_ratio_bound}]
First, take a fixed event $A$ with $\mathbb{P}(A)>0$ and a probe $w\in\mathcal{H}$.
Apply the argument of Lemma~\ref{lem:rate-bound} to the conditional
distribution of $(Z,G)$ given $A$ by replacing all probabilities
and expectations there by conditional ones given $A$. Lemma~\ref{lem:rate-bound} applied to the conditional distribution given $A$ yields
\begin{equation}
\bigl|\Delta_A(w)\bigr|
\;\le\;
L\,\bigl|\Delta_{g,A}(w)\bigr|
\;+\;
L\bigl(v_{0,A}(w)+v_{1,A}(w)\bigr).
\end{equation}

By definition,
$
p_{g,A}(w)
=\mathbb{P}(\hat y = 1 \mid G=g,A)$ and
$R_A(w)
=\frac{p_{0,A}(w)}{p_{1,A}(w)},
$
so that
\begin{equation}
R_A(w)-1
=
\frac{p_{0,A}(w)-p_{1,A}(w)}{p_{1,A}(w)}
=
-\frac{p_{1,A}(w)-p_{0,A}(w)}{p_{1,A}(w)}
=
-\frac{\Delta_A(w)}{p_{1,A}(w)}.
\end{equation}
Using the assumption $p_{1,A}(w)\ge p_{\min,A}>0$ for all $w\in\mathcal{H}$,
we obtain
\begin{align}
\bigl|R_A(w)-1\bigr|
=
\frac{\bigl|\Delta_A(w)\bigr|}{p_{1,A}(w)}
&\le
\frac{L}{p_{\min,A}}\,
\bigl|\Delta_{g,A}(w)\bigr|
+
\frac{L}{p_{\min,A}}\,
\bigl(v_{0,A}(w)+v_{1,A}(w)\bigr),
\end{align}
which is Eq. \eqref{eq:ratio_explicit_bound_A}.

\end{proof}


\begin{proof}[Proof of Lemma~\ref{lem:fw_inv_mean_gap}]
We work under the conditional law given $A$ (so all expectations and probabilities
below are taken with respect to $(Z,G)\mid A$). Let $(Z',G')$ be an independent copy
of $(Z,G)$ under this conditional law. Expanding the squared norm yields
\begin{align}
I^{\methodsuperscript}_{\mathrm{reg},A}&=\mathbb{E}\big[\|Z-Z'\|_2^2 \mid A\big] \\
&=\mathbb{E}\big[\|Z\|_2^2 \mid A\big]+\mathbb{E}\big[\|Z'\|_2^2 \mid A\big]
-2\,\mathbb{E}\big[Z^\top Z' \mid A\big].
\end{align}
Since $Z'$ has the same conditional distribution as $Z$ given $A$,
$\mathbb{E}[\|Z'\|_2^2\mid A]=\mathbb{E}[\|Z\|_2^2\mid A]$. Moreover, conditional
independence implies
\begin{equation}
\mathbb{E}[Z^\top Z' \mid A]
=\big(\mathbb{E}[Z\mid A]\big)^\top\big(\mathbb{E}[Z'\mid A]\big)
=\big\|\mathbb{E}[Z\mid A]\big\|_2^2.
\end{equation}
Therefore,
\begin{align}
I^{\methodsuperscript}_{\mathrm{reg},A}
&=2\Big(\mathbb{E}\big[\|Z\|_2^2\mid A\big]-\big\|\mathbb{E}[Z\mid A]\big\|_2^2\Big)
=2\,\mathrm{tr}\!\big(\mathrm{Cov}(Z\mid A)\big).
\end{align}
Using the law of total covariance under $A$,
\begin{equation}
\mathrm{Cov}(Z\mid A)
=\mathbb{E}\big[\mathrm{Cov}(Z\mid G,A)\mid A\big]
+\mathrm{Cov}\big(\mathbb{E}[Z\mid G,A]\mid A\big)
=\sum_{g\in\{0,1\}}\pi_{g,A}\,\Sigma_{g,A}
+\mathrm{Cov}(\mu_{G,A}\mid A),
\end{equation}
where $\mu_{G,A}$ is the random vector that equals $\mu_{0,A}$ with probability $\pi_{0,A}$, and equals $\mu_{1,A}$ with probability $\pi_{1,A}$ under the conditional law given $A$. Since $G$ is binary,
\begin{equation}
\mathrm{Cov}(\mu_{G,A}\mid A)
=\pi_{0,A}\pi_{1,A}\,(\mu_{1,A}-\mu_{0,A})(\mu_{1,A}-\mu_{0,A})^\top,
\qquad
\mathrm{tr}\!\big(\mathrm{Cov}(\mu_{G,A}\mid A)\big)
=\pi_{0,A}\pi_{1,A}\,\|\mu_{1,A}-\mu_{0,A}\|_2^2.
\end{equation}
Hence,
\begin{align}
I^{\methodsuperscript}_{\mathrm{reg},A}
&=2\,\mathrm{tr}\!\big(\mathrm{Cov}(Z\mid A)\big) \\
&=2\sum_{g\in\{0,1\}}\pi_{g,A}\,\mathrm{tr}(\Sigma_{g,A})
\;+\;2\,\pi_{0,A}\pi_{1,A}\,\|\mu_{1,A}-\mu_{0,A}\|_2^2 \\
&\ge 2\,\pi_{0,A}\pi_{1,A}\,\|\mu_{1,A}-\mu_{0,A}\|_2^2,
\end{align}
which implies
\begin{equation}
\|\mu_{1,A}-\mu_{0,A}\|_2^2
\le \frac{I^{\methodsuperscript}_{\mathrm{reg},A}}{2\pi_{0,A}\pi_{1,A}}.
\end{equation}
Taking the square roots then gives the equivalent implication stated in the lemma.
\end{proof}


\begin{proof}[Proof of Lemma~\ref{lem:two_sided_bound}]
Following the typical linear-probe evaluation protocols, where downstream
classifiers are trained with standard $\ell_2$ regularization (weight decay)
or early stopping, the bounded-norm assumption $\|w\|_2 \le C$ is trivially
satisfied, since the regularizer (or implicit regularization) prevents the
probe weights from growing without bound (e.g., with an $\ell_2$
penalty $\frac{\lambda}{2}\|w\|_2^2$ one can take
$C = \sqrt{2\,\mathcal{L}(0)/\lambda}$). Thus, we take $w$ with $\|w\|_2=C>0$, and for each $g\in\{0,1\}$, we define
$\mu_{g,A}:=\mathbb{E}[Z\mid G=g,A]$ and $\Sigma_{g,A}:=\mathrm{Cov}(Z\mid G=g,A)$.

\paragraph{Upper bound.}
For each $g\in\{0,1\}$ define $X_{g,A}:=w^\top Z-\mathbb{E}[w^\top Z\mid G=g,A]=w^\top(Z-\mu_{g,A})$.
By Cauchy--Schwarz,
\begin{align}
v_{g,A}(w)
=\mathbb{E}\big[|X_{g,A}|\mid G=g,A\big]
&\le \sqrt{\mathbb{E}\big[X_{g,A}^2\mid G=g,A\big]}
= \sqrt{\mathrm{Var}(w^\top Z\mid G=g,A)}
= \sqrt{w^\top \Sigma_{g,A} w}.
\end{align}
Using $w^\top \Sigma_{g,A} w\le \|w\|_2^2\,\mathrm{tr}(\Sigma_{g,A})$ for any PSD matrix $\Sigma_{g,A}$, $v_{g,A}(w)\le \|w\|_2\,\sqrt{\mathrm{tr}(\Sigma_{g,A})}=C\sqrt{\mathrm{tr}(\Sigma_{g,A})}.$
Next, with an independent copy $(Z',G')$ of $(Z,G)$,
\begin{eqnarray}
I^{\methodsuperscript}_{\mathrm{reg},A} & = & \mathbb{E}\big[\|Z-Z'\|_2^2 \mid A\big]
=2\,\mathrm{tr}\!\big(\mathrm{Cov}(Z\mid A)\big) \\
& = & 2\sum_{h\in\{0,1\}}\pi_{h,A}\,\mathrm{tr}(\Sigma_{h,A})
+2\pi_{0,A}\pi_{1,A}\|\mu_{1,A}-\mu_{0,A}\|_2^2
\ge 2\sum_{h\in\{0,1\}}\pi_{h,A}\,\mathrm{tr}(\Sigma_{h,A}).
\end{eqnarray}
Therefore, for each $g$, $
\pi_{g,A}\,\mathrm{tr}(\Sigma_{g,A})
\le \sum_{h\in\{0,1\}}\pi_{h,A}\,\mathrm{tr}(\Sigma_{h,A})
\le \frac{I^{\methodsuperscript}_{\mathrm{reg},A}}{2},
\quad\text{hence}\quad
\mathrm{tr}(\Sigma_{g,A})\le \frac{I^{\methodsuperscript}_{\mathrm{reg},A}}{2\pi_{g,A}}.
$
Plugging into the bound on $v_{g,A}(w)$ then gives $
v_{g,A}(w)\le C\sqrt{\frac{I^{\methodsuperscript}_{\mathrm{reg},A}}{2\pi_{g,A}}}.
$
Summing over $g=0,1$ results to the upper bound
\begin{equation}
v_{0,A}(w)+v_{1,A}(w)
\le C\sqrt{\frac{I^{\methodsuperscript}_{\mathrm{reg},A}}{2}}\left(\frac{1}{\sqrt{\pi_{0,A}}}+\frac{1}{\sqrt{\pi_{1,A}}}\right)
\le C\sqrt{\frac{2I^{\methodsuperscript}_{\mathrm{reg},A}}{\pi_{\min,A}}}.
\end{equation}
Since $\frac{1}{\sqrt{\pi_{0,A}}}+\frac{1}{\sqrt{\pi_{1,A}}}\le \frac{2}{\sqrt{\pi_{\min,A}}}$.

\paragraph{Lower bound.}
Assume $V_{\mathrm{reg},A}^{(g)}=0$ and $C_{\mathrm{reg},A}^{(g)}\le\delta$ for both groups.
From $V_{\mathrm{reg},A}^{(g)}=0$ we have $\mathrm{Std}(Z_j\mid G=g,A)\ge \gamma$ for every $j$,
so
\begin{equation}
(\Sigma_{g,A})_{jj}=\mathrm{Var}(Z_j\mid G=g,A)\ge \gamma^2,\qquad j=1,\dots,d.
\end{equation}
From $C_{\mathrm{reg},A}^{(g)}\le\delta$ we obtain
$
\sum_{j\neq k}(\Sigma_{g,A})_{jk}^2 \le d\,\delta.
$ Then, fix any row index $j$. By Cauchy--Schwarz,
\begin{align}
\sum_{k\neq j}|(\Sigma_{g,A})_{jk}|
&\le \sqrt{d-1}\,\sqrt{\sum_{k\neq j}(\Sigma_{g,A})_{jk}^2}
\le \sqrt{d-1}\,\sqrt{\sum_{a\neq b}(\Sigma_{g,A})_{ab}^2}
\le \sqrt{d-1}\,\sqrt{d\,\delta}
= \sqrt{d(d-1)\delta}.
\end{align}
By the Gershgorin circle theorem,
\begin{equation}
\lambda_{\min}(\Sigma_{g,A})
\ge \min_{j}\left((\Sigma_{g,A})_{jj}-\sum_{k\neq j}|(\Sigma_{g,A})_{jk}|\right)
\ge \gamma^2-\sqrt{d(d-1)\delta}.
\end{equation}
Hence, since $\|w\|_2=C$,
\begin{equation}
\mathrm{Var}(w^\top Z\mid G=g,A)=w^\top\Sigma_{g,A} w \ge \|w\|_2^2\,\lambda_{\min}(\Sigma_{g,A})
\ge C^2\Big(\gamma^2-\sqrt{d(d-1)\delta}\Big).
\end{equation}
Now use the additional assumption $\|Z\|_2\le M_A$ almost surely under $A$. Then
$\|\mu_{g,A}\|_2=\|\mathbb{E}[Z\mid G=g,A]\|_2\le \mathbb{E}[\|Z\|_2\mid G=g,A]\le M_A$,
so $\|Z-\mu_{g,A}\|_2\le 2M_A$ almost surely, and therefore
\begin{equation}
|X_{g,A}|=|w^\top(Z-\mu_{g,A})|\le \|w\|_2\,\|Z-\mu_{g,A}\|_2\le 2CM_A\qquad\text{a.s.}
\end{equation}
Since $|X_{g,A}|^2\le (2CM_A)|X_{g,A}|$ almost surely, taking conditional expectations yields
\begin{equation}
\mathbb{E}[X_{g,A}^2\mid G=g,A]
\le (2CM_A)\,\mathbb{E}[|X_{g,A}|\mid G=g,A]=(2CM_A)\,v_{g,A}(w),
\end{equation}
and thus
\begin{equation}
v_{g,A}(w)\ge \frac{\mathbb{E}[X_{g,A}^2\mid G=g,A]}{2CM_A}
=\frac{\mathrm{Var}(w^\top Z\mid G=g,A)}{2CM_A}
\ge \frac{C}{2M_A}\Big(\gamma^2-\sqrt{d(d-1)\delta}\Big).
\end{equation}
Summing the two groups and applying $(\cdot)_+$ gives
\begin{equation}
v_{0,A}(w)+v_{1,A}(w)\ge \frac{C}{M_A}\Big(\gamma^2-\sqrt{d(d-1)\delta}\Big)_+.
\end{equation}
Combining this with the upper bound completes the proof.
\end{proof}


\section{Additional Results}
\label{sect:additional_results}

\subsection{DVlog}
\label{sect:add_results_dvlog}

In Table \ref{tab:supp_dvlog_combined}, we observe that DVlog shows a clear split between audio-only and visual-only models. For audio, DeiT attains the top classification scores (Acc 0.68, F1 0.74) and the highest overall fairness aggregate ($AGG_F$ 0.70). DenseNet-121 follows with a lower Acc of 0.60 but with a $AGG_F$ of 0.69, indicating stronger faieness metrics (notably $\mathcal{M}{\text{SP}}$ 0.62). Classify-Net offers balanced performance and fairness (Acc 0.65, $AGG_F$ 0.68) whereas MSCDR shows issues in fairness ($AGG_F$ 0.34). The Uni Audio gives moderate Acc 0.61 and an $AGG_F$ of 0.56. For visuals, FacialPulse matches the stronger audio baselines in fairness ($AGG_F$ 0.69) despite lower recall, while ST-GCN’s disparity (infinite $\mathcal{M}{\text{EOdd}}$) makes its $AGG_F$ undefined. Uni Visual improves recall (0.73) over other visual methods but does so at a cost to fairness ($AGG_F$ 0.48).

In Table \ref{tab:supp_dvlog_multimodal} we observe recall and precision separate the models into two clear groups. Recall-oriented systems include \methodname M1, which attains perfect recall (1.00) but keeps precision modest at 0.58, and SEResnet, whose recall 0.94 likewise comes with precision 0.58. These models prioritise capturing nearly all positives at the cost of false positives. Precision-leaning transformers, Bimod-cross (0.70) and Perceiver (0.69), deliver the highest precision while still recovering around two-thirds of the targets (recall 0.74 and 0.64, respectively), giving the most balanced retrieval of relevant instances. CNN baselines (Xception variants) are in the mid-range on both metrics, whereas our \methodname variants (M2–M4) mostly trade recall for slight precision gains. 

In addition, Table \ref{tab:supp_dvlog_ssl} shows that precision is high in multiple baselines (e.g., DeCUR-Baseline 0.74, Ours-Baseline 0.74, CoMM-Baseline 0.73) and they consistently show the highest precision but pair it with mid-range recall. Meaning that they capture fewer positives while keeping false positives low. In contrast, \methodname-M1 reaches perfect recall 1.00 with a marked drop in precision 0.58, with Quest-M2 (0.97) and Quest-M3 (0.87) showing a similar pattern. Double-pooling amplifies this trend. M3+DP lifts recall to 0.94 at precision 0.64, while M2+DP achieves 0.86/0.62.


\begin{table}[hbt!]
  \caption{\textbf{MODMA:} Combined EEG and Audio results.
  Best and second-best results are highlighted in \first{bold} and \second{underline}, respectively. * denotes baseline.
  }
  \small
  \addtolength{\tabcolsep}{-1.8mm}
  \centering
  \begin{tabular}{l|l|cccc|cccc|c}
    \hline
      && \multicolumn{4}{c|}{\textbf{Performance}}
      & \multicolumn{5}{c}{\textbf{Group Fairness}}
       \\
    & 
      & Acc. & Prec. & Rec. & F1 
      & $SP$ & $EOpp$  & $EOdd$  & $EAcc$  & $AGG_F$ \\
    \hline

\multirow{5}{*}{\rot{\textbf{EEG}}}
& DeprNet      & 0.69 & \second{0.81} & 0.42 & 0.51 
& 0.16 & 0.12 & 0.00 & \first{0.94} & 0.31 \\

& EEGNet       & 0.70 & 0.80 & 0.50 & 0.54 
& 0.11 & 0.08 & 0.00 & 0.77 & 0.24 \\
& FeatureNet   & \first{0.86} & \first{1.00} & \first{0.67} & \first{0.80}
& \second{1.33} & \first{1.00} & 0.00 & 1.33 & 0.59 \\

& HybGNN IA    & \second{0.85} & \first{1.00} & \second{0.65} & \second{0.79} 
& 0.00 & 0.00 & 0.00 & 0.68 & 0.17 \\

& FeatNet*      & 0.73 & \first{1.00} & 0.37 & 0.54 
& 0.00 & 0.00 & 0.00 & \second{0.86} & 0.22 \\
\hline
\multirow{5}{*}{\rot{\textbf{Audio}}}
& MSCDR         & 0.57 & 0.50 & 0.43 & 0.46 
& 0.18 & 0.30 & \second{1.19} & 0.62 & 0.38 \\

& Classify-Net  & 0.70 & 0.76 & 0.43 & 0.55 
& \first{0.73} & 1.71 & 1.28 & 0.37 & \first{0.63} \\
\cdashline{3-10}[.4pt/2pt]
& DeiT          & 0.73 & 0.76 & 0.54 & 0.62 
& 1.34 & 3.00 & \first{1.17} & 0.83 & 0.17 \\

& DenseNet-121  & 0.61 & 0.64 & 0.23 & 0.34 
& 0.57 & \second{1.50} & 1.40 & 0.39 & \second{0.61} \\

& FeatNet*       & 0.69 & 0.74 & 0.61 & 0.67 
& 0.59 & 1.78 & 0.59 & 0.45 & 0.55 \\
\hline
  \end{tabular}
  \label{tab:supp_modma_combined}
\end{table}

\begin{table}[hbt!]
\caption{\textbf{MODMA:} 
Baseline comparison of the performance and fairness across different \textbf{multimodal} methods. Best and second-best results are highlighted in \first{bold} and \second{underline}, respectively. * denotes baseline.
}
\small
\addtolength{\tabcolsep}{-1.8mm}
  \centering
  \begin{tabular}{l|l|cccc|cccc|c}
    \hline
 & \multirow{2}{*}{\textbf{Method}} & \multicolumn{4}{c|}{\textbf{Performance}} &\multicolumn{5}{c}{\textbf{Group Fairness}}   \\
     
     &     & Acc.	&Prec.	&Rec.	&F1	& $SP$ & $EOpp$  & $EOdd$  & $EAcc$ &$AGG_F$ \\ \cline{1-11}
     
\multirow{3}{*}{\rot{\textbf{CNN}}}
& MultiDepr       & \first{0.79} & \first{1.00} & 0.50 & \first{0.67} & 0.00 & 0.00 & 0.00 & 0.76 & 0.19 \\
& Effnetv2s       & \second{0.71} & \first{1.00} & 0.33 & 0.50 & 0.00 & 0.00 & 0.00 & 0.88 & 0.22 \\
& FeatNet* & \second{0.71} & 0.67 & 0.67 & \first{0.67} & 0.67 & 0.50 & 0.00 & 0.33 & 0.38 \\
\cdashline{1-11}[.4pt/2pt]
& EMO GCN         & 0.60 & 0.52 & \first{0.93} & \first{0.67} & 0.78 & 0.59 & 0.86 & 0.82 & 0.76 \\
\cdashline{2-11}[.4pt/2pt]
& Transformer     & 0.54 & \second{0.80} & 0.20 & 0.57 & 0.86 & 0.60 & 1.43 & 0.62 & 0.66 \\
\cdashline{1-11}[.4pt/2pt]
\multirow{4}{*}{\rot{\textbf{Ours}}}
& M1 & 0.59 & 0.50 & 0.67 & 0.57 & \first{0.95} & \first{0.71} & \first{1.00} & 0.95 & \first{0.90} \\
& M2 & 0.67 & 0.60 & 0.66 & 0.63 & \second{0.89} & \second{0.66} & \second{1.03} & \second{1.01} & \second{0.88} \\
& M3 & 0.60 & 0.56 & 0.34 & 0.42 & 0.67 & 0.50 & \first{1.00} & \first{1.00} & 0.79 \\
& M4 & 0.65 & 0.56 & \second{0.79} & \second{0.66} & 0.85 & 0.64 & 1.08 & 1.10 & 0.83 \\
\hline
\cdashline{2-11}[.4pt/2pt]
    \hline    
    \end{tabular}
    \label{tab:supp_modma_multimodal}
\end{table}


\begin{table}[hbt!]
\caption{\textbf{MODMA:} SSL.
Baseline comparison of the performance and fairness across different \textbf{SSL} methods. Best and second-best results are highlighted in \first{bold} and \second{underline}, respectively.
}
\small
\addtolength{\tabcolsep}{-1.8mm}
\centering
\begin{tabular}{l|l|cccc|cccc|c}
\hline
& \multirow{2}{*}{\textbf{Method}}
& \multicolumn{4}{c|}{\textbf{Performance}}
& \multicolumn{5}{c}{\textbf{Group Fairness}}  \\[-2pt]
& & Acc. & Prec. & Rec. & F1
& $SP$ & $EOpp$  & $EOdd$  & $EAcc$ &$AGG_F$ \\ \cline{1-11}
%
 \multirow{5}{*}{\rot{\textbf{CoMM}}}  
& Baseline   & 0.58 & 0.54 & 0.09 & 0.14 & 0.83 & 0.63 & 4.00 & 1.11 & 0.09 \\
& M1         & 0.57 & 0.50 & 0.22 & 0.31 & 0.44 & 0.33 & 1.00 & 0.95 & 0.68 \\
& M2  &0.53	&0.42&	0.34	&0.38	&1.06&	0.75&	1.00&	1.11	&0.89 \\
& M3  &0.69&	0.97&	0.33&	0.50&	0.67&	0.50&	0.00&	1.17&	0.50  \\
& M4 &0.53&	0.43&	0.33&	0.38	&0.97	&0.74&	0.95	&1.12&	0.88  \\
\cline{1-11}
 \multirow{5}{*}{\rot{\textbf{FACTORCL}}}    
& Baseline   & 0.66 & 0.60 & 0.42 & 0.50 & 1.21 & \second{0.91} & 2.00 & 1.39 & 0.58 \\
& M1         & \second{0.67} & 0.67 & \second{0.67} & \first{0.67} & 1.07 & 0.80 & 2.00 & 0.89 & 0.66 \\
& M2  &\second{0.67}&	0.67&	0.44	&0.53	&0.27	&0.20	&1.00	&0.74&	0.55 \\
& M3         & 0.57 & 0.50 & 0.51 & 0.57 & 0.67 & 0.50 & \second{0.67} & 1.33 & 0.63 \\
& M4         & 0.66 & 0.64 & 0.52 & 0.57 & 0.43 & 0.33 & \second{0.67} & 0.88 & 0.58 \\
\cline{1-11}
%
 \multirow{5}{*}{\rot{\textbf{QUEST}}}   
& Baseline   & 0.62 & 0.60 & 0.42 & 0.50 & 1.21 & 0.88 & 2.00 & 1.39 & 0.57 \\
& M1         & 0.62 & 0.60 & 0.33 & 0.43 & 0.89 & 0.67 & \first{1.00} & 1.14 & 0.85 \\
& M2         & 0.60 & 0.57 & 0.30 & 0.39 & \first{0.99} & 0.74 & 1.18 & 1.19 & 0.84 \\
& M3         & 0.58 & 0.50 & 0.33 & 0.40 & 0.67 & 0.50 & \first{1.00} & 1.33 & 0.71 \\
& M4         & 0.57 & 0.50 & 0.22 & 0.31 & 1.33 & \first{1.00} & 2.00 & 1.33 & 0.58 \\
\cline{1-11}
%
 \multirow{5}{*}{\rot{\textbf{DeCUR}}}  
& Baseline   & \second{0.67} & \second{0.75} & 0.33 & 0.46 & 0.44 & 0.33 & \first{1.00} & 1.33 & 0.61 \\
& M1         & \second{0.67} & 0.63 & 0.56 & 0.59 & 0.44 & 0.43 & 0.50 & \first{1.00} & 0.59 \\
& M2         & 0.65 & 0.70 & 0.33 & 0.45 & 0.58 & 0.43 & \first{1.00} & 1.28 & 0.68 \\
& M3         & 0.62 & 0.56 & 0.56 & 0.56 & 0.67 & 0.50 & \first{1.00} & 0.83 & 0.75 \\
& M4         & \first{0.69} & \first{0.77} & 0.43 & 0.55 & 0.35 & 0.25 & 0.00 & 0.67 & 0.32 \\
\cline{1-11}
%
 \multirow{5}{*}{\rot{\textbf{Ours}}}  
& M1 & 0.59 & 0.50 & \second{0.67} & 0.57 & \second{0.95} & 0.71 & \first{1.00} & 0.95 & \first{0.90} \\
& M2 & \second{0.67} & 0.60 & 0.66 & 0.63 & 0.89 & 0.66 & \second{1.03} & \second{1.01} & \second{0.88} \\
& M3 & 0.60 & 0.56 & 0.34 & 0.42 & 0.67 & 0.50 & \first{1.00} & \first{1.00} & 0.79 \\
& M4 & 0.65 & 0.56 & \first{0.79} & \second{0.66} & 0.85 & 0.64 & 1.08 & 1.10 & 0.83 \\
\hline
\end{tabular}
\label{tab:supp_modma_ssl}
\end{table}

\subsection{MIMIC-III}
\label{sect:add_results_mimic}

Refering to Table \ref{tab:supp_MIMIC_ssl_auroc}, we observe SSL methods finetuned on AUROC exhibit clear trade-offs between accuracy and fairness. 
CoMM-M2 slightly reduces accuracy relative to the baseline but achieves the highest aggregate fairness. 
DeCUR-M2 maintains baseline accuracy (0.87) while boosting $AGG_F$ to 0.79 from 0.54. FOCAL-M4 also preserves the accuracy (0.87) and raises $AGG_F$ to 0.89. FACTORCL-M2 sets the fairness benchmark with $AGG_F$ 0.94 at 0.86 accuracy, while FACTORCL-M3 reaches the top accuracy (0.89) and F1 (0.30) but with a lower $AGG_F$ (0.84). SimCLR’s baseline shows infinite $EOdd$ and undefined $AGG_F$, yet SimCLR-M2 balances accuracy and fairness together. 

In Table \ref{tab:supp_MIMIC_ssl_accuracy}, every method displays modest precision, never exceeding 0.34. SimCLR-baseline (0.32) and our M4 (0.34) show the best results in this column. Most other runs, including the classical baselines for CoMM, FOCAL and FACTORCL, cluster around 0.16–0.19, showing only marginal improvement over the lowest performance models.  Recall, in contrast, shows a wider spread of values. CoMM-baseline is lowest at 0.10, while \methodname-M2 has the highest recall of 0.57. Among the baseline models, DeCUR-baseline (0.36) and SimCLR-baseline (0.35) have the highest recall.

\subsection{MODMA}
\label{sect:add_results_modma}

With reference to Table \ref{tab:supp_modma_combined}, MODMA shows a clear modality split. EEG models reach strong classification, FeatureNet shoes the best Acc 0.86 and F1 0.80, but every EEG method records $\infty$ on $EOdd$, so $AGG_F$ is undefined and group fairness remains hard to compare.
Audio models on the other hand have finite fairness which are more direct to analyze. DeiT leads accuracy (Acc 0.73, F1 0.62) but shows the weakest fairness ($AGG_F$ 0.17), while Classify-Net balances performance (Acc 0.70) with the top $AGG_F$ 0.63. MSCDR trades accuracy for fairness, and FeatNet shows the highest F1 among audio models at a mid-level $AGG_F$. 

In Table \ref{tab:supp_modma_ssl}, we see that, for most cases, \methodname+SSL models on MODMA reveal consistent gains in fairness over their baselines, with M4 showing slight fairness decrease in DeCUR. For CoMM, M1 leaves accuracy almost unchanged but improves the $AGG_F$. FACTORCL shows its best balance at M1. For QUEST, both M1 and M2 keep baseline accuracy yet drive $AGG_F$ up to 0.85 / 0.84. DeCUR-M3 also has relatively good accuracy with the top fairness in its block ($AGG_F$ 0.75), while M4 gains a higher accuracy (0.69) but shows a zero $EOdd$, making $AGG_F$ the lowest in DeCUR's variants. Finally, our methods give the strongest overall equity: M1 reaches the highest reported $AGG_F$ (0.90) alongside good recall (0.67), and M2 has the joint-best accuracy (0.67) with an $AGG_F$ of 0.88.

\textbf{Computational overhead.}
Algorithmically, \methodname does not introduce a new heavy model component. 
To clarify the computational overhead, we measured peak GPU memory and wall-clock training time under matched settings as shown in Table \ref{tab:computation_overhead}.
\methodname remains in essentially the same efficiency regime as VICReg: across M1–M4, peak **memory is unchanged, and total training time increases by only 0.12–0.27\% relative to VICReg. This is consistent with the design of \methodname, which mainly modifies the training objective through subject-aware pooling and alignment rather than introducing a substantially larger model.
%

\begin{table}[hbt!]
\caption{Computational overhead measured on MODMA under 
10 SSL epochs, 5 probe epochs, and 2 subjects per batch. \label{tab:computation_overhead}}
\footnotesize
\addtolength{\tabcolsep}{-1.4mm}
  \centering
\begin{tabular}{lccccc}
\toprule
\textbf{Method} & \textbf{Params (M)} & \textbf{Peak Mem (GB)} & \textbf{Total Time (s)} & \textbf{Time vs VICReg} & \textbf{Mem vs VICReg} \\
\midrule
COMM       & 3.89 & 18.28 & 99.78 & +24.66\% & +73.08\% \\
QUEST      & 4.94 & 10.56 & 79.96 & -0.10\%  & +0.00\%  \\
CLIP       & 3.89 & 10.54 & 79.52 & -0.65\%  & -0.19\%  \\
FACTORCL   & 4.94 & 18.29 & 98.62 & +23.21\% & +73.21\% \\
DECUR      & 4.94 & 18.29 & 98.24 & +22.74\% & +73.21\% \\
VICReg     & 4.94 & 10.56 & 80.04 & +0.00\%  & +0.00\%  \\
FairSSL-M1 & 3.89 & 10.55 & 80.14 & +0.12\%  & -0.07\%  \\
FairSSL-M2 & 4.94 & 10.56 & 80.20 & +0.20\%  & +0.00\%  \\
FairSSL-M3 & 4.94 & 10.56 & 80.25 & +0.27\%  & +0.00\%  \\
FairSSL-M4 & 4.94 & 10.56 & 80.24 & +0.25\%  & +0.00\%  \\
\bottomrule
\end{tabular}
\end{table}


\section{Hyperparameters}
\label{sect:hyperparameters}

Tables \ref{tab:dvlog_hparams} (DVlog), 
\ref{tab:mimic_hparams} (MIMIC-III)
and \ref{tab:modma_hparams} (MODMA)
display the hyper-paramater settings of the different models used in our experiments across the different datasets.  All values in these tables have been rounded to three decimal places for readability; the full-precision settings are available in the code repository shared within the main paper.

\begin{table*}[htb!]
\caption{\textbf{DVlog:} Complete hyper-parameter listings for the different models. These values are obtained by Bayesian optimization over all the given hyperparameters with the purpose to maximize the model performance. For further details, see Sections S1 and S2.
Abbreviations: DP: double pooling.
\label{tab:dvlog_hparams}}
\centering
\scriptsize
\setlength{\tabcolsep}{3pt}
\begin{tabular}{p{0.08\linewidth} | p{0.05\linewidth} | p{0.75\linewidth}}
\toprule
 \textbf{Model} & \textbf{Variant} & \textbf{Hyper-parameters} \\ \midrule

CoMM & Baseline & batch\_size = 256, ssl\_epochs = 20, cov\_coeff = 15, linear\_lr = {0.691}, lr = {0.057}, mi\_coeff = 7, mlp = 512-256-128, sl\_epochs = 20, tau = {0.682}, var\_coeff = 11, wd = {0.149} \\

CoMM & M1 &batch\_size = 256, ssl\_epochs = 10, cov\_coeff = 13, linear\_lr = {0.821}, lr = {0.977}, mi\_coeff = 18, mlp = 512-256-128, sl\_epochs = 10, tau = {0.800}, var\_coeff = 7, wd = {0.611} \\

CoMM & M2 & batch\_size = 64, ssl\_epochs = 19, cov\_coeff = {8}, linear\_lr = {0.486}, lr = {0.033}, mi\_coeff = 15, mlp = 512-256-128, sl\_epochs = 20, tau = {0.194}, var\_coeff = 19, wd = {0.894} \\

CoMM & M3 & batch\_size = 128, ssl\_epochs = 25, cov\_coeff = {3}, linear\_lr = {0.910}, lr = {0.113}, mi\_coeff = 9, mlp = 512-256-128, sl\_epochs = 9, tau = {0.987}, var\_coeff = 12, wd = {0.718} \\

CoMM & M4 & batch\_size = 256, ssl\_epochs = 22, cov\_coeff = {4}, linear\_lr = {0.982}, lr = {0.046}, mi\_coeff = 14, mlp = 512-256-128, sl\_epochs = 20, tau = {0.935}, var\_coeff = 8, wd = {0.308} \\

\cline{1-3}

QUEST &Baseline   &  align\_coeff = 13, batch\_size = 64, linear\_lr = {0.010}, lr = {0.020}, mlp = 512-256-128, orth\_coeff = 6, sl\_epochs = 10, sp\_coeff = 0, uniq\_coeff = 14, var\_coeff = 14, ssl\_epochs = 10, wd = {0.715} \\

QUEST &M1 & align\_coeff = 0, batch\_size = 64, linear\_lr = {0.398}, lr = {0.645}, mlp = 512-256-128, orth\_coeff = 2, sl\_epochs = 15, sp\_coeff = 10, uniq\_coeff = 5, var\_coeff = 7, ssl\_epochs = 20, wd = {0.946} \\

QUEST & M2   & align\_coeff = 16, batch\_size = 256, linear\_lr = {0.173}, lr = {0.232}, mlp = 512-256-128, orth\_coeff = 17, sl\_epochs = 15, sp\_coeff = 7, uniq\_coeff = 16, var\_coeff = 18, ssl\_epochs = 10, wd = {0.435} \\

QUEST &M3   &  align\_coeff = 7, batch\_size = 64, linear\_lr = {0.447}, lr = {0.135}, mlp = 512-256-128, orth\_coeff = 16, sl\_epochs = 10, sp\_coeff = 8, uniq\_coeff = 11, var\_coeff = 12, ssl\_epochs = 20, wd = {0.758} \\

QUEST &M4   &  align\_coeff = 3, batch\_size = 256, linear\_lr = {0.548}, lr = {0.242}, mlp = 512-256-128, orth\_coeff = 14, sl\_epochs = 15, sp\_coeff = 13, uniq\_coeff = 8, var\_coeff = 14, ssl\_epochs = 10, wd = {0.434} \\

\cline{1-3}

DeCUR & Baseline & batch\_size = 128, common\_coeff = {4}, linear\_lr = {0.329}, lr = {0.400}, mlp = 512-256-128, sl\_epochs = 15, unique\_coeff = {11}, var\_coeff = {13}, ssl\_epochs = 50, wd = {0.568} \\

DeCUR & M1 & batch\_size = 128, common\_coeff = {9}, linear\_lr = {0.598}, lr = {0.072}, mlp = 512-256-128, sk\_epochs = 15, unique\_coeff = {18}, var\_coeff = {7}, ssl\_epochs = 50, wd = {0.169} \\

DeCUR & M2 & batch\_size = 256, common\_coeff = {1}, linear\_lr = {0.027}, lr = {0.540}, mlp = 512-256-128, sl\_epochs = 15, unique\_coeff = {7}, var\_coeff = {18}, ssl\_epochs = 10, wd = {0.669} \\

DeCUR &M4 &  batch\_size = 256, common\_coeff = {1}, linear\_lr = {0.162}, lr = {0.734}, mlp = 512-256-128, sk\_epochs = 15, unique\_coeff = {3}, var\_coeff = {7}, ssl\_epochs = 8, wd = {0.068} \\

\cline{1-3}
VICReg &Baseline  & batch\_size = 256, cov\_coeff = {1.127}, linear\_lr = {0.061}, lr = {0.084}, mlp = 256-128, sl\_epochs = 3, sim\_coeff = {25.148}, std\_coeff = {18.782}, ssl\_epochs = 10, wd = {0.007} \\

Ours &M1   &  batch\_size = 256, cov\_coeff = {21.384}, linear\_lr = {0.085}, lr = {0.075}, mlp = 512-256-128, sl\_epochs = 3, sim\_coeff = {46.996}, std\_coeff = {30.687}, ssl\_epochs = 10, wd = {0.009} \\

Ours &M2 & batch\_size = 256, cov\_coeff = {2.907}, linear\_lr = {0.082}, lr = {0.036}, mlp = 512-256-128, sl\_epochs = 3, sim\_coeff = {31.201}, std\_coeff = {1.694}, ssl\_epochs = 10, wd = {0.004} \\

Ours &M3 & batch\_size = 256, cov\_coeff = {28.682}, linear\_lr = {0.054}, lr = {0.017}, mlp = 512-256-128, sl\_epochs = 3, sim\_coeff = {46.250}, std\_coeff = {47.997}, ssl\_epochs = 11, wd = {0.002} \\

Ours &M4 & batch\_size = 256, cov\_coeff = {8.042}, linear\_lr = {0.027}, lr = {0.022}, mlp = 512-256-128, sl\_epochs = 3, sim\_coeff = {0.597}, std\_coeff = {29.935}, ssl\_epochs = 15, wd = {0.003} \\

\cline{1-3}

DP &M1 & batch\_size = 256, cov\_coeff = {19.755}, linear\_lr = {0.080}, lr = {0.006}, mlp = 256-128, sl\_epochs = 3, sim\_coeff = {0.314}, std\_coeff = {7.061}, ssl\_epochs = 10, wd = {0.004} \\

DP &M2   & batch\_size = 256, cov\_coeff = {40.315}, linear\_lr = {0.088}, lr = {0.058}, mlp = 256-128, sl\_epochs = 3, sim\_coeff = {38.311}, std\_coeff = {45.620}, ssl\_epochs = 10, wd = {0.007} \\

DP &M3   & batch\_size = 64, cov\_coeff = {41.426}, linear\_lr = {0.057}, lr = {0.071}, mlp = 512-256-128, sl\_epochs = 10, sim\_coeff = {23.459}, std\_coeff = {6.917}, ssl\_epochs = 30, wd = {0.047} \\

DP & M4   &  batch\_size = 32, cov\_coeff = {1.467}, linear\_lr = {0.040}, lr = {0.036}, mlp = 256-128, sl\_epochs = 10, sim\_coeff = {7.170}, std\_coeff = {11.042}, ssl\_epochs = 30, wd = {0.049} \\

\bottomrule
\end{tabular}

\end{table*}


\begin{table*}[htbp]
\caption{\textbf{MIMIC-III:} Complete hyper-parameter listings for the different models. These values are obtained by Bayesian optimization over all the given hyperparameters with the purpose to maximize the model performance. For further details, see Sections S1 and S2.
\label{tab:mimic_hparams}}
\centering
\scriptsize
\setlength{\tabcolsep}{3pt}
\begin{tabular}{p{0.08\linewidth} | p{0.05\linewidth} | p{0.75\linewidth}}
\toprule
 \textbf{Model} & \textbf{Variant}  & \textbf{Hyper-parameters} \\ \midrule

CoMM & Baseline & batch\_size = 32, cov\_coeff = {17.325}, epochs\_ft = 13, epochs\_ssl = 11, log\_every = 20, lr\_ft = {0.007}, lr\_ssl = {0.000}, mi\_coeff = {4.415}, mlp = 96-128-50, tau = {0.177}, var\_coeff = {17.007} \\ 

CoMM & M1 & batch\_size = 32, cov\_coeff = {17.386}, epochs\_ft = 9, epochs\_ssl = 11, log\_every = 20, lr\_ft = {0.003}, lr\_ssl = {0.000}, mi\_coeff = {4.633}, mlp = 96-128-50, tau = {0.137}, var\_coeff = {17.485} \\ 

CoMM & M2 & batch\_size = 32, cov\_coeff = {14.804}, epochs\_ft = 10, epochs\_ssl = 10, log\_every = 20, lr\_ft = {0.047}, lr\_ssl = {0.001}, mi\_coeff = {3.062}, mlp = 96-128-50, tau = {0.139}, var\_coeff = {18.542} \\

CoMM &M3   &  batch\_size = 32, cov\_coeff = {4.337}, epochs\_ft = 10, epochs\_ssl = 10, log\_every = 20, lr\_ft = {0.035}, lr\_ssl = {0.000}, mi\_coeff = {4.313}, mlp = 96-128-50, tau = {0.184}, var\_coeff = {18.142} \\

CoMM &M4   &  batch\_size = 32, cov\_coeff = {9.216}, epochs\_ft = 10, epochs\_ssl = 10, log\_every = 20, lr\_ft = {0.080}, lr\_ssl = {0.000}, mi\_coeff = {2.639}, mlp = 96-256-64, tau = {0.193}, var\_coeff = {10.201} \\

 \cline{1-3}

DeCUR & Baseline   & batch\_size = 32, common\_coeff = {18.195}, epochs\_ft = 12, epochs\_ssl = 10, intra\_coeff = {11.025}, lambda\_off = {1.500}, log\_every = 20, lr\_ft = {0.067}, lr\_ssl = {0.001}, mlp = 96-256-64, unique\_coeff = {15.009}, var\_coeff = {8.923} \\ 

DeCUR & M1 & batch\_size = 32, common\_coeff = {5.380}, epochs\_ft = 11, epochs\_ssl = 10, intra\_coeff = {11.564}, lambda\_off = {0.534}, log\_every = 20, lr\_ft = {0.083}, lr\_ssl = {0.001}, mlp = 96-256-64, unique\_coeff = {16.486}, var\_coeff = {2.997} \\ 

DeCUR & M2  & batch\_size = 32, common\_coeff = {14.573}, epochs\_ft = 9, epochs\_ssl = 12, intra\_coeff = {15.948}, lambda\_off = {1.297}, log\_every = 20, lr\_ft = {0.042}, lr\_ssl = {0.000}, mlp = 96-256-64, unique\_coeff = {18.863}, var\_coeff = {6.338} \\

DeCUR & M3 & batch\_size = 32, epochs\_ft = 10, epochs\_ssl = 10, log\_every = 20, lr\_ft = {0.052}, lr\_ssl = {0.001}, tau = {0.181}, w\_shared = {10.371}, w\_unique = {12.329} \\

DeCUR & M4  & batch\_size = 32, epochs\_ft = 10, epochs\_ssl = 10, log\_every = 20, lr\_ft = {0.008}, lr\_ssl = {0.001}, tau = {0.062}, w\_shared = {18.626}, w\_unique = {17.977} \\

 \cline{1-3}
 
FOCAL &Baseline   &  batch\_size = 32, dim\_shared = 48, epochs\_ft = 9, epochs\_ssl = 13, lambda\_c = {1.693}, lambda\_o = {1.181}, lambda\_p = {0.333}, lambda\_t = {1.815}, log\_every = 20, lr\_ft = {0.077}, lr\_ssl = {0.001}, mlp = 96-128-50, temperature = {0.061} \\

FOCAL &M1   &  batch\_size = 32, dim\_shared = null, epochs\_ft = 10, epochs\_ssl = 12, lambda\_c = {1.179}, lambda\_o = {1.882}, lambda\_p = {1.309}, lambda\_t = {0.792}, log\_every = 20, lr\_ft = {0.086}, lr\_ssl = {0.001}, mlp = 96-128-50, temperature = {0.197} \\

FOCAL &M2 & batch\_size = 32, dim\_shared = 48, epochs\_ft = 9, epochs\_ssl = 12, lambda\_c = {1.418}, lambda\_o = {0.825}, lambda\_p = {0.393}, lambda\_t = {0.919}, log\_every = 20, lr\_ft = {0.080}, lr\_ssl = {0.001}, mlp = 96-128-50, temperature = {0.081} \\

FOCAL &M3   &  batch\_size = 32, dim\_shared = null, epochs\_ft = 10, epochs\_ssl = 10, lambda\_c = {1.299}, lambda\_o = {1.814}, lambda\_p = {0.400}, lambda\_t = {0.147}, log\_every = 20, lr\_ft = {0.066}, lr\_ssl = {0.001}, mlp = 96-128-50, temperature = {0.041} \\

FOCAL &M4   &  batch\_size = 32, dim\_shared = 48, epochs\_ft = 10, epochs\_ssl = 10, lambda\_c = {1.946}, lambda\_o = {0.898}, lambda\_p = {0.734}, lambda\_t = {1.961}, log\_every = 20, lr\_ft = {0.095}, lr\_ssl = {0.001}, mlp = 96-128-50, temperature = {0.061} \\

 \cline{1-3}
 
FACTORCL & Baseline   &batch\_size = 32, epochs\_ft = 10, epochs\_ssl = 13, log\_every = 20, lr\_ft = {0.055}, lr\_ssl = {0.001}, tau = {0.129}, w\_shared = {17.887}, w\_unique = {18.501} \\

FACTORCL &M1   &  batch\_size = 32, epochs\_ft = 10, epochs\_ssl = 11, log\_every = 20, lr\_ft = {0.052}, lr\_ssl = {0.000}, tau = {0.174}, w\_shared = {19.363}, w\_unique = {13.611} \\

FACTORCL &M2 & batch\_size = 32, epochs\_ft = 9, epochs\_ssl = 12, log\_every = 20, lr\_ft = {0.035}, lr\_ssl = {0.001}, tau = {0.123}, w\_shared = {9.477}, w\_unique = {17.197} \\

FACTORCL & M3   & batch\_size = 32, epochs\_ft = 10, epochs\_ssl = 12, log\_every = 20, lr\_ft = {0.008}, lr\_ssl = {0.001}, tau = {0.062}, w\_shared = {18.626}, w\_unique = {17.977} \\

FACTORCL &M4   &  batch\_size = 32, epochs\_ft = 10, epochs\_ssl = 10, log\_every = 20, lr\_ft = {0.052}, lr\_ssl = {0.001}, tau = {0.181}, w\_shared = {10.371}, w\_unique = {12.329} \\

 \cline{1-3}
 
SIMCLR & Baseline    & batch\_size = 256, learning\_rate = {0.018}, total\_epoch\_sl = 10, total\_epoch\_ssl = 45, weight\_decay = {0.140} \\

SIMCLR &M1    &  batch\_size = 256, learning\_rate = {0.018}, total\_epoch\_sl = 10, total\_epoch\_ssl = 47, weight\_decay = {0.140} \\

SIMCLR & M2 & batch\_size = 256, learning\_rate = {0.018}, total\_epoch\_sl = 10, total\_epoch\_ssl = 43, weight\_decay = {0.015} \\

SIMCLR &M3    &  batch\_size = 128, learning\_rate = {0.016}, total\_epoch\_sl = 10, total\_epoch\_ssl = 48, weight\_decay = {0.021} \\

SIMCLR &M4 & batch\_size = 256, learning\_rate = {0.019}, total\_epoch\_sl = 10, total\_epoch\_ssl = 49, weight\_decay = {0.041} \\

 \cline{1-3}

VICReg & M1 & batch\_size = 32, cov\_coeff = {1.120}, epochs\_ft = 12, epochs\_ssl = 13, log\_every = 20, lr\_ft = {0.029}, sim\_coeff = {33.254}, std\_coeff = {18.252} \\

VICReg & M2 & batch\_size = 32, cov\_coeff = {1.017}, epochs\_ft = 10, epochs\_ssl = 10, log\_every = 20, lr\_ft = {0.076}, sim\_coeff = {43.671}, std\_coeff = {47.167} \\

VICReg & M3 & batch\_size = 32, cov\_coeff = {1.001}, epochs\_ft = 10, epochs\_ssl = 18, log\_every = 20, lr\_ft = {0.071}, sim\_coeff = {19.808}, std\_coeff = {19.824} \\

VICReg & M4 & batch\_size = 32, cov\_coeff = {1.000}, epochs\_ft = 10, epochs\_ssl = 15, log\_every = 20, lr\_ft = {0.024}, sim\_coeff = {42.756}, std\_coeff = {13.698} \\

\bottomrule
\end{tabular}

\end{table*}

\begin{table*}[htbp]
\caption{\textbf{MODMA:} Complete hyper-parameter listings for the different models. These values are obtained by Bayesian optimization over all the given hyperparameters with the purpose to maximize the model performance. For further details, see Sections S1 and S2.
\label{tab:modma_hparams}}
\centering
\scriptsize
\setlength{\tabcolsep}{3pt}
\begin{tabular}{p{0.08\linewidth} | p{0.05\linewidth} | p{0.75\linewidth}}
\toprule
\textbf{Model} & \textbf{Variant} & \textbf{Hyper-parameters} \\ \midrule

CoMM & Baseline  & cov\_coeff = {8.523}, ssl\_epoch = 10, , sl\_epoch = 9, learning\_rate = {4.306}, mi\_coeff = {8.029}, mlp = 512-256-1024, patience = 5, subjects\_per\_batch = 4, tau = {0.070}, var\_coeff = {9.831}, weight\_decay = {0.629} \\

CoMM & M1  & cov\_coeff = {5.440}, ssl\_epoch = 20, , sl\_epoch = 10, learning\_rate = {2.662}, mi\_coeff = {6.071}, mlp = 512-256-1024, patience = 5, subjects\_per\_batch = 2, tau = {0.070}, var\_coeff = {6.899}, weight\_decay = {0.286} \\

CoMM & M2   &  cov\_coeff = {6.483}, ssl\_epoch = 7, sl\_epoch = 8, learning\_rate = {2.388}, mi\_coeff = {7.884}, mlp = 512-256-1024, patience = 5, subjects\_per\_batch = 4, tau = {0.038}, var\_coeff = {7.816}, weight\_decay = {3.122} \\

CoMM & M3   & cov\_coeff = {8.513}, ssl\_epoch = 8,, sl\_epoch = 9,  learning\_rate = {3.159}, mi\_coeff = {8.167}, mlp = 512-256-1024, patience = 6, subjects\_per\_batch = 4, tau = {0.066}, var\_coeff = {7.511}, weight\_decay = {2.893} \\

 CoMM &M4   & cov\_coeff = {8.446}, ssl\_epoch = 4, , sl\_epoch = 8, learning\_rate = {3.210}, mi\_coeff = {6.574}, mlp = 512-256-1024, patience = 6, subjects\_per\_batch = 4, tau = {0.039}, var\_coeff = {2.266}, weight\_decay = {1.433} \\

 \cline{1-3}

FACTORCL & Baseline   & ssl\_epoch = 20, sl\_epoch = 4, , inv\_coeff = {8}, learning\_rate = {1.327}, mlp = 512-1024-256, patience = 6, shared\_coeff = {6}, subjects\_per\_batch = 3, tau = {0.000}, weight\_decay = {0.687} \\

FACTORCL & M2   & ssl\_epoch = 20, sl\_epoch = 2, inv\_coeff = {8}, learning\_rate = {1.451}, mlp = 512-1024-256, patience = 3, shared\_coeff = {10}, subjects\_per\_batch = 3, tau = {0.001}, weight\_decay = {0.005} \\

FACTORCL &M1   &  ssl\_epoch = 20, sl\_epoch = 3, inv\_coeff = {9}, learning\_rate = {1.349}, mlp = 512-256-1024, patience = 4, shared\_coeff = {6}, subjects\_per\_batch = 3, tau = {0.000}, weight\_decay = {0.311} \\

FACTORCL &M3   & ssl\_epoch = 5, sl\_epoch = 3, sl\_epoch = 10, inv\_coeff = 9, learning\_rate = {3.859}, mlp = 512-1024-256, patience = 3, shared\_coeff = 10, subjects\_per\_batch = 3, tau = {0.001}, weight\_decay = {0.882} \\

FACTORCL &M4   &  ssl\_epoch = 20, sl\_epoch = 3, inv\_coeff = 9, learning\_rate = {1.867}, mlp = 512-1024-256, patience = 6, shared\_coeff = 10, subjects\_per\_batch = 3, tau = {0.000}, weight\_decay = {0.116} \\

 \cline{1-3}

QUEST &Baseline   &  align\_coeff = 1, ssl\_epoch = 10, , lp\_epochs = 2, learning\_rate = {0.961}, mlp = 512-256-1024, orth\_coeff = 7, patience = 6, sp\_coeff = 10, subjects\_per\_batch = 1, tau = 6, uniq\_coeff = 10, var\_coeff = 3, weight\_decay = {1.014} \\

QUEST &M1   &  align\_coeff = {6}, ssl\_epoch = 10, , lp\_epochs = 3, learning\_rate = {0.047}, mlp = 512-256-1024, orth\_coeff = {5}, patience = 6, subjects\_per\_batch = 2, uniq\_coeff = {1}, var\_coeff = {5}, weight\_decay = {0.190} \\

QUEST & M2   &  align\_coeff = 9, ssl\_epoch = 20, , lp\_epochs = 2, learning\_rate = {3.852}, mlp = 512-256-1024, orth\_coeff = 4, patience = 5, sp\_coeff = 6, subjects\_per\_batch = 2, tau = 4, uniq\_coeff = 4, var\_coeff = 9, weight\_decay = {1.414} \\

QUEST &M3   &  align\_coeff = 3, ssl\_epoch = 10, lp\_epochs = 3, learning\_rate = {3.437}, mlp = 512-256-1024, orth\_coeff = 6, patience = 5, sp\_coeff = 9, subjects\_per\_batch = 2, tau = 3, uniq\_coeff = 3, var\_coeff = 3, weight\_decay = {0.094} \\

QUEST &M4   &  align\_coeff = 10, ssl\_epoch = 20, lp\_epochs = 3, learning\_rate = {4.204}, mlp = 512-256-1024, orth\_coeff = 4, patience = 5, sp\_coeff = 10, subjects\_per\_batch = 2, tau = 7, uniq\_coeff = 7, var\_coeff = 7, weight\_decay = {2.412} \\

 \cline{1-3}
 DeCUR   & Baseline & embed\_dim = 256, ssl\_epoch = 20, lp\_epochs = 2,  lambd = {0.005}, learning\_rate = {0.797}, mlp = 512-256-1024, patience = 5, subjects\_per\_batch = 1, weight\_decay = {1.766} \\

DeCUR & M1   & embed\_dim = 256, ssl\_epoch = 20, lp\_epochs = 2, lambd = {0.005}, learning\_rate = {1.384}, mlp = 512-256-1024, patience = 5, subjects\_per\_batch = 1, weight\_decay = {3.152} \\

DeCUR & M2 & embed\_dim = 256, ssl\_epoch = 20, lp\_epochs = 3, lambd = {0.576}, learning\_rate = {0.597}, mlp = 512-256-1024, patience = 5, subjects\_per\_batch = 1, weight\_decay = {0.258} \\

DeCUR & M3   &  embed\_dim = 256, ssl\_epoch = 20, lp\_epochs = 3, lambd = {0.365}, learning\_rate = {0.684}, mlp = 512-256-1024, patience = 5, subjects\_per\_batch = 1, weight\_decay = {0.298} \\

DECUR & M4   & embed\_dim = 256, ssl\_epoch = 5, lp\_epochs = 3, lambd = {0.803}, learning\_rate = {0.650}, mlp = 512-256-1024, patience = 6, subjects\_per\_batch = 1, weight\_decay = {0.602} \\

 \cline{1-3}
 
VICReg & Baseline   &  cov\_coeff = {3}, ssl\_epoch = 10,  lp\_epochs = 3, learning\_rate = {0.845}, mlp = 512-256-1024, patience = 3, sim\_coeff = {0}, std\_coeff = 68, subjects\_per\_batch = 3, weight\_decay = {1.554} \\

VICReg &M1   &  base\_lr = {1.018}, batch\_size = 32, cov\_coeff = {46.202}, ssl\_epochs = 30, lp\_epochs = 2, mlp = 8192-8192-8192, sim\_coeff = {21.315}, std\_coeff = {35.875}, wd = {1.083} \\

VICReg &M2   & base\_lr = {1.086}, batch\_size = 32, cov\_coeff = {49.345}, ssl\_epochs = 20, lp\_epochs = 3, mlp = 4096-4096-4096, sim\_coeff = {27.695}, std\_coeff = {18.361}, wd = {1.087} \\
  
VICReg &M3   & base\_lr = {1.079}, batch\_size = 32, cov\_coeff = {31.326}, ssl\_epochs = 20, lp\_epochs = 3, mlp = 4096-4096-4096, sim\_coeff = {38.462}, std\_coeff = {22.943}, wd = {1.081} \\
   
VICReg & M4   & ov\_coeff = {87}, ssl\_epoch = 20, , lp\_epochs = 5,learning\_rate = {2.661}, mlp = 512-256-1024, patience = 4, sim\_coeff = {67}, std\_coeff = {18}, subjects\_per\_batch = 2, weight\_decay = {1.094} \\

\bottomrule
\end{tabular}
\end{table*}


\section{Extended Related Work}
\label{app:extended_related_work}

\subsection{Multimodal Fairness}
\label{app:multimodal_ssl_related_work}
Fairness is a multifaceted issue \cite{cheong2021hitchhiker,kuzucu2024uncertainty}. Existing works chiefly investigated multimodal fairness
across ML models trained using multiple data modalities.
\citet{booth2021bias} demonstrated how using multiple modalities marginally improves prediction at the cost of reducing fairness for automated video interviews. 
\citet{schmitz2022bias} studied how different multimodal approaches affect gender bias in emotion recognition. 
\citet{janghorbani2023multi} presented a visual-textual benchmark dataset to assess the bias present in existing multimodal models.
%
%
\citet{pena2023human} presented a new dataset of synthetic resumes to evaluate demographic bias in
multimodal ML.
\citet{kathan2022personalised} 
and \citet{alasadi_cyberbullying}
proposed a weighted fusion approach 
to achieve fairness in audiovisual humour recognition.
whereas \citet{yan2020mitigating} focused on adversarial bias mitigation for 
personality assessment. 
\citet{alasadi_cyberbullying} proposed a fairness-aware fusion framework for cyberbullying detection using a weighted approach. 
\citet{chen2023fmmrec} proposed a fairness-aware method for multimodal recommendations.
\citet{cheong2024fairrefuse} 
proposed a causal-based multimodal fusion network 
for depression detection.
%


\subsection{Self-Supervised Learning for Fairness}
\label{app:ssl_fairness_related_work} 

Fairer SSL methods have only been investigated within a unimodal setting.
\citet{yfantidou2024using} demonstrated that SSL can significantly improve model
fairness, while maintaining performance on par with supervised
method.
\citet{chai2022self} proposed a
novel reweighing-based contrastive learning method to learn a generally fair representation without observing sensitive attributes.
\citet{ma2021conditional} proposed a Conditional Contrastive Learning
(CCL) approach by sampling 
samples positive and negative pairs from distributions conditioning on the sensitive attribute to improve the fairness of contrastive SSL methods. 
\citet{chakraborty2022fair} proposed a semi-supervised method which uses a small proportion of labelled data as input in order to generate pseudo-lables for unlabelled data.

\subsection{Fairness in Healthcare}
\label{app:fair_health_related_work}

Several works
have investigated 
ML fairness across a variety of health and wellbeing settings ranging from 
chest x-ray analysis \cite{zhang2022improving,seyyed2020chexclusion}, 
affect analysis \cite{cheong2022counterfactual,cheong2023causal,cheong_acii,cheong2025small,akgul2026investigating,green2025gender}
to depression detection \cite{cheong2024fairrefuse,kwok2025machine,cameron2024multimodal}.
However, most of the studies have chiefly focused on a unimodal setup. 
As healthcare systems become increasingly integrated \cite{yildirim2024multimodal,dai2025climb,daifairgrpo,dai2025fairgrpo}, 
investigating multimodal fairness in ML for healthcare becomes increasingly relevant and pressing.
%
Our work is distinct from recent related efforts.
Compared with \citet{luo2024fairclip,zhang2025joint}, our focus is not CLIP type vision-language debiasing, but fairness-aware multimodal self-supervised representation learning for heterogeneous, variable-length modalities.
Compared with  \citet{ye2024continual,shao2026vmfcoop} our focus is not continual or universal multimodal medical pretraining, but how subject-aware pooling and subject-aware VICReg regularization can improve fairness under heterogeneity, together with a corresponding theoretical analysis. 
%


\end{document}